%% file: neurips_26_camera_ready.tex
\documentclass{article}
\PassOptionsToPackage{numbers,sort&compress}{natbib}
\usepackage[main,final]{neurips_2026}
\usepackage[utf8]{inputenc}
\usepackage[T1]{fontenc}
\usepackage{xcolor}
\definecolor{mydarkred}{rgb}{0.6,0,0}
\definecolor{mydarkgreen}{rgb}{0,0.6,0}
\usepackage{amsmath,amsfonts,amssymb,amsthm,mathtools}
\usepackage{booktabs}
\usepackage{array}
\usepackage{longtable}
\usepackage{float}
\usepackage{nicefrac}
\usepackage{microtype}
\usepackage{pifont}
\usepackage{fontawesome5}
\usepackage{tikz}
\usetikzlibrary{positioning,patterns,arrows.meta,calc,shapes.geometric}
\usepackage{hyperref}
\hypersetup{colorlinks, linkcolor=mydarkred, citecolor=mydarkgreen}

\theoremstyle{plain}
\newtheorem{theorem}{Theorem}
\newtheorem{proposition}{Proposition}
\newtheorem{lemma}{Lemma}
\newtheorem{corollary}{Corollary}

\theoremstyle{definition}
\newtheorem{definition}{Definition}
\newtheorem{example}{Example}

\theoremstyle{remark}
\newtheorem{remark}{Remark}

\theoremstyle{plain}

\newtheorem{condition}{Condition}

\usepackage[capitalize,noabbrev]{cleveref}

\crefname{theorem}{Theorem}{Theorems}
\Crefname{theorem}{Theorem}{Theorems}
\crefname{proposition}{Proposition}{Propositions}
\Crefname{proposition}{Proposition}{Propositions}
\crefname{lemma}{Lemma}{Lemmas}
\Crefname{lemma}{Lemma}{Lemmas}
\crefname{corollary}{Corollary}{Corollaries}
\Crefname{corollary}{Corollary}{Corollaries}
\crefname{definition}{Definition}{Definitions}
\Crefname{definition}{Definition}{Definitions}
\crefname{example}{Example}{Examples}
\Crefname{example}{Example}{Examples}
\crefname{remark}{Remark}{Remarks}
\Crefname{remark}{Remark}{Remarks}
\crefname{assumption}{Assumption}{Assumptions}
\Crefname{assumption}{Assumption}{Assumptions}
\crefname{condition}{Condition}{Conditions}
\Crefname{condition}{Condition}{Conditions}
\renewcommand{\eqref}[1]{Eqn.~\textup{(\ref{#1})}}

\newcommand{\Risk}{R_{\rm T}}
\newcommand{\RiskStar}{\Risk^{\ast}}
\newcommand{\RiskClsOptimum}{\Risk^{\FMR, \ast}}
\newcommand{\EstRisk}{\mathcal E_{\rm T}^{\mathrm{est}}}
\newcommand{\Loss}{\mathcal{L}_{{\rm T},\ell}}
\newcommand{\LossStar}{\mathcal{L}_{{\rm T},\ell}^{\ast}}
\newcommand{\FMR}{\mathcal F_{\mathrm{MR}}}
\newcommand{\Rreach}{\mathcal R}
\newcommand{\tauhard}{\tau_{\mathrm{hard}}}
\newcommand{\tauscore}{\tau_{\mathrm{score}}}
\newcommand{\epshard}{\epsilon_{\mathrm{hard}}}
\newcommand{\epsscore}{\epsilon_{S,\ell}}
\newcommand{\Deltahard}{\mathrm{disc}_{\mathrm{hard}}}
\newcommand{\Deltascore}{\mathrm{disc}_{\ell}}
\newcommand{\lamhard}{\mathrm{mis}_{\mathrm{hard}}}
\newcommand{\lamscore}{\mathrm{mis}_{\mathrm{score}}}
\newcommand{\PihatR}{\Pi_{\mathcal R}}
\newcommand{\PihatH}{\Pi_{\mathcal H\mid\mathcal R}}

\newcommand{\Prob}{\mathbb P}
\newcommand{\E}{\mathbb E}
\newcommand{\one}{\mathbf 1}
\newcommand{\Ndim}{\operatorname{Ndim}}
\newcommand{\Hfamily}{\mathcal H}
\newcommand{\Gfamily}{\mathcal G}
\newcommand{\Wclass}{\mathcal W}

\newcommand{\PihatHall}{\Pi_{\mathcal H}^{\mathrm{all}}}
\newcommand{\lexargmax}{\operatorname{lexargmax}}
\newcommand{\CAff}{\mathcal C_{\mathrm{Aff}}}
\newcommand{\CRelu}{\mathcal C_{\mathrm{ReLU}}}
\newcommand{\Phip}{\Phi_p}
\newcommand{\tR}{t_{\mathcal R}}
\newcommand{\dHR}{d_{\mathcal H}^{\mathcal R}}
\newcommand{\reachset}{T}

\newcommand{\fs}{f_{\rm S}}
\newcommand{\XS}{\mathcal{X}_{\rm S}}

\newcommand{\XT}{\mathcal{X}_{\rm T}}

\newcommand{\PT}{P_{\rm T}}
\newcommand{\PS}{P_{\rm S}}

\newcommand{\ReaClass}{\mathrm{Rea}_{\Gfamily}}
\newcommand{\RelClass}{\mathrm{Rel}_{\Hfamily}}
\newcommand{\nT}{n_{\rm T}}

\usepackage[most]{tcolorbox}
\tcolorboxenvironment{theorem}{
  colback=white,
  colframe=black,
  boxrule=0.75pt,
  arc=1.5pt,
  left=0.5pt,
  right=0.5pt,
  top=0.75pt,
  bottom=0.75pt
}

\definecolor{SteelBlue}{RGB}{70, 130, 180}
\newtcolorbox{takeaway}{
    enhanced,
    breakable,                
    colback=SteelBlue!8, 
    frame hidden,             
    borderline west={2pt}{0pt}{black!70}, 
    sharp corners,            
    fontupper=\small\itshape, 
    left=5pt, right=5pt,
    before skip=2pt, after skip=2pt
}

\newcommand{\appendixtocname}{Appendix Contents}

\makeatletter
\newcommand{\listofappendices}{%
  \section*{\appendixtocname}
  \@starttoc{atoc}
}

\newcommand{\appsection}[1]{%
  \section{#1}%
  \addcontentsline{atoc}{section}{\protect\numberline{\thesection}#1}%
}

\makeatother

\title{On the Learnability of Model Reprogramming: \\A Reachability and Relabeling Perspective}

\author{%
  Zesheng Ye$^{1}$ \quad Pin-Yu Chen$^{2}$ \quad Feng Liu$^{1}$ \\
  $^{1}$The University of Melbourne \quad $^{2}$IBM Research \\
  \texttt{\{zeshengye.ml, fengliu.ml\}@gmail.com}, \quad \texttt{pin-yu.chen@ibm.com}
}

\begin{document}
\maketitle

\begin{abstract}
Model reprogramming adapts a {\em frozen} source model to a target task by wrapping it in a {\em learnable} input transformation from target inputs to the source-model input space and an output mapping from source-model outputs to target predictions.
Despite extensive empirical progress, the {\em finite-sample learnability} of the induced wrapper hypothesis class $\FMR$ remains underexplored.
We address this question for source models that expose hard source labels and analyze it through the wrapper structure: on any target sample, the input transformation selects {\em reachable} source labels that the output mapping relabels.
Through this lens, we answer two questions: \ding{172} how many target samples are needed to approach the best predictor in $\FMR$, and \ding{173} what gap remains between the optimum over $\FMR$ and the target Bayes risk.
For \ding{172}, the wrapper structure factorizes the sample-wise complexity of $\FMR$ into a {\em reachability} term counting reachable source-label patterns and a {\em conditional relabeling} term counting their relabelings.
This yields an agnostic-PAC bound on the {\em estimation error}, with sample complexity scaling additively in the two terms; this bound becomes closed-form on additive input transformations and is matched by a minimax lower bound on an explicit binary classification construction.
For \ding{173}, the same structure decomposes the {\em approximation error} exactly into source-label information loss and output-mapping restriction, together with a source-conditional upper bound on this gap.
We extend the approximation analysis to logit-exposing source models, where the upper bound is empirically estimable, and verify both the estimation and approximation mechanisms in controlled synthetic and pretrained-MNIST experiments.
The wrapper-side analysis localizes both errors to specific components, giving a structural view of reprogramming practice. Code: \href{https://github.com/zyecs/repgm-slt}{\faGithub}.
\end{abstract}

\section{Introduction}\label{sec:intro}

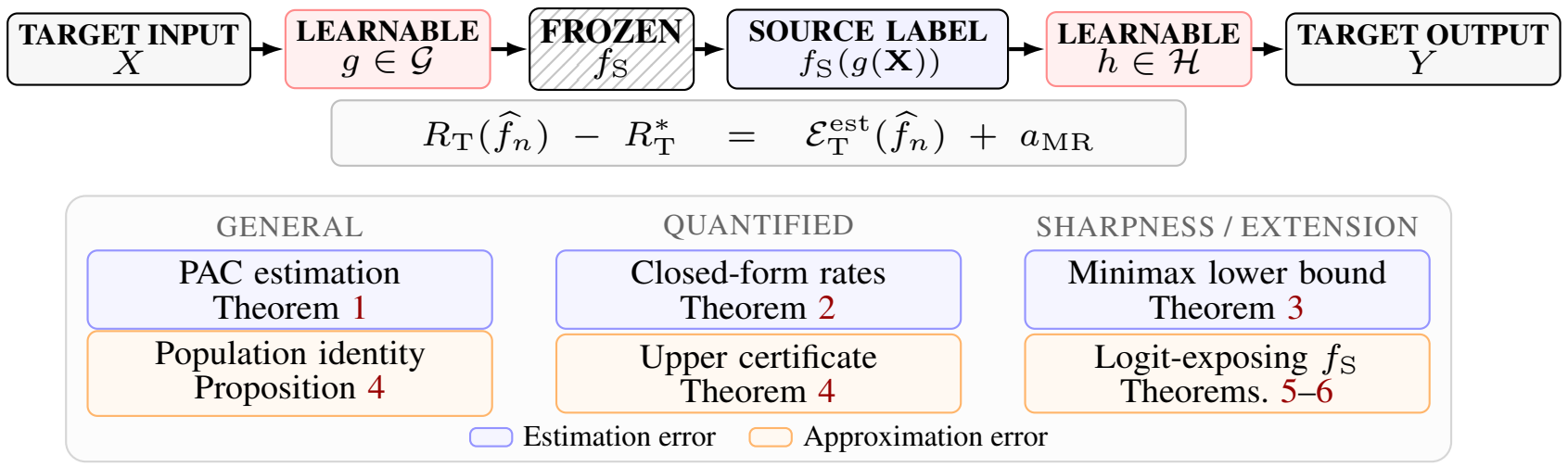
\begin{figure}[t]
  \centering
  \setlength{\belowcaptionskip}{-1.8em}
  \resizebox{0.80\textwidth}{!}{\input{figures/frozen_access_funnel.tex}}
  \caption{We present a {\em finite-sample learnability} analysis for hard-label MR through the MR wrapper structure, organized by the {\em estimation error} (\Cref{sec:T1,sec:T1P,sec:T3P}) and {\em approximation error} (\Cref{sec:T2}) perspectives, respectively.}
\label{fig:organization}
\end{figure}

As pre-trained machine learning models grow in scale, much of their value lies in capabilities that are difficult or expensive to reproduce by retraining~\citep{brown2020language, hoffmann2022training, zhou2023comprehensive, zhao2023survey, gu2023systematic, ye2025neural}.
Meanwhile, a rising fraction of them is accessible only through restricted interfaces, e.g., released checkpoints or even deployment-time inference APIs, making adaptation methods that do not modify source weights, and can operate under restricted access, increasingly important~\citep{houlsby2019parameter, lester2021power, luo2024incontext}.
{\em Model reprogramming}~(MR)~\citep{elsayed2019adversarial, neekhara2019adversarial, chen2024model} stands out as an active model-reuse paradigm for this problem.
It adapts a {\em frozen} pre-trained model (i.e., a source model) to a downstream task (i.e., a target task) by wrapping it in a composite predictor\footnote{This paper focuses on the settings where the reprogrammed model is a {\em classifier}.} with a {\em learnable input transformation} that maps target inputs into the source-model input space, and a {\em learnable output mapping} that converts source-model outputs (e.g., source labels) into target predictions (\Cref{fig:organization}).
This wrapper architecture is considered valuable in frozen-source-model reuse scenarios where target-task data are limited~\citep{chen2024model} and has succeeded across image~\citep{elsayed2019adversarial, chen2023understanding, cai2024sample, cai2024bayesian}, text~\citep{neekhara2019adversarial}, time-series~\citep{tsai2020transfer, yang2021voice2series}, protein sequences~\citep{melnyk2023reprogramming, vinod2025reprogramming}, and cross-modal~\citep{neekhara2022cross} adaptation tasks.

\textbf{Gap.}
The empirical progress of MR, however, runs ahead of the theory: particularly, {\em finite-sample} understanding of {\em what controls the behavior of such a wrapper}, the regime that limited-target-data use cases demand, remains underexplored.
Existing reprogramming theory has not yet addressed this directly, but relies on structural assumptions on the source model, including source--target alignment~\citep{yang2021voice2series}, specific source-model architectures~\citep{englert2022adversarial}, and spectral regularity of the linearized source model~\citep{chung2025model}.
Such routes characterize which source models MR may succeed on, but they do not isolate which properties of the {\em reprogramming wrapper} itself control its finite-sample behavior.

According to~\citep{vapnik2013nature, mohri2018foundations}, understanding that behavior on a target distribution amounts to controlling the wrapper's target excess risk~\citep{vapnik2013nature}, split into two complementary parts, namely the {\em estimation error}, which measures the gap from the learned predictor to the best predictor in the MR-induced wrapper hypothesis class $\FMR$, and the {\em approximation error}, which measures the gap from that in-class optimum to the target Bayes risk~\citep{vapnik1999overview}.
Each part gives rise to a corresponding question naturally:
\begin{list}{}{%
  \setlength{\leftmargin}{0pt}%
  \setlength{\labelsep}{0.3em}%
  \setlength{\itemsep}{2pt}%
  \setlength{\topsep}{1pt}%
  \setlength{\parsep}{0pt}%
}
  \item[] \ding{172} How many target samples does the learned predictor need to approach the best predictor in $\FMR$?
  \item[] \ding{173} What residual gap remains between the optimum over $\FMR$ and the target Bayes risk, even with unlimited target samples?
\end{list}
We address this for source models that expose hard source labels, i.e., {\em hard-label} MR~\citep{elsayed2019adversarial, tsai2020transfer, chen2023understanding}, grounded in the wrapper structure (defined in \Cref{sec:prelim:wrapper}):
when the source model is fixed, only the input transformation and the output mapping are learnable, and they act sequentially.
Thus, on each target sample, the input transformation determines which source-model output (i.e., a source label) reaches the output mapping, which then relabels those reachable patterns into target labels.
This MR-specific structure governs both questions, and our answers below trace its consequences.

\textbf{Answer to \ding{172}.}
We discover that sample-wise complexity of $\FMR$ factorizes into a {\em reachability} term and a {\em conditional relabeling} term, each recording the realizable behavior of one wrapper component on the target sample.
This factorization yields an agnostic {\em probably approximately correct}~(PAC) bound~\citep{vapnik2013nature} on the {\em estimation error} of approximate {\em empirical risk minimization}~(ERM) over $\FMR$ (\Cref{thm:T1}), with sample complexity scaling additively in the growth of these two terms.
On additive input transformations~\citep{elsayed2019adversarial, tsai2020transfer, bahng2022exploring, chen2023understanding, tsao2023autovp}, both terms admit closed-form rates under structural conditions on the source model (\Cref{thm:T1P}), verified for affine and finite-depth ReLU source-score maps (Corollaries~\ref{cor:T1P:affine}-\ref{cor:T1P:relu}).
We further prove the necessity of this sample-complexity dependence via a matching minimax lower bound~\citep{yu1997assouad} on an explicit binary classification construction, whose Bayes-optimal predictor lies in $\FMR$, so that the lower bound applies to the estimation error (\Cref{thm:T3P}).

\textbf{Answer to \ding{173}.}
We establish that the {\em approximation error} of $\FMR$ admits an exact population identity, decomposing it into the information loss when source labels do not preserve enough target structure for any relabeling to recover the target Bayes rule, and the additional loss when the output-mapping class does not contain that Bayes relabeling (Proposition~\ref{prop:T2P}).
Substituting Bayes selectors of the source and target distributions into the Proposition~\ref{prop:T2P} gives three source-conditional quantities, whose sum upper-bounds the residual gap (Theorem~\ref{thm:T2C}).
We extend the same decomposition to logit-exposing source models, with similar population identities (\cref{prop:score-01-identity,prop:score-surrogate-identity}) and empirically estimable upper bounds on the residual (\cref{thm:score-certificate,cor:score-finite-library,thm:score-rademacher}).

\textbf{Contribution.}
We give the first {\em wrapper-side} finite-sample learnability analysis for hard-label MR.
By analyzing the reprogramming wrapper itself, our framework isolates reachability and conditional relabeling, the two structural components that control finite-sample behavior of MR, and we verify both mechanisms in controlled synthetic and pretrained-MNIST experiments (\Cref{sec:empirical}).
Beyond the bounds themselves, our results yield a theoretical perspective to understand the design choices and failure cases of existing MR practices (\Cref{sec:discussion}).
The wrapper-side viewpoint opens finite-sample MR analysis to source models accessed only through their input--output interface, complementing existing MR theory that focuses on source-model properties~\citep{yang2021voice2series, englert2022adversarial, chung2025model}.

\section{Hard-Label Model Reprogramming}
\vspace{-0.5em}
\label{sec:prelim}

We formalize the hard-label MR setup and isolate the two combinatorial objects on which our analysis rests.
\Cref{sec:prelim:wrapper} fixes the MR-induced hypothesis class and the approximate-ERM learner;
\Cref{sec:prelim:excess} sets up the estimation-approximation decomposition;
\Cref{sec:prelim:growth} introduces the sample-wise growth functions whose factorization drives \Cref{sec:T1}.
We summarize the notations in Appendix~\ref{app:assumptions}.

\subsection{Formulation of Model Reprogramming}
\label{sec:prelim:wrapper}
Let $\XT$ and $\XS$ denote measurable target and source input spaces with finite label alphabets $[K_{\rm T}]$ and $[K_{\rm S}]$, respectively, where $[K] := \{1, \ldots, K\}$.
Throughout the main text, $\fs: \XS \to [K_{\rm S}]$ is a fixed measurable hard-label source classifier.

{\em Model reprogramming}~(MR)~\citep{chen2024model} repurposes $\fs$ for a target task governed by a distribution $\PT$ over $\XT \times [K_{\rm T}]$ by composing $\fs$ with two \emph{learnable} wrapper stages: an \emph{input transformation} $g \in \Gfamily$ (a nonempty class of measurable maps $\XT \to \XS$) maps target inputs into $\XS$, and an \emph{output mapping} $h \in \Hfamily \subseteq [K_{\rm T}]^{[K_{\rm S}]}$ (nonempty) converts source labels into target predictions.
Each $(g, h) \in \Gfamily \times \Hfamily$ defines a reprogrammed predictor $f_{g,h} := h \circ \fs \circ g$ (abbreviated $f$).
The MR-induced hypothesis class collects all such compositions:
\begin{equation*}
    \FMR := \{ f_{g,h} = h \circ \fs \circ g : g \in \Gfamily, h \in \Hfamily \} \subseteq [K_{\rm T}]^{\XT}.    
\end{equation*}
Given an i.i.d.\ target sample $\{(X_i, Y_i)\}_{i=1}^{\nT} \sim \PT^{\nT}$, write $\widehat R_n(f) := \nT^{-1} \sum_{i=1}^{\nT} \one\{f(X_i) \neq Y_i\}$ for the empirical $0$--$1$ risk.
A measurable predictor $\widehat f_n \in \FMR$ is an \emph{$\eta_n$-approximate ERM} if $\widehat R_n(\widehat f_n) \le \inf_{f \in \FMR} \widehat R_n(f) + \eta_n$ ($\eta_n \ge 0$).
Unless stated otherwise, all results below quantify over every target distribution $\PT$ over $\XT \times [K_{\rm T}]$, every fixed measurable $\fs: \XS \to [K_{\rm S}]$, and every nonempty measurable wrapper pair $(\Gfamily, \Hfamily)$.
\subsection{Estimation error and approximation residual}
\label{sec:prelim:excess}
For a target predictor $f: \XT \to [K_{\rm T}]$, write $\Risk(f) := \Prob_{(X,Y) \sim \PT}(f(X) \neq Y)$ for its population $0$--$1$ risk.
Comparing a learned wrapper $\widehat f_n \in \FMR$ against the in-class optimum $\RiskClsOptimum := \inf_{f \in \FMR} \Risk(f)$ isolates finite-sample estimation cost, whereas comparing $\RiskClsOptimum$ against the unrestricted Bayes risk $\RiskStar := \inf_{f} \Risk(f)$ over all measurable $f: \XT \to [K_{\rm T}]$ isolates the structural expressivity limit of $\FMR$.

\begin{definition}[Estimation--approximation error decomposition~\citep{shalev2014understanding}]
\label{def:errors}
For every measurable predictor $\widehat f_n: \XT \to [K_{\rm T}]$, the target {\em estimation error} relative to $\FMR$ is $\EstRisk(\widehat f_n) := \Risk(\widehat f_n) - \RiskClsOptimum$ and the {\em approximation error} of $\FMR$ is $a_{\mathrm{MR}} := \RiskClsOptimum - \RiskStar$, so that $\Risk(\widehat{f}_n) - \RiskStar = \EstRisk(\widehat{f}_n) + a_{\mathrm{MR}}$.
\end{definition}
The {\em estimation error} measures the finite-sample learning gap between the predictor obtained from $n$ target examples and the best predictor in $\FMR$.
The {\em approximation error} measures the residual gap between the best attainable risk obtained by any predictor within $\FMR$ and the target Bayes risk and cannot be reduced even with unlimited target data.
In this paper, we handle both terms through the {\em two-stage} pipeline of the MR wrapper itself: $x \xrightarrow{g} g(x) \xrightarrow{\fs} \fs(g(x)) \xrightarrow{h} h(\fs(g(x)))$.
Namely, on any target sample, $g$ first determines which source labels are exposed by $\fs$, and $h$ then relabels them.
We use this structure to both {\em decompose} the finite-sample complexity $\FMR$ that controls $\EstRisk$ and the population residual $a_{\rm MR}$, in each case into matching input-side and output-side contributions.
We next introduce the combinatorial objects entering the finite-sample complexity analysis.

\subsection{Reachable source-label patterns and conditional relabelings}
\label{sec:prelim:growth}
On a fixed target tuple $x_{1:m} := (x_1, \ldots, x_m) \in \XT^m$, the two-stage pipeline induces two coupled pattern sets: the source-label vectors reachable by varying $g \in \Gfamily$ through $\fs$, and, for each such reachable vector $z$, the target-label vectors realizable by varying $h \in \Hfamily$.
\begin{definition}[Reachable and realizable label patterns]
\label{def:patterns}
Let $\Rreach := \{\, r := \fs \circ g : g \in \Gfamily \,\}$ denote the reachable source-label predictor class.
For every integer $m \ge 1$, every target tuple $x_{1:m} = (x_1, \ldots, x_m) \in \XT^m$, and every source-label vector $z = (z_1, \ldots, z_m) \in [K_{\mathrm S}]^m$, the reachable source-label pattern set on $x_{1:m}$ and the realizable target-label pattern set on $z$ are
\begin{equation*}
    \ReaClass(x_{1:m}) := \{(r(x_1),\ldots,r(x_m)) : r \in \Rreach\}, \quad
    \RelClass(z) := \{(h(z_1),\ldots,h(z_m)) : h \in \Hfamily\}.
\end{equation*}
\end{definition}
For a multiclass hypothesis class $\mathcal F \subseteq [K]^{\mathcal X}$, write $\Pi_{\mathcal F}(m) := \sup_{x_{1:m} \in \mathcal X^m} |\{(f(x_1),\ldots,f(x_m)) : f \in \mathcal F\}|$ for its sample-wise growth function \citep{shalev2014understanding}.
Applying worst-case sample-wise counting to $\ReaClass(x_{1:m})$ and to $\RelClass(z)$ over reachable $z \in \ReaClass(x_{1:m})$ yields the two complexity measures.

\begin{definition}[Reachability and conditional-relabeling growth]
\label{def:growth}
For every $m \ge 1$, the {\em sample-wise reachability growth} and {\em conditional-relabeling growth} of $\FMR$ are $\PihatR(m) := \sup_{x_{1:m} \in \XT^m} |\ReaClass(x_{1:m})|$ and $\PihatH(m) := \sup_{x_{1:m} \in \XT^m} \sup_{z \in \ReaClass(x_{1:m})} |\RelClass(z)|$.
\end{definition}

\begin{remark}
    Because the inner supremum in $\PihatH(m)$ ranges only over reachable patterns $z \in \ReaClass(x_{1:m})$, relaxing it to all $z \in [K_{\rm S}]^m$ defines the {\em structure-oblivious} proxy $\PihatHall(m) := \sup_{z \in [K_{\rm S}]^m} |\RelClass(z)|$, which can exponentially over-count head complexity. 
    We revisit the distinction in \Cref{sec:T1:conditional-growth} and show that the gap between $\PihatH$ and $\PihatHall$ can be large in regimes practically relevant to MR.
\end{remark}

\section{Factorized PAC Bound on the Estimation Error}
\vspace{-0.5em}
\label{sec:T1}

We now answer question~\ding{172} and bound the sample complexity of learning a hard-label MR.
The MR wrapper's two-stage pipeline factorizes the sample-wise growth of $\FMR$ into two independent terms, one per wrapper component (Section~\ref{sec:T1-factorization}), and this factorization induces an additive PAC bound on the estimation error of approximate ERM (Section~\ref{sec:T1:bound}).
Proofs are deferred to Appendix~\ref{app:proof_T1}.

\subsection{Reachability before relabeling}
\label{sec:T1-factorization}

On any fixed $x_{1:m} \in \XT^m$, each $f_{g,h} = h \circ \fs \circ g\in \FMR$ first selects a reachable source-label vector $z \in \ReaClass(x_{1:m})$ via $g$ and then relabels $z$ via $h$.
The target-label patterns realized by $\FMR$ on $x_{1:m}$ therefore lie in $\bigcup_{z \in \ReaClass(x_{1:m})} \RelClass(z)$. 
Bounding this union by $|\ReaClass(x_{1:m})| \max_{z \in \ReaClass(x_{1:m})} |\RelClass(z)|$ and taking the supremum over $x_{1:m}$ (\cref{app:proof_T1}) gives:

\begin{proposition}[Sample-wise factorization]
    \label{prop:factorization}
    For every $m \ge 1$, $\Pi_{\FMR}(m) \le \PihatR(m) \cdot \PihatH(m)$.
\end{proposition}
Proposition~\ref{prop:factorization} is the structural reduction behind the analysis.
Rather than simply treating $\FMR$ as an opaque composition, it charges $g$ only for the source-label patterns it can expose on $x_{1:m}$, and $h$ only for relabelings of those exposed patterns.
This enables the sample-complexity rate established below.

\subsection{Agnostic-PAC estimation bound}
\label{sec:T1:bound}
\Cref{prop:factorization} bounds combinatorial realizations on a fixed tuple, leaving open how that product governs the finite-sample estimation error $\EstRisk(\widehat f_n)$ of an $\eta_n$-approximate ERM without assuming target realizability.
As the $0$--$1$ loss indicator $\one\{f(x) \neq y\}$ depends on $f \in \FMR$ solely through $f(x)$, the induced loss class inherits growth at most $\Pi_{\FMR}(2\nT) \le \PihatR(2\nT)\PihatH(2\nT)$.
Applying the two-sided growth-function uniform-deviation bound~\citep[Corollary~3.9 and Eq.~(3.23)]{mohri2018foundations} and telescoping empirical risk against $\RiskClsOptimum$ (\cref{thm:T1-app}) yields an {\em agnostic-PAC} bound (\cref{def:pac}) on the estimation error as follows.

\begin{theorem}[Factorized agnostic-PAC bound]
\label{thm:T1}
For every $\nT \ge 1$, $\delta \in (0,1)$, and $\eta_n \ge 0$, with probability at least $1-\delta$ over an i.i.d.\ sample of size $\nT$ from $\PT$, every measurable $\eta_n$-approximate ERM $\widehat f_n \in \FMR$ satisfies
\begin{equation}
\EstRisk(\widehat f_n) \;\le\; \eta_n + 4\sqrt{6} \, \sqrt{\frac{\log \PihatR(2 \nT) + \log \PihatH(2 \nT) + \log(1/\delta)}{\nT}}.
\label{eq:T1}
\end{equation}
\end{theorem}

\Cref{thm:T1} bounds $\EstRisk(\widehat f_n) = \Risk(\widehat f_n) - \RiskClsOptimum$ for any measurable $\fs$: $\log \PihatR(2\nT)$ measures input-side reachability capacity and $\log \PihatH(2\nT)$ measures output-side relabeling capacity on reached patterns, vanishing whenever $\PihatR(2\nT)=1$ or $\PihatH(2\nT)=1$, respectively.
Inverting \eqref{eq:T1} under polynomial growth via \cref{lem:log-inversion} (\cref{cor:T1-sample-app}) gives an additive sample complexity.

\begin{corollary}[Polynomial-growth sample complexity]
\label{cor:T1-sample}
Suppose there exist constants $d_{\Rreach}, d_{\Hfamily} \ge 0$ and $C_{\Rreach}, C_{\Hfamily} \ge 1$ such that 
\begin{equation*}
    \log \Pi_{\Rreach}(m) \le d_{\Rreach} \log(C_{\Rreach} m) 
\quad\text{and}\quad 
\log \Pi_{\Hfamily \mid \Rreach}(m) \le d_{\Hfamily} \log(C_{\Hfamily} m)
\qquad \text{for every } m \ge 1.
\end{equation*}
Then for every $\varepsilon, \delta \in (0,1)$ and $\eta_n \ge 0$, there exists $n_0 = \widetilde{O}\!\left( (d_{\Rreach} + d_{\Hfamily} + \log(1/\delta)) / \varepsilon^2 \right)$ such that whenever $\nT \ge n_0$, every measurable $\eta_n$-approximate ERM $\widehat f_n \in \FMR$ satisfies $\EstRisk(\widehat f_n) \le \eta_n + \varepsilon$ with probability at least $1 - \delta$ over $\PT^{\nT}$.
\end{corollary}

Thus, if reachable source-label patterns and conditional relabelings grow polynomially, then $\FMR$ is {\em agnostically PAC learnable} by approximate ERM, with sample complexity controlled additively in the two {\em independent} growth exponents: $d_{\Rreach}$ measures how distinct source-label patterns the input transformation can expose through $\fs$ on a target sample, while $d_{\Hfamily}$ measures how richly the output mapping relabels a fixed reachable pattern.
\Cref{sec:T1P} verifies the polynomial-growth condition for additive input transformations~\citep{elsayed2019adversarial, tsai2020transfer, bahng2022exploring, chen2023understanding, tsao2023autovp} under structural conditions on $\fs$, and \cref{sec:T3P} establishes that the additive $d_{\Rreach} + d_{\Hfamily}$ dependence is necessary on at least one explicit family.

\subsection{Why conditioning on reachability matters}
\label{sec:T1:conditional-growth}
\vspace{-0.5em}

We next compare the reachability-aware term $\PihatH$ with the structure-oblivious proxy
$\PihatHall(m)$ from~\cref{sec:prelim:growth}, which counts relabelings on arbitrary length-$m$ source-label vectors, regardless of whether those vectors can be reached by
$\fs\circ\Gfamily$ on a target sample.

\begin{proposition}
\label{prop:no-worse}
For every $m \ge 1$,
$\Pi_{\mathcal{F}_{\mathrm{MR}}}(m)
  \le \Pi_{\mathcal{R}}(m) \cdot \Pi_{\mathcal{H} \mid \mathcal{R}}(m)
  \le \Pi_{\mathcal{R}}(m) \cdot \Pi_{\mathcal{H}}^{\mathrm{all}}(m).$
\end{proposition}
\vspace{-0.5em}

Thus, conditioning on reachability never worsens the growth bound and can be strictly sharper when source-label vectors are unreachable, with concrete consequences for \cref{thm:T1}.

\begin{example}[Strict gap from unreachable source-label patterns]
\label{ex:gap}
Let $\Hfamily = [K_{\rm T}]^{[K_{\rm S}]}$ be the full relabeling class. 
For $z \in [K_{\rm S}]^m$, write $|z|_{\#} := |\{z_1, \ldots, z_m\}|$ for its alphabet size, and define the \emph{reachable alphabet} $\tR(m) := \sup_{x_{1:m}} \sup_{z \in \ReaClass(x_{1:m})} |z|_{\#} \leq \min\{m, K_{\rm S}\}$ in general.
Then,
\begin{equation*}
    \PihatHall(m) = K_{\rm T}^{\min\{m, K_{\rm S}\}}, \qquad \PihatH(m) = K_{\rm T}^{\tR(m)},
\end{equation*}
The gap $\PihatHall / \PihatH$ can be exponential: when $K_{\rm S} \ge 2\nT$ and $\tR(2\nT) = O(1)$, the conditional term contributes only $O(\log K_{\rm T})$ to \eqref{eq:T1}, against $2\nT \log K_{\rm T}$ for the structure-oblivious proxy.

\end{example}


\textbf{From factorized growth to explicit rates.}
\cref{thm:T1} reduces hard-label PAC estimation on $\FMR$ to two distribution-free sample-wise growth functions: $\PihatR$, controlled by the input-transformation class together with $\fs$, and $\PihatH$, controlled by the output-mapping class together with the source labels actually reached on the target sample.
The bound is only as explicit as these two growth terms.
We next turn it into concrete polynomial rates for additive input transformations.
\vspace{-0.5em}
\section{Explicit PAC Rates for Additive Input Transformation}
\vspace{-0.5em}
\label{sec:T1P}

We now instantiate \Cref{thm:T1} on \emph{additive} input transformations~\citep{elsayed2019adversarial, tsai2020transfer, bahng2022exploring, chen2023understanding, tsao2023autovp}: under structural conditions on $\fs$ and $\Hfamily$ (\Cref{sec:T1P:setup}), both growth functions admit explicit polynomial bounds (\Cref{sec:T1P:rate}), verified for affine and fixed-depth ReLU source-score classifiers (\Cref{sec:T1P:discharge}).
\subsection{Additive input transformations and structural conditions}
\label{sec:T1P:setup}

We consider additive input transformations of the form $g_\theta(x) = x + M\theta$, with learnable parameter $\theta \in \Theta \subseteq \mathbb R^p$ and a fixed perturbation matrix $M \in \mathbb R^{d \times p}$ specifying the directions available to the wrapper.
In visual reprogramming~\citep{elsayed2019adversarial, bahng2022exploring}, for example, $M$ is a binary pixel mask that determines where on the target image the perturbation $\theta$ is placed.

\begin{condition}[Reachability arrangement]
\label{cond:arrangement}
There exist constants $c_{\Rreach} \ge 0$ and $C_{\Rreach} \ge 1$ such that, for every integer $m \ge 1$ and every target tuple $x_{1:m} \in \XT^m$, the parameter set $\Theta$ admits a partition into at most $(C_{\Rreach} m)^{c_{\Rreach} p}$ {\em cells} on each of which the source-label vector is constant in $\theta$.
\end{condition}

\begin{condition}[Output-mapping complexity]
\label{cond:nat}
The output-mapping class $\Hfamily \subseteq [K_{\rm T}]^{[K_{\rm S}]}$ has Natarajan dimension~\citep{natarajan1989learning, daniely2015multiclass} $\Ndim(\Hfamily) \le d_{\Hfamily} < \infty$.
\end{condition}

\Cref{cond:arrangement} stipulates that varying $\theta \in \Theta$ induces at most polynomially many source-label patterns on $x_{1:m}$ and concentrates all dependence on $\fs$ (\Cref{sec:T1P:discharge}).
\Cref{cond:nat} controls multiclass head capacity independently of $\fs$.

\subsection{Sample complexity and reachable-label arrangements}
\label{sec:T1P:rate}
Under \Cref{cond:arrangement}, the cell partition directly bounds $|\ReaClass(x_{1:m})| \le (C_{\Rreach} m)^{c_{\Rreach} p}$; under \Cref{cond:nat}, identifying $\RelClass(z)$ with restrictions $\Hfamily|_{\reachset(z)}$ on $|\reachset(z)| \le m$ and applying the multivalued Sauer--Shelah lemma~\citep[Theorem~9]{ben1992characterizations} (\cref{lem:restricted-head}; \cref{thm:T1P-app}) bounds $\PihatH(m)$ polynomially in $m$.

\begin{theorem}[Explicit PAC bound for additive MR]
\label{thm:T1P}
Under \Cref{cond:arrangement,cond:nat} for $\Gfamily_p$, set $d_m := d_{\Hfamily} \wedge m$ (with $B_m := 0$ when $d_m = 0$ or $K_{\rm T} = 1$).
For every $m \ge 1$,
\begin{equation*}
\log \Pi_{\Rreach}(m) \le \underbrace{c_{\Rreach}\, p \log(C_{\Rreach}\, m)}_{=: \,A_m}, \qquad \log \Pi_{\Hfamily \mid \Rreach}(m) \le \underbrace{d_m \log\!\bigl(e \tbinom{K_{\rm T}}{2} m / d_m\bigr)}_{=: \,B_m}.
\end{equation*}
Consequently, for every $\nT \ge 1$, $\delta \in (0,1)$, and $\eta_n \ge 0$, with probability at least $1-\delta$ over $\PT^{\nT}$, every measurable $\eta_n$-approximate ERM $\widehat f_n \in \FMR$ satisfies
\begin{equation}
\EstRisk(\widehat f_n) \;\le\; \eta_n + 4\sqrt{6} \,\cdot \sqrt{\nT^{-1}\, (A_{2\nT} + B_{2\nT} + \log(1/\delta))}.
\label{eq:T1P}
\end{equation}
\end{theorem}

\Cref{thm:T1P} charges the input transformation by $c_{\Rreach} p$ and the head by its $\fs$-free Natarajan dimension $d_{\Hfamily}$ (with $A_{2\nT}=0$ when $p=0$ or $c_{\Rreach}=0$, and $B_{2\nT}=0$ when $d_{\Hfamily}=0$ or $K_{\rm T}=1$).
\begin{corollary}[Polynomial sample complexity for additive MR]
\label{cor:T1P-sample}
Under \Cref{cond:arrangement,cond:nat} for $\Gfamily_p$, for every $\varepsilon, \delta \in (0,1)$ and $\eta_n \ge 0$, there exists 
\begin{equation*}
    n_0 = \widetilde{O}\!\left( \frac{c_{\Rreach}\, p + d_{\Hfamily} + \log(1/\delta)}{\varepsilon^2} \right),    
\end{equation*}
such that whenever $\nT \ge n_0$, every measurable $\eta_n$-approximate ERM $\widehat{f}_n \in \FMR$ satisfies $\EstRisk(\widehat{f}_n) \le \eta_n + \varepsilon$ (and $\Risk(\widehat{f}_n) - \RiskStar \le a_{\mathrm{MR}} + \eta_n + \varepsilon$) with probability at least $1 - \delta$ over $\PT^{\nT}$.
\end{corollary}

Additive hard-label MR is thus {\em agnostically PAC learnable} whenever Conditions~\ref{cond:arrangement}-\ref{cond:nat} hold, with sample complexity additive in $c_{\Rreach}\, p$ and $d_{\Hfamily}$, isolating architectural dependence on $\fs$ inside $(c_{\Rreach}, C_{\Rreach})$.

\subsection{Discharging Condition~\ref{cond:arrangement} through source-score structure}
\label{sec:T1P:discharge}

\Cref{cond:arrangement} is the only $\fs$-side hypothesis in \Cref{thm:T1P}, whereas \Cref{thm:T1} holds for any measurable $\fs$.
Thus, obtaining closed-form rates concentrates architecture dependence into bounding constant-label cells over $\Theta$.
We verify it for two families sharing a structure that $\fs$ is induced by a {\em score map} $s_{\rm S}: \XS \to \mathbb{R}^{K_{\rm S}}$ via $\fs(u) = \arg \max_{k \in [K_{\rm S}]} s_{{\rm S}, k}(u)$~\citep{crammer2001algorithmic}, with ties broken by a fixed measurable rule.
The verification proceeds through weak sign-pattern counting on score-difference functions~\citep{zaslavsky1975facing, warren1968lower, cover2006geometrical} via \cref{lem:weak-sign} bounds cells through $\Phi_p(N) := \sum_{j=0}^{\min\{p, N\}} 2^j \binom{N}{j} \le (1+2N)^p$, which denotes the number of weak sign patterns in $\{-1, 0, +1\}^N$ generated by $N$ affine functions over $\mathbb R^p$, for every $N \ge 1$.

\begin{corollary}[Affine source-score maps]
\label{cor:T1P:affine}
If $s_S(u) = Au + b$ for some $A \in \mathbb R^{K_{\rm S} \times d}$ and $b \in \mathbb R^{K_{\rm S}}$, then for every $m \ge 1$,
\begin{equation*}
\log \PihatR(m) \;\le\; \log \Phi_p\!\bigl(m \binom{K_{\rm S}}{2}\bigr) \;\le\; p \log \bigl(1 + 2 m \binom{K_{\rm S}}{2}\bigr).
\end{equation*}
\end{corollary}
\begin{corollary}[Fixed finite-depth ReLU source-score networks]
\label{cor:T1P:relu}
If $s_S$ is a fixed feedforward ReLU network with $L$ hidden layers of widths $q_1, \ldots, q_L$ (definition in Appendix \ref{app:proof_T1P}), then for every $m \ge 1$,
\begin{equation*}
\log \PihatR(m) \;\le\; \log \Phi_p\!\bigl(m \binom{K_{\rm S}}{2}\bigr) + \sum_{\ell=1}^L \log \Phi_p(m\, q_\ell) \;=\; O\!\bigl(p(L+1) \log m\bigr).
\end{equation*}
\end{corollary}
\begin{remark}
\label{rem:arch-scope}
    \Cref{thm:T1,thm:T1P} separate the architecture-free generalization bound from per-architecture capacity verification (\Cref{cond:arrangement}), mirroring how threshold and piecewise-linear bounds instantiate general VC guarantees~\citep{baum1989what, karpinski1997polynomial, bartlett2019nearly}.
    \Cref{cor:T1P:relu} refines $\Theta$ layer by layer across ReLU pre-activations~\citep{montufar2014number, raghu2017expressive}, yielding $\widetilde{O}((p + d_{\Hfamily})/\varepsilon^2)$ for affine scores and $\widetilde{O}((p(L+1) + d_{\Hfamily})/\varepsilon^2)$ for fixed-depth ReLU scores; extension routes to ConvNets~\citep{he2016deep, bartlett2019nearly, long2020generalization} and Transformers~\citep{vaswani2017attention, trauger2024sequence} are discussed in \Cref{sec:discussion}.
\end{remark}

\section{Matching Lower Bound on an Explicit Binary Construction}
\vspace{-0.5em}
\label{sec:T3P}

\Cref{sec:T1,sec:T1P} upper-bound the sample complexity of hard-label MR by $\widetilde{O}((d_{\Rreach} + d_{\Hfamily})/\varepsilon^2)$.
We now ask whether the additive $(d_{\Rreach} + d_{\Hfamily})$ structure is an artifact of \Cref{prop:factorization} or intrinsic to MR.
We resolve this on an explicit binary family that realizes $\PihatR(m) = 2^{d_{\Rreach}}$ and $\PihatH(m) = 2^{d_{\Hfamily}}$ while keeping the target Bayes classifier inside $\FMR$ (i.e., $a_{\mathrm{MR}} = 0$).
Proofs are deferred to Appendix~\ref{app:proof_T3P}.

\subsection{A binary family separating reachability and relabeling}
\label{sec:T3P:family}

\begin{proposition}[Binary MR family; full construction in \cref{app:proof_T3P:construction}]
\label{con:binary-family}
For all integers $d_{\Rreach}, d_{\Hfamily} \ge 1$ and $\gamma \in (0, 1/2]$, set $m := d_{\Rreach} + d_{\Hfamily}$, $K_{\rm T} := 2$, and $K_{\rm S} := d_{\Hfamily} + 2$.
There exist a canonical tuple $x_{1:m} = (x_1, \ldots, x_m) \in \XT^m$, a fixed $\fs: \XS \to [K_{\rm S}]$, and wrapper classes $\Gfamily = \{g_\alpha : \alpha \in \{0,1\}^{d_{\Rreach}}\}$ ($|\Gfamily| = 2^{d_{\Rreach}}$) and $\Hfamily = \{h_\beta : \beta \in \{0,1\}^{d_{\Hfamily}}\}$ ($|\Hfamily| = 2^{d_{\Hfamily}}$) such that $\FMR = \{h_\beta \circ \fs \circ g_\alpha\}$ realizes every $\omega = (\alpha, \beta) \in \{0,1\}^m$ on $x_{1:m}$, with $\PihatR(m') \le 2^{d_{\Rreach}}$ and $\PihatH(m') \le 2^{d_{\Hfamily}}$ for all $m' \ge 1$ and equality at $m' = m$.
Let $P_{\rm T}^{\omega}$ be uniform on $\{x_1, \ldots, x_m\}$ with $P_{\rm T}^{\omega}(Y = \omega_i \mid X = x_i) = (1 + \gamma)/2$ for $i \in [m]$, and define $\mathfrak{F}_{d_{\Rreach},d_{\Hfamily}}^{\mathrm{bin}}(\gamma) := \{P_{\rm T}^{\omega} : \omega \in \{0,1\}^m\}$.
\end{proposition}
\Cref{con:binary-family} realizes $d_{\Rreach}$ and $d_{\Hfamily}$ as exact reachability and conditional-relabeling log-growths on $x_{1:m}$, while placing the Bayes rule $x_i \mapsto \omega_i$ of each $P_{\rm T}^{\omega}$ inside $\FMR$, so $a_{\mathrm{MR}} = 0$ and $\Risk(\widehat{f}_n) - \RiskStar = \EstRisk(\widehat{f}_n)$ across $\mathfrak{F}^{\mathrm{bin}}_{d_{\Rreach}, d_{\Hfamily}}(\gamma)$.

\subsection{Lower bound and matching consequence}
\label{sec:T3P:bound}
With approximation error eliminated on $\mathfrak{F}^{\mathrm{bin}}_{d_{\Rreach}, d_{\Hfamily}}(\gamma)$, we ask whether any learning rule, not merely approximate ERM, can achieve smaller estimation error than \Cref{thm:T1}.
The lower bound on $\mathfrak F^{\rm bin}_{d_{\Rreach}, d_{\Hfamily}}(\gamma)$ follows from Assouad's lemma~\citep{tsybakov2009introduction, yu1997assouad} applied to the binary classification problem induced by \cref{con:binary-family}, with margin $\gamma \asymp \sqrt{(d_{\Rreach} + d_{\Hfamily})/\nT}$.

\begin{theorem}[Matching minimax lower bound]
\label{thm:T3P}
With $c_0 = 1$, $c_1 = 1/4$, and $c_2 = e^{-1/4}/16$, for all integers $d_{\Rreach}, d_{\Hfamily} \ge 1$ and $\nT \ge 1$ satisfying $d_{\Rreach} + d_{\Hfamily} \le c_0 \nT$, on $\mathfrak{F}_{d_{\Rreach},d_{\Hfamily}}^{\mathrm{bin}}(\gamma)$ with margin $\gamma = c_1 \sqrt{(d_{\Rreach} + d_{\Hfamily})/\nT}$, every learning rule $\widehat{f}_n$ mapping $\nT$ i.i.d.\ target samples to a measurable predictor satisfies $\sup_{\PT \in \mathfrak{F}_{d_{\Rreach},d_{\Hfamily}}^{\mathrm{bin}}(\gamma)} \E_{\PT^{\nT}}[\EstRisk(\widehat{f}_n)] \ge c_2 \sqrt{(d_{\Rreach} + d_{\Hfamily})/\nT}$.
\end{theorem}

\Cref{thm:T3P} matches \Cref{thm:T1} on $\mathfrak{F}^{\mathrm{bin}}_{d_{\Rreach}, d_{\Hfamily}}(\gamma)$, where substituting $\log \PihatR(2\nT) \le d_{\Rreach} \log 2$ and $\log \PihatH(2\nT) \le d_{\Hfamily} \log 2$ into \eqref{eq:T1} gives $\EstRisk(\widehat f_n) \le \eta_n + 4\sqrt{6}\, \sqrt{\nT^{-1}((d_{\Rreach} + d_{\Hfamily})\log 2 + \log(1/\delta))}$.
Operationally, $d_{\Rreach}$ independent bits must be learned through reachability and $d_{\Hfamily}$ bits through conditional relabeling, with neither component substituting for the other (when $d_{\Rreach} \gg d_{\Hfamily}$ or $d_{\Hfamily} \gg d_{\Rreach}$, the larger capacity alone governs the minimax rate); note that \Cref{thm:T3P} establishes tightness on this explicit binary family only.

\begin{corollary}[Sample-complexity lower bound]
\label{cor:T3P-sample}
For all $d_{\Rreach}, d_{\Hfamily} \ge 1$, every $\varepsilon \in (0, c_2/2)$, any learner $\widehat{f}_n$ achieving $\sup_{\PT \in \mathfrak{F}^{\mathrm{bin}}_{d_{\Rreach}, d_{\Hfamily}}(\gamma)} \mathbb{E}_{\PT^{\nT}}[\EstRisk(\widehat{f}_n)] \le \varepsilon$ at margin $\gamma = \min\bigl\{1/4,\, c_1\sqrt{(d_{\Rreach} + d_{\Hfamily})/\nT}\bigr\}$ requires $\nT \ge c_2^2(d_{\Rreach} + d_{\Hfamily})/\varepsilon^2 = \Omega\bigl((d_{\Rreach} + d_{\Hfamily})/\varepsilon^2\bigr)$ target samples.
\end{corollary}

\Cref{cor:T3P-sample} matches \cref{cor:T1-sample,cor:T1P-sample} up to constants and shows that the two growth sources contribute additively as they are \emph{independently identifiable} on $\mathfrak F^{\rm bin}_{d_{\Rreach}, d_{\Hfamily}}(\gamma)$: $d_{\Rreach}$ bits are learned through reachability and $d_{\Hfamily}$ bits through conditional relabeling, with neither side substituting for the other.

Sections~\ref{sec:T1}--\ref{sec:T3P} answer question~\ding{172} with matching $\widetilde{O}((d_{\Rreach} + d_{\Hfamily})/\varepsilon^2)$ and $\Omega((d_{\Rreach} + d_{\Hfamily})/\varepsilon^2)$ rates.
We next address question~\ding{173}: when the estimation error vanishes, the wrapper class may still be separated from the target Bayes rule by the population residual $a_{\mathrm{MR}}$, which we now characterize exactly.

\section{Approximation Error: Identity and Certificate}
\label{sec:T2}
We have bounded the estimation error of hard-label MR, but left open whether $\FMR$ contains a predictor close to the target Bayes optimum.
Through the wrapper-side viewpoint, we show that the approximation error $a_{\mathrm{MR}}=\RiskClsOptimum-\RiskStar$ is still governed by the two-stage pipeline of MR, which asks whether the reachable source labels through $g \in \Gfamily$ preserve enough target-label information, and whether the allowed output mappings $h \in \Hfamily$ can use that information.
See Appendix~\ref{app:proof-identities}-\ref{app:proof_T2} for proofs.

\subsection{Exact population decomposition}
\label{sec:T2:identity}
To isolate what prevents $\RiskClsOptimum$ from reaching $\RiskStar$ even with infinite target data, fix $g \in \Gfamily$ and write $Z_g := \fs(g(X)) \in [K_{\rm S}]$ on $(X, Y) \sim \PT$.
Given $Z_g$, the best output mapping in $\Hfamily$ achieves $R_{\Hfamily}^{\ast}(g) := \inf_{h\in\Hfamily} \E_{\PT}[\one\{h(Z_g)\neq Y\}]$, whereas optimizing over {\em all} unrestricted relabelings $h: [K_{\rm S}] \to [K_{\rm T}]$ achieves the channel Bayes risk $R_{\mathrm{all}}^{\ast}(g) := \inf_{h:[K_{\rm S}]\to[K_{\rm T}]} \E_{\PT}[\one\{h(Z_g)\neq Y\}] = \E_{\PT}[1-\max_{y\in[K_{\rm T}]}\PT(Y=y \mid Z_g)]$.
Writing $\RiskClsOptimum = \inf_{g \in \Gfamily} R_{\Hfamily}^{\ast}(g)$, we have:
\begin{proposition}[Exact decomposition of $a_{\rm MR}$]
\label{prop:T2P}
For every $\PT$, fixed measurable $\fs$, nonempty measurable $\Gfamily$, and nonempty $\Hfamily \subseteq [K_{\rm T}]^{[K_{\rm S}]}$,
\begin{equation}
    a_{\mathrm{MR}}
    \;=\;
    \inf_{g\in\Gfamily}
    \Bigl\{
        \underbrace{R_{\mathrm{all}}^{\ast}(g)-\RiskStar}_{\textup{source-label information loss}}
        \;+\;
        \underbrace{R_{\Hfamily}^{\ast}(g)-R_{\mathrm{all}}^{\ast}(g)}_{\textup{target-relabeling restriction}}
    \Bigr\},
    \label{eq:T2P}
\end{equation}
where both underbraced summands are nonnegative for every $g \in \Gfamily$.
\end{proposition}

\Cref{prop:T2P} attributes $a_{\mathrm{MR}}$ to two coupled wrapper bottlenecks evaluated at a shared $g$: {\em source-label information loss} arises whenever distinct target Bayes-decision regions collapse onto a single source label of $Z_g$ so no relabeling can separate them, whereas {\em target-relabeling restriction} measures the cost of restricting the head to $\Hfamily$.
In the extreme cases, information loss vanishes iff $R_{\mathrm{all}}^{\ast}(g) = \RiskStar$, and relabeling restriction vanishes iff $\Hfamily$ contains a channel-Bayes relabeling $h_g^{\star}$ satisfying $h_g^{\star}(k) \in \arg\max_{y \in [K_{\rm T}]} \PT(Y=y \mid Z_g = k)$ on the support of $Z_g$ (identically zero when $\Hfamily = [K_{\rm T}]^{[K_{\rm S}]}$).
Note that both losses share $g$ inside the infimum, minimizing one need not minimize the other.

\subsection{Source-conditional upper bound}
\label{sec:T2:certificate}

Although \Cref{prop:T2P} is exact, evaluating $R_{\mathrm{all}}^{\ast}(g)$ and $\RiskStar$ requires unknown target conditional posteriors.
We replace these with three source-conditional quantities computable from the source and target distributions, yielding an empirically tractable upper bound.
For a source distribution $\PS$ over $\XS \times [K_{\rm S}]$ with source-input marginal $P_{{\rm S},X}$, target-input marginal $P_{{\rm T},X}$, and measurable Bayes selectors $b_{\rm T}:\XT\to[K_{\rm T}]$ and $b_{\rm S}:\XS\to[K_{\rm S}]$, for $(g, h) \in \Gfamily \times \Hfamily$, define the target-side Bayes-mismatch term $\lamhard(g, h)$ and source-side relabeled-classifier error $\epshard(h)$ as
\begin{equation*}
    \lamhard(g, h) := \Prob_{X \sim \PT}\!\bigl(b_{\rm T}(X) \ne h(b_{\rm S}(g(X)))\bigr),
    \;
    \epshard(h) := \Prob_{U \sim P_{S, X}}\!\bigl(h(\fs(U)) \ne h(b_{\rm S}(U))\bigr).
\end{equation*}
With $\phi_h(u) = \one\{h(\fs(u))\neq h(b_{\rm S}(u))\}$, define the head-dependent pushforward discrepancy~\citep{mansour2009domain}
\begin{equation*}
    \Deltahard(g, h) := \bigl| \mathbb E_{X \sim \PT}\, \phi_h(g(X)) - \mathbb E_{U \sim P_{S, X}}\, \phi_h(U) \bigr|.
\end{equation*}


\begin{theorem}[Population upper certificate for $a_{\rm MR}$]
\label{thm:T2C}
Suppose $\PT$ and $\PS$ admit measurable Bayes-optimal classifiers $b_{\rm T}: \XT \to [K_{\rm T}]$ and $b_{\rm S}: \XS \to [K_{\rm S}]$.
Then
\begin{equation}
    a_{\mathrm{MR}}
    \;\le\;
    \inf_{(g,h)\in\Gfamily\times\Hfamily}
    \Bigl\{
        \underbrace{\lamhard(g, h)}_{\textup{mismatch}}
        \;+\;
        \underbrace{\epshard(h)}_{\textup{source error}}
        \;+\;
        \underbrace{\Deltahard(g, h)}_{\textup{discrepancy}}
    \Bigr\}.
    \label{eq:T2C}
\end{equation}
\end{theorem}
\Cref{thm:T2C} upper-bounds $a_{\mathrm{MR}}$ through three operational terms, where $\lamhard(g, h)$ measures whether the relabeled source Bayes rule $h \circ b_{\rm S} \circ g$ agrees with $b_{\rm T}$ on $P_{{\rm T},X}$, $\epshard(h)$ measures the $h$-relabeled error of $\fs$ against $b_{\rm S}$ on $P_{{\rm S},X}$, and $\Deltahard(g, h)$ measures the shift of $\phi_h$ between $g_\# P_{{\rm T},X}$ and $P_{{\rm S},X}$.
When $\fs = b_{\rm S}$ pointwise on $\XS$, $\epshard(h) = \Deltahard(g, h) = 0$ and \eqref{eq:T2C} reduces to $\inf_{g,h}\lamhard(g,h)$.
Plugging consistent estimators of $b_{\rm S}$ and $b_{\rm T}$ into \eqref{eq:T2C} yields i.i.d.-estimable proxies inheriting their rates~\citep{mohri2018foundations}.

\begin{remark}
\label{rem:one-way-certificate}
\Cref{prop:T2P} is an exact, slack-free identity, whereas \Cref{thm:T2C} is a one-directional upper certificate (smallness certifies small $a_{\rm MR}$, while large values are inconclusive).
\cref{app:proof_T2} constructs a converse: (i) bounding $\Risk(f_{g,h}) - \RiskStar$ by $\Prob_{P_{{\rm T},X}}(f_{g,h}(X) \neq b_{\rm T}(X))$ is loose wherever the target posterior is near-uniform without an unassumed low-noise margin condition on $\PT$; and (ii) once $b_{\rm S}$ is fixed or estimated, $\Deltahard(g,h)$ is estimable from unlabeled inputs because $\phi_h$ is evaluated pointwise at the tested $h$ rather than taking a supremum over $\Hfamily\Delta\Hfamily$~\citep{ben2010theory}.
As such, \Cref{thm:T2C} is a wrapper-side upper bound on $a_{\rm MR}$, not an $\mathcal{H}\Delta\mathcal{H}$-style class-symmetric transfer divergence~\citep{ben2010theory}.
\end{remark}

\textbf{Extension to logit-exposing MR.}
The approximation analysis also transfers to source models exposing real-valued logit vectors $U_g = s_{\rm S}(g(X)) \in \mathbb{R}^{K_{\rm S}}$, with discrete relabeling replaced by a learnable linear map $W \in \Wclass \subseteq \mathbb{R}^{K_{\rm T} \times K_{\rm S}}$ (identifying each matrix $W$ with $u \mapsto Wu$).
\Cref{app:scoreoutput} establishes exact population identities (\cref{prop:score-01-identity,prop:score-surrogate-identity}), aand a population certificate (\cref{thm:score-certificate}), which is empirically estimable over a data-independent candidate set of MR wrappers (\cref{thm:score-rademacher}).
\section{Controlled Empirical Verification}
\label{sec:empirical}
\begin{table}[t]
\centering
\caption{Empirical verification under exact ERM. 
}
\label{tab:empirical-main}
\scriptsize
\begin{minipage}[t]{0.475\textwidth}
\centering
\setlength{\tabcolsep}{3.0pt}
\begin{tabular*}{\textwidth}{@{\extracolsep{\fill}}lccccc@{}}
\toprule
\multicolumn{6}{@{}c@{}}{\textbf{(a) Synthetic Estimation $\EstRisk(\widehat f_n)\downarrow$ (Input $\equiv$ Head)}} \\
\cmidrule{1-6}
Sample size & $r=0$ & $r=2$ & $r=4$ & $r=6$ & $r=8$ \\
\midrule
$\nT = 8$   & $0.0000$ & $0.0475$ & $0.0995$ & $0.1340$ & $0.1765$ \\
$\nT = 32$  & $0.0000$ & $0.0080$ & $0.0155$ & $0.0255$ & $0.0295$ \\
$\nT = 64$  & $0.0000$ & $0.0020$ & $0.0035$ & $0.0045$ & $0.0050$ \\
$\nT = 512$ & $0.0000$ & $0.0000$ & $0.0000$ & $0.0000$ & $0.0000$ \\
\bottomrule
\end{tabular*}
\end{minipage}%
\hfill
\begin{minipage}[t]{0.505\textwidth}
\centering
\setlength{\tabcolsep}{2.2pt}
\begin{tabular*}{\textwidth}{@{\extracolsep{\fill}}lccccc@{}}
\toprule
\multicolumn{6}{@{}c@{}}{\textbf{(b) Synthetic Approximation Split at Matched $\RiskClsOptimum$}} \\
\cmidrule{1-6}
Mechanism ($m$) & $\RiskClsOptimum$ & $I$ & $H$ & $\widehat I$ & $\widehat H$ \\
\midrule
Merge labels ($2$)  & $0.2000$ & $0.1000$ & $0.0000$ & $0.0979$ & $0.0000$ \\
Restrict head ($2$) & $0.2000$ & $0.0000$ & $0.1000$ & $0.0000$ & $0.1000$ \\
Merge labels ($8$)  & $0.5000$ & $0.4000$ & $0.0000$ & $0.3923$ & $0.0000$ \\
Restrict head ($8$) & $0.5000$ & $0.0000$ & $0.4000$ & $0.0000$ & $0.4000$ \\
\bottomrule
\end{tabular*}
\end{minipage}

\vspace{0.35em}
\setlength{\tabcolsep}{1.6pt}
\begin{tabular*}{\textwidth}{@{\extracolsep{\fill}}llcccccccc@{}}
\toprule
\multicolumn{10}{@{}c@{}}{\textbf{(c) Pretrained Neural Sources on $90^\circ$-Rotated MNIST Parity (Held-Out $0$--$1$ Risk $\downarrow$)}} \\
\cmidrule{1-10}
 & & \multicolumn{4}{c}{\textbf{Frozen MLP Source}} & \multicolumn{4}{c@{}}{\textbf{Frozen Minimal-LeNet Source}} \\
\cmidrule(lr){3-6} \cmidrule(l){7-10}
Wrapper $(\Gfamily, \Hfamily)$ & Regime & $\nT{=}50$ & $200$ & $800$ & $3200$ & $\nT{=}50$ & $200$ & $800$ & $3200$ \\
\midrule
$(\Gfamily_{\mathrm{id}}, \Hfamily_{\mathrm{thr}})$   & Fixed, threshold     & $0.4701$ & $0.4515$ & $0.4439$ & $0.4385$ & $0.4206$ & $0.4029$ & $0.3979$ & $0.3960$ \\
$(\Gfamily_{\mathrm{id}}, \Hfamily_{\mathrm{full}})$  & Fixed, full binary   & $0.4416$ & $0.4069$ & $0.4006$ & $0.3980$ & $0.4006$ & $0.3634$ & $0.3552$ & $0.3538$ \\
$(\Gfamily_{\mathrm{rot}}, \Hfamily_{\mathrm{thr}})$  & $4$-rot., threshold  & $0.3967$ & $0.3718$ & $0.3636$ & $0.3613$ & $0.3621$ & $0.3460$ & $0.3374$ & $0.3352$ \\
$(\Gfamily_{\mathrm{rot}}, \Hfamily_{\mathrm{full}})$ & $4$-rot., full binary & $\mathbf{0.0098}$ & $\mathbf{0.0098}$ & $\mathbf{0.0098}$ & $\mathbf{0.0098}$ & $\mathbf{0.0058}$ & $\mathbf{0.0058}$ & $\mathbf{0.0058}$ & $\mathbf{0.0058}$ \\
\midrule
Untrained $(\Gfamily_{\mathrm{rot}}, \Hfamily_{\mathrm{full}})$ & Random-weight source & \multicolumn{3}{c}{--} & $0.4355$ & \multicolumn{3}{c}{--} & $0.5075$ \\
\bottomrule
\end{tabular*}
\vspace{-0.6em}
\end{table}

\Cref{thm:T1} predicts that finite-sample estimation error responds symmetrically to input-side reachability and output-side relabeling capacity, while \Cref{prop:T2P} attributes population approximation error to lost source-label distinctions or head restriction.
To isolate wrapper structure from optimizer dynamics, we evaluate both predictions under exact ERM ($\eta_n = 0$) on a synthetic benchmark and on frozen MNIST classifiers (\Cref{app:empirical}; code at \url{https://github.com/zyecs/repgm-slt}).

\textbf{Capacity symmetry and separable approximation bottlenecks.}
Consider a uniform binary task over $16$ basis vectors in $\mathbb{R}^{16}$, grouped into $8$ ordered pairs whose clean labels $(0, 1)$ flip independently with probability $0.1$ ($\RiskStar = 0.1$) under a bijective source $\fs(e_i) = i$ (\cref{app:empirical:synthetic}).
On the first $r \in \{0, 2, 4, 6, 8\}$ pairs, independently allowing $g$ to swap each pair's vectors under a fixed parity head ($\log_2 \PihatR = r$, $\log_2 \PihatH = 0$) realizes the exact same $2^r$ predictors ($a_{\rm MR} = 0$) as fixing $g = \mathrm{id}$ and independently flipping the head's pair assignments ($\log_2 \PihatR = 0$, $\log_2 \PihatH = r$).
Both parameterizations yield identical $\EstRisk(\widehat f_n)$ across all $(r, \nT)$ (\Cref{tab:empirical-main}a), rising at $\nT = 8$ from $0.0475$ ($r = 2$) to $0.1765$ ($r = 8$) before vanishing by $\nT = 512$.
Across $m \in \{0, 2, 4, 6, 8\}$ affected pairs, two wrappers also share the exact suboptimal risk $\RiskClsOptimum = 0.1 + 0.05m$ via opposite terms of \Cref{prop:T2P} (\Cref{tab:empirical-main}b): merging each affected pair onto one source label under an unrestricted head pools opposite Bayes labels into information loss ($I = 0.05m$, $H = 0$), whereas keeping all $16$ source labels distinct while fixing the wrong label $1$ on each pair's first vector places the same $0.05m$ residual into head restriction ($I = 0$, $H = 0.05m$), tracked by plug-ins $(\widehat I, \widehat H)$.

\textbf{Joint input and output adaptation on pretrained neural sources.}
Reprogramming frozen upright-MNIST~\citep{lecun1998gradient} classifiers (MLP and minimal LeNet exposing only predicted digit IDs in $\{0, \ldots, 9\}$) for binary parity on unseen $90^\circ$-rotated digits illustrates how estimation decay and approximation floors interact across $\nT \in \{50, 200, 800, 3200\}$ (\cref{app:empirical:mnist}; \Cref{tab:empirical-main}c).
Crossing identity access $\Gfamily_{\mathrm{id}}$ versus four rotations $\Gfamily_{\mathrm{rot}} = \{\mathrm{rot}_0, \mathrm{rot}_{90}, \mathrm{rot}_{180}, \mathrm{rot}_{270}\}$ with single-threshold heads $\Hfamily_{\mathrm{thr}}$ versus full binary relabelings $\Hfamily_{\mathrm{full}} = \{0, 1\}^{10}$, only $(\Gfamily_{\mathrm{rot}}, \Hfamily_{\mathrm{full}})$ reaches sub-$1\%$ error, locking onto $\mathrm{rot}_{270}$ already at $\nT = 50$.
When either stage is restricted, growing $\nT$ from $50$ to $3200$ reduces estimation error ($0.4416 \to 0.3980$ under $(\Gfamily_{\mathrm{id}}, \Hfamily_{\mathrm{full}})$ and $0.3967 \to 0.3613$ under $(\Gfamily_{\mathrm{rot}}, \Hfamily_{\mathrm{thr}})$ on MLP) before stalling at high approximation floors from unaligned digits or non-alternating heads.
Applying $(\Gfamily_{\mathrm{rot}}, \Hfamily_{\mathrm{full}})$ to untrained networks stays near chance, confirming that wrapper capacity helps only when the frozen source already separates task-relevant structure.

\section{On Understanding the Practice of Model Reprogramming}
\vspace{-0.5em}
\label{sec:discussion}

Our {\em wrapper-side} analysis turns the MR wrapper into a diagnostic tool: \Cref{thm:T1,prop:T2P} attribute finite-sample estimation cost and the approximation residual to specific wrapper stages.

\textbf{Two failure modes of MR.}
\Cref{prop:T2P} splits $a_{\rm MR}$ into source-label information loss $R_{\rm all}^*(g) - R_{\rm T}^*$, governed by $Z_g = \fs(g(X))$, and target-relabeling restriction $R_{\Hfamily}^*(g) - R_{\rm all}^*(g)$, governed by $\Hfamily$.
Notably, \Cref{prop:T2P} depends on whether the partition induced by $Z_g$ preserves target Bayes decisions rather than whether $g_\# P_{{\rm T},X}$ matches $P_{{\rm S},X}$ in input space, whereas input alignment enters \Cref{thm:T2C} via $\Deltahard(g, h)$ only as a sufficient condition for small $a_{\rm MR}$.
Because the two terms require distinct remedies (modifying $g$ vs.\ enlarging $\Hfamily$), separating them is key to diagnosing failures.

\textbf{Input-side trade-off.}
\Cref{thm:T1P} and \Cref{prop:T2P} expose a wrapper-side bias--variance trade-off~\citep{geman1992neural, hastie01statisticallearning}: enlarging $\Gfamily_p$ via $p$ can make $Z_g$ more informative about $Y$ and reduce $\inf_g R_{\rm all}^*(g)$, at the cost of sample complexity $\widetilde O((c_{\Rreach} p + d_{\Hfamily})/\varepsilon^2)$.
Recent input-side designs sit at complementary points along this curve: \citet{cai2024sample} {\em raised} $p$ with a small ConvNet generating input-dependent masks to cut approximation loss, whereas \citet{jin2025lor} {\em lowered} $p$ via low-rank sub-region matrices to curb the statistical exponent without empirically sacrificing approximation power.

\textbf{Output-side implications.}
Our results distinguish two subsets of $[K_{\rm S}]$ on the output side.
In finite samples, \Cref{thm:T1} charges $\Hfamily$ via $\PihatH$, scaling with the sample-wise reachable alphabet size $\tR(m) \le K_{\rm S}$ rather than $K_{\rm S}$ (e.g., $K_{\rm S}=1000$ on ImageNet~\citep{deng2009imagenet} versus $K_{\rm T}=10$ on CIFAR-10).
At the population level, \Cref{prop:T2P} defines the optimal head given $g$ as a channel-Bayes relabeling on the active support $\mathcal Z_g := \{k \in [K_{\rm S}] : \Prob_{\PT}(Z_g = k) > 0\}$.
This shifts head design from picking a one-to-one injection $[K_{\rm S}] \to [K_{\rm T}]$~\citep{chen2023understanding} to ensuring expressivity on $\mathcal Z_g$: when every Bayes relabeling on $\mathcal Z_g$ is many-to-one, injections incur positive $R_{\Hfamily}^*(g) - R_{\rm all}^*(g)$, whereas many-to-many or Bayesian-guided mappings~\citep{cai2024bayesian} close this gap at cost governed only by $\tR$ (\Cref{ex:gap}).

\textbf{Limitations and outlook.}
Three boundaries of our analysis point to natural extensions.
First, while \Cref{thm:T1} holds for any measurable $\fs$, the closed-form rate in \Cref{thm:T1P} relies on the polynomial cell-partition condition
(\Cref{cond:arrangement}), which we verified for affine score maps and finite-depth ReLU networks (\Cref{cor:T1P:affine,cor:T1P:relu}).
Extending explicit rates to broader architectures depends on how $\fs$ transforms wrapper parameters. 
In frozen piecewise-affine ConvNets~\citep{he2016deep},
convolutional, ReLU, residual, and pooling layers keep pre-activations piecewise affine in $\theta$, motivating computation-graph cell bounds, whereas smooth
operations in Transformers~\citep{vaswani2017attention} (e.g., softmax attention, LayerNorm, and GELU) break piecewise affinity and instead call for margin-based
covering-number or Rademacher bounds~\citep{bartlett2019nearly, long2020generalization, trauger2024sequence}.
Second, the growth factorization in \Cref{prop:factorization} relies on $\fs$ taking values in a finite alphabet $[K_{\rm S}]$.
When the source model instead exposes continuous logits in $\mathbb{R}^{K_{\rm S}}$, our approximation identities and held-out certificates still hold
(\Cref{app:scoreoutput}), but the PAC bound of \Cref{thm:T1} does not extend automatically; proving same-sample ERM guarantees there calls for combining capacity
bounds on $\{x \mapsto s_{\rm S}(g(x)) : g \in \Gfamily\}$ and norm-constrained heads~\citep{mohri2018foundations, bartlett2002rademacher} with vector
contraction~\citep{maurer2016vector} and calibrated surrogate transfer~\citep{zhang2004statistical, tewari2007consistency}.
Third, \Cref{thm:T2C} gives a one-directional upper bound for a fixed pair $(g, h)$ (\Cref{rem:one-way-certificate}), whereas practical methods update $(g_t, h_t)$
iteratively.
Analyzing how training dynamics reduce the two terms of \Cref{prop:T2P}, and combining our wrapper-side bounds with source-centered reprogrammability
conditions~\citep{yang2021voice2series, englert2022adversarial, chung2025model}, would link wrapper design to both optimization and source selection.

\textbf{Summary.}
We developed a finite-sample analysis of hard-label MR: the estimation analysis factorizes $\FMR$'s sample-wise complexity into reachability and conditional-relabeling growth (\Cref{thm:T1}), giving closed-form PAC rates on additive transformations (\Cref{thm:T1P}) matched by a minimax lower bound (\Cref{thm:T3P}), while the approximation analysis splits the residual into source-label information loss and output-mapping restriction (\Cref{prop:T2P,thm:T2C}), with both mechanisms verified in synthetic and pretrained-MNIST studies (\Cref{sec:empirical}).
Our analysis complements source-centered MR theory and offers a wrapper-side viewpoint to guide the design of MR wrappers for a given frozen $\fs$.
\clearpage
\begin{ack}
The authors thank the anonymous reviewers and the Area Chair for their constructive feedback and insightful suggestions during the review process.
This research was partly supported by the University of Melbourne’s Research Computing Services and the Petascale Campus Initiative. Feng Liu is supported by the ARC with grant numbers DP230101540, DE240101089 and DP260103996.
Pin-Yu Chen is solely funded by IBM Research.
\end{ack}
\bibliography{references}
\bibliographystyle{IEEEtranN}

\newpage
\appendix
\crefalias{section}{appendix}
\crefalias{subsection}{appendix}

\listofappendices
\newpage

\appsection{Notations and Assumptions}
\input{appendix/assumptions}
\newpage

\appsection{Approximation Analysis on Logit-Exposing Source Models}
\label{app:scoreoutput}
\input{appendix/app_approximation_error}
\newpage

\appsection{Proof of the Factorized Excess-Risk Bound and Comparison Chain}
\input{appendix/proof_T1}
\newpage

\appsection{Proofs for the Additive Specialization and Representation-Side Discharges}
\input{appendix/proof_T1P_C1prime}
\newpage

\appsection{Proof of the Lower-Bound Result}
\input{appendix/proof_T3P}
\newpage

\appsection{Proofs for Exact Representation-Conditioned Approximation Identities}
\input{appendix/proof_identities}
\newpage

\appsection{Proof of the Hard-Label Source-Conditional Upper Certificate}
\input{appendix/proof_T2}
\newpage

\appsection{Proofs for the Score-Output Certificates and Empirical Corollary}
\input{appendix/proof_T2P_C2prime}
\newpage

\appsection{Uniform Empirical Score-Output Certificate over a Fixed Candidate Class}
\input{appendix/rademacher_certificate}
\newpage

\appsection{Worked Examples}
\input{appendix/worked_examples.tex}
\newpage

\appsection{Controlled Empirical Verification: Protocols and Complete Results}
\input{appendix/empirical_verification}
\newpage
\input{checklist}

\end{document}

%% file: figures/frozen_access_funnel.tex
\begin{tikzpicture}[
  x=1cm,y=1cm,
  >={Latex[length=1.55mm,width=1.05mm]},
  font=\scriptsize,
  box/.style={
    draw,
    rounded corners=2pt,
    align=center,
    inner sep=2.2pt,
    minimum height=0.44cm
  },
  io/.style={
    box,
    fill=black!3,
    minimum width=0.62cm
  },
  learned/.style={
    box,
    fill=red!6,
    draw=red!45,
    minimum width=1.05cm
  },
  frozen/.style={
    box,
    fill=gray!13,
    pattern=north east lines,
    pattern color=gray!45,
    minimum width=1.05cm
  },
  source/.style={
    box,
    fill=blue!5,
    text width=1.85cm
  },
  risk/.style={
    box,
    fill=black!2,
    draw=black!25,
    text width=5.95cm,
    minimum height=0.42cm,
    font=\scriptsize
  },
  panel/.style={
    draw=black!16,
    fill=black!1,
    rounded corners=4pt
  },
  coltitle/.style={
    font=\tiny\scshape,
    text=black!60,
    align=center
  },
  result/.style={
    box,
    font=\scriptsize,
    text width=2.75cm,
    minimum height=0.56cm,
    inner sep=2pt
  },
  est/.style={
    result,
    fill=blue!4,
    draw=blue!42
  },
  app/.style={
    result,
    fill=orange!5,
    draw=orange!58
  },
  legendbox/.style={
    draw,
    rounded corners=1.3pt,
    minimum width=0.30cm,
    minimum height=0.13cm,
    inner sep=0pt
  },
  legend/.style={
    font=\tiny,
    text=black!70,
    align=left
  },
  line/.style={->,thick}
]

\node[io]      (x)    at (0.55,0) {{\tiny\textbf{TARGET INPUT}}\\[-2pt]$X$};
\node[learned] (g)    at (2.40,0) {{\tiny\textbf{LEARNABLE}}\\[-2pt]$g\in\mathcal G$};
\node[frozen]  (fs)   at (4.00,0) {\textbf{FROZEN}\\[-2pt]$\fs$};
\node[source]  (r)    at (5.83,0) {\tiny \textbf{SOURCE LABEL}\\[-2pt]$\fs(g(\mathbf{X}))$};
\node[learned] (h)    at (7.84,0) {{\tiny\textbf{LEARNABLE}}\\[-2pt]$h\in\mathcal H$};
\node[io]      (yhat) at (9.80,0) {{\tiny\textbf{TARGET OUTPUT}}\\[-2pt]$Y$};

\draw[line] (x) -- (g);
\draw[line] (g) -- (fs);
\draw[line] (fs) -- (r);
\draw[line] (r) -- (h);
\draw[line] (h) -- (yhat);

\node[risk] (risk) at (5.05,-0.60)
  {$\Risk(\widehat f_n) \, - \, \RiskStar
    \; = \; \EstRisk(\widehat f_n) \,+ \, a_{\mathrm{MR}}$};

\path[panel]
  (0.10,-2.95) rectangle (10.00,-1.05);

\node[coltitle] at (1.70,-1.27) {GENERAL};
\node[coltitle] at (5.05,-1.27) {QUANTIFIED};
\node[coltitle] at (8.40,-1.27) {SHARPNESS / EXTENSION};

\node[est] (pac) at (1.70,-1.72)
  {PAC estimation\\[-1pt]
   {Theorem~\ref{thm:T1}}};

\node[est] (rates) at (5.05,-1.72)
  {Closed-form rates\\[-1pt]
   {Theorem~\ref{thm:T1P}}};

\node[est] (lb) at (8.40,-1.72)
  {Minimax lower bound\\[-1pt]
   {Theorem~\ref{thm:T3P}}};

\node[app] (id) at (1.70,-2.32)
  {Population identity\\[-1pt]
   {Proposition~\ref{prop:T2P}}};

\node[app] (cert) at (5.05,-2.32)
  {Upper certificate\\[-1pt]
   {Theorem~\ref{thm:T2C}}};

\node[app] (score) at (8.40,-2.32)
  {Logit-exposing $f_{\rm S}$\\[-1pt]
   {Theorems.~\ref{thm:score-certificate}--\ref{thm:score-rademacher}}};

\node[anchor=center,inner sep=0pt] at (5.05,-2.78) {%
  \begin{tabular}{@{}c@{\hspace{0.30em}}l@{\hspace{0.95em}}c@{\hspace{0.30em}}l@{}}
    \tikz[baseline=-0.55ex]
      \node[legendbox,fill=blue!4,draw=blue!42] {}; &
    {\tiny Estimation error} &
    \tikz[baseline=-0.55ex]
      \node[legendbox,fill=orange!5,draw=orange!58] {}; &
    {\tiny Approximation error}
  \end{tabular}%
};

\end{tikzpicture}

%% file: appendix/assumptions.tex
\label{app:assumptions}

This appendix consolidates all notation used across the main text and appendices into a four-layer reference table (\Cref{tab:notations}), records our global tie-breaking, reachable-alphabet, and degenerate-regime conventions, and indexes the mathematical assumptions (\texttt{A1}--\texttt{A17} in \Cref{tab:assumptions}) by the exact results that invoke them.

\paragraph{Global tie-breaking, reachable-alphabet, and edge conventions.}
Whenever a classifier is induced by maximizing a real-valued score vector $t = (t_1, \ldots, t_K) \in \mathbb{R}^K$ ($K \ge 1$), ties among maximizers are broken deterministically in lexicographic order:
\begin{equation*}
  \lexargmax_{j \in [K]} t_j \;:=\; \min \arg\max_{j \in [K]} t_j.
\end{equation*}
Because $\{t \in \mathbb{R}^K : \lexargmax_{j \in [K]} t_j = k\}$ equals $\bigcap_{j < k}\{t_k > t_j\} \cap \bigcap_{j > k}\{t_k \ge t_j\}$, a finite intersection of open and closed half-spaces, composing $\lexargmax$ with any measurable score map preserves measurability (\Cref{sec:T1P:discharge,app:scoreoutput}).
For any sample size $m \ge 1$ and source-label vector $z = (z_1, \ldots, z_m) \in [K_{\rm S}]^m$, we write
\begin{equation*}
  \reachset(z) \;:=\; \{z_1, \ldots, z_m\} \;\subseteq\; [K_{\rm S}],
  \qquad
  |z|_{\#} \;:=\; |\reachset(z)|,
\end{equation*}
for the set of distinct source labels appearing in $z$ and its cardinality (\Cref{ex:gap}).
Given a target tuple $x_{1:m} = (x_1, \ldots, x_m) \in \XT^m$, the reachable source-label pattern set is $\ReaClass(x_{1:m}) := \{(r(x_1), \ldots, r(x_m)) : r \in \Rreach\} \subseteq [K_{\rm S}]^m$, and the realizable target-label pattern set on $z \in [K_{\rm S}]^m$ is $\RelClass(z) := \{(h(z_1), \ldots, h(z_m)) : h \in \Hfamily\} \subseteq [K_{\rm T}]^m$ (\Cref{def:patterns}).
The worst-case reachable alphabet size and restricted head Natarajan dimension on samples of size $m$ are
\begin{equation*}
  \tR(m) \;:=\; \sup_{x_{1:m} \in \XT^m} \sup_{z \in \ReaClass(x_{1:m})} |z|_{\#},
  \qquad
  \dHR(m) \;:=\; \sup_{x_{1:m} \in \XT^m} \sup_{z \in \ReaClass(x_{1:m})} \Ndim\bigl(\Hfamily|_{\reachset(z)}\bigr).
\end{equation*}
Throughout the main text and appendices, we adopt four standard degenerate-regime conventions: the empty product satisfies $\prod_{s=1}^{0} (\cdot) := 1$, the zeroth combinatorial power satisfies $\binom{K_{\rm T}}{2}^{0} := 1$ when $K_{\rm T} = 1$, the additive-MR output-capacity term $B_m := d_m\log\bigl(e\binom{K_{\rm T}}{2}m/d_m\bigr)$ with $d_m := d_{\Hfamily} \wedge m$ satisfies $B_m := 0$ when $d_m = 0$ or $K_{\rm T} = 1$ (\Cref{thm:T1P}), and $0\log(a/0) := 0$ for every $a \ge 0$.
Following standard empirical-process conventions~\citep{bartlett2002rademacher, mohri2018foundations}, whenever an empirical supremum or symmetrization step is taken over an uncountable hypothesis or loss class (including $\FMR$ in \Cref{thm:T1,cor:T1-sample,thm:T1P,cor:T1P-sample} and the score-output classes in \Cref{app:rademacher}), the class is assumed pointwise measurable (i.e., admitting a countable subset dense under pointwise convergence) so that empirical suprema and Rademacher averages are measurable and symmetrization applies.

\begingroup
\small
\renewcommand{\arraystretch}{1.15}
\begin{longtable}{@{}p{0.28\textwidth}p{0.68\textwidth}@{}}
\caption{Grouped summary of notation across the four structural layers of the analysis.}
\label{tab:notations}\\
\toprule
Symbol & Meaning and definition \\
\midrule
\endfirsthead
\toprule
Symbol & Meaning and definition \\
\midrule
\endhead
\bottomrule
\endlastfoot
\multicolumn{2}{@{}l}{\textbf{(1) Spaces and Distributions}} \\
\midrule
$\XT$, $\XS$ & Target and source input spaces ($\XT \subseteq \mathbb{R}^d$ and $\XS = \mathbb{R}^d$ in the additive setting of \Cref{sec:T1P}). \\
$[K_{\rm T}]$, $[K_{\rm S}]$ & Finite target and source label alphabets, where $[K] := \{1, \ldots, K\}$. \\
$\PT$, $\PS$ & Target joint distribution on $\XT \times [K_{\rm T}]$ and source joint distribution on $\XS \times [K_{\rm S}]$. \\
$P_{{\rm T},X}$, $P_{{\rm S},X}$; $g_\# P_{{\rm T},X}$ & Target- and source-input marginals, and pushforward of $P_{{\rm T},X}$ under $g: \XT \to \XS$. \\
$D_{\rm T}$, $D_{\rm S}$; $\nT$, $n_{\rm S}$ & i.i.d.\ target sample $\{(X_i, Y_i)\}_{i=1}^{\nT} \sim \PT^{\nT}$ and unlabeled source sample $\{U_j\}_{j=1}^{n_{\rm S}} \sim P_{{\rm S},X}^{n_{\rm S}}$. \\
$b_{\rm T}$, $b_{\rm S}$ & Measurable Bayes-optimal classifiers $b_{\rm T}: \XT \to [K_{\rm T}]$ and $b_{\rm S}: \XS \to [K_{\rm S}]$. \\
$\eta^g_y(u)$ & Regular conditional posterior $\Prob_{\PT}(Y = y \mid U_g = u)$ on standard Borel spaces (\Cref{app:scoreoutput:identity}). \\
\midrule
\multicolumn{2}{@{}l}{\textbf{(2) Wrapper and Hypothesis Classes}} \\
\midrule
$\fs$; $s_{\rm S}$, $B_s$ & Frozen hard-label source classifier $\fs: \XS \to [K_{\rm S}]$ and frozen source score map $s_{\rm S}: \XS \to [-B_s, B_s]^{K_{\rm S}}$ ($B_s \in [0, \infty)$). \\
$\lexargmax$; $\hat b_{\rm S}$, $c(u)$ & Lexicographic argmax $\min\arg\max$, score-induced source classifier $\hat b_{\rm S}(u) := \lexargmax_{k \in [K_{\rm S}]} s_{{\rm S},k}(u)$, and one-hot vector $c(u) := e_{\hat b_{\rm S}(u)} \in \mathbb{R}^{K_{\rm S}}$. \\
$g \in \Gfamily$; $\Gfamily_p$, $M$ & Learnable input transformation $\XT \to \XS$ and additive family $\Gfamily_p = \{g_\theta(x) = x + M\theta : \theta \in \Theta \subseteq \mathbb{R}^p\}$ with fixed $M \in \mathbb{R}^{d \times p}$. \\
$h \in \Hfamily$; $W \in \Wclass$, $B_h$ & Discrete output mapping $h: [K_{\rm S}] \to [K_{\rm T}]$ and linear score-output map $W \in \Wclass \subseteq \mathbb{R}^{K_{\rm T} \times K_{\rm S}}$ (identified with $u \mapsto Wu$) with $\|W\|_F \le B_h$ ($B_h \in [0, \infty)$). \\
$\Rreach$; $\FMR$, $\mathcal{F}_{\mathrm{score}}$ & Reachable source-label class $\{\fs \circ g : g \in \Gfamily\}$, hard-label wrapper class $\{f_{g,h} := h \circ \fs \circ g\}$, and logit-exposing wrapper class $\{f_{g,W}\}$ (\Cref{def:score-wrapper}). \\
$\widehat R_n(f)$; $\widehat f_n$, $\eta_n$ & Empirical $0$--$1$ risk $\nT^{-1}\sum_{i=1}^{\nT}\one\{f(X_i)\neq Y_i\}$ and measurable $\eta_n$-approximate ERM satisfying $\widehat R_n(\widehat f_n) \le \inf_{f \in \FMR}\widehat R_n(f) + \eta_n$ ($\eta_n \ge 0$). \\
$\mathfrak{F}^{\mathrm{bin}}_{d_{\Rreach}, d_{\Hfamily}}(\gamma)$ & Explicit binary target distribution family on $m = d_{\Rreach} + d_{\Hfamily}$ canonical points with margin $\gamma \in (0, 1/2]$ (\Cref{con:binary-family,app:proof_T3P:construction}). \\
\midrule
\multicolumn{2}{@{}l}{\textbf{(3) Reachability and Relabeling Growth Measures}} \\
\midrule
$\ReaClass(x_{1:m})$, $\RelClass(z)$ & Reachable source-label patterns on $x_{1:m} \in \XT^m$ and realizable target-label patterns on $z \in [K_{\rm S}]^m$ (\Cref{def:patterns}). \\
$\reachset(z)$, $|z|_{\#}$, $\tR(m)$ & Distinct source labels $\{z_1, \ldots, z_m\}$, alphabet size $|\reachset(z)|$, and worst-case reachable alphabet size (\Cref{ex:gap}). \\
$\PihatR(m)$, $\PihatH(m)$, $\PihatHall(m)$ & Sample-wise reachability growth $\sup_{x_{1:m}}|\ReaClass(x_{1:m})|$, conditional-relabeling growth $\sup_{x_{1:m}}\sup_{z \in \ReaClass(x_{1:m})}|\RelClass(z)|$, and structure-oblivious proxy $\sup_{z \in [K_{\rm S}]^m}|\RelClass(z)|$ (\Cref{def:growth}). \\
$\Ndim(\Hfamily)$, $d_{\Hfamily}$, $\dHR(m)$ & Multiclass Natarajan dimension of $\Hfamily$, its global upper bound $d_{\Hfamily}$ (\Cref{cond:nat}), and its reachable-alphabet restriction $\sup_{x_{1:m}}\sup_{z \in \ReaClass(x_{1:m})}\Ndim(\Hfamily|_{\reachset(z)})$. \\
$c_{\Rreach}$, $C_{\Rreach}$; $\Phip(N)$ & Cell-count exponent and scale constants in $(C_{\Rreach} m)^{c_{\Rreach} p}$ (\Cref{cond:arrangement}), and weak affine sign-pattern bound $\sum_{j=0}^{\min\{p,N\}} 2^j \binom{N}{j}$ (\Cref{lem:weak-sign}). \\
\midrule
\multicolumn{2}{@{}l}{\textbf{(4) Risks, Residuals, and Certificates}} \\
\midrule
$\Risk(f)$, $\RiskClsOptimum$, $\RiskStar$ & Population target $0$--$1$ risk $\Prob_{\PT}(f(X) \neq Y)$, in-class optimum $\inf_{f \in \FMR}\Risk(f)$, and unrestricted target Bayes risk $\inf_f \Risk(f)$. \\
$\EstRisk(\widehat f_n)$, $a_{\mathrm{MR}}$ & Target estimation error $\Risk(\widehat f_n) - \RiskClsOptimum$ relative to $\FMR$ and population approximation error $\RiskClsOptimum - \RiskStar$ (\Cref{def:errors}). \\
$Z_g$, $U_g$; $R_{\Hfamily}^{\ast}(g)$, $R_{\mathrm{all}}^{\ast}(g)$ & Induced source label $\fs(g(X))$ and logit vector $s_{\rm S}(g(X))$; $\Hfamily$-restricted and unrestricted $0$--$1$ channel optima given $Z_g$ (\Cref{prop:T2P}). \\
$R^{\star}_{\rm T}(U_g)$, $L^{\star}_\ell(U_g)$ & Unrestricted $0$--$1$ and surrogate channel Bayes risks given $U_g$ (\Cref{prop:score-01-identity,prop:score-surrogate-identity}). \\
$\Loss(g,W)$, $\LossStar$; $a_{\mathrm{score},0/1}$, $a_{\mathrm{score},\ell}$ & Population surrogate target risk $\E_{\PT}[\ell(Y, W s_{\rm S}(g(X)))]$, unrestricted surrogate Bayes risk, and logit-wrapper $0$--$1$ and surrogate approximation errors (\Cref{app:scoreoutput:wrapper}). \\
$\lamhard$, $\epshard$, $\Deltahard$, $\tauhard$ & Hard-label target Bayes mismatch $\Prob_{P_{{\rm T},X}}(b_{\rm T}(X) \neq h(b_{\rm S}(g(X))))$, source relabeled error $\Prob_{P_{{\rm S},X}}(h(\fs(U)) \neq h(b_{\rm S}(U)))$, pushforward discrepancy $|\E_{g_\# P_{{\rm T},X}}[\phi_h] - \E_{P_{{\rm S},X}}[\phi_h]|$ with $\phi_h(u) := \one\{h(\fs(u)) \neq h(b_{\rm S}(u))\}$, and population certificate $\tauhard := \inf_{g,h}\{\lamhard + \epshard + \Deltahard\}$ (\Cref{thm:T2C}). \\
$\lamscore$, $\zeta_W$, $\epsscore$, $\Deltascore$, $\tauscore$, $\widehat{\tauscore}$ & Score-output plug-in mismatch, pointwise surrogate distortion, source distortion, pushforward discrepancy, population certificate $\tauscore := \lamscore + \epsscore + \Deltascore$, and empirical plug-in certificate $\widehat{\tauscore}$ (\Cref{app:scoreoutput:certificate,app:scoreoutput:empirical}). \\
$\widehat{\mathfrak{R}}_{Z_{1:n}}(\mathcal{F})$, $\Gamma_Q$; $R_{U1}$, $B_\lambda$, $B_\zeta$ & Empirical Rademacher complexity on $Z_{1:n}$ and uniform concentration slack on candidate set $Q$; score-ball radius $B_h\max\{B_s\sqrt{K_{\rm S}}, 1\}$ and loss/distortion envelopes (\Cref{cor:score-finite-library,thm:score-rademacher}). \\
\end{longtable}
\endgroup

\paragraph{Minimal requirements for exact identities versus certificates.}
The population approximation identities require no boundedness, Lipschitz, concentration, or calibration conditions. Specifically, \Cref{prop:T2P} uses only measurability of $\fs$ and $g \in \Gfamily$, finite label alphabets, and nonemptiness of $\Gfamily$ and $\Hfamily \subseteq [K_{\rm T}]^{[K_{\rm S}]}$. \Cref{prop:score-01-identity} adds standard Borel target spaces to guarantee regular conditional posteriors $\eta^g_y(u) = \Prob_{\PT}(Y = y \mid U_g = u)$ under $\lexargmax$ tie-breaking, and \Cref{prop:score-surrogate-identity} assumes measurable $s_{\rm S}$, measurable $g \in \Gfamily$, nonempty $\Gfamily$ and $\Wclass$, measurable $\ell: [K_{\rm T}] \times \mathbb{R}^{K_{\rm T}} \to [0, \infty)$, and finite risks $\LossStar < \infty$ and $\inf_{W \in \Wclass} \Loss(g, W) < \infty$ (with the pointwise conditional Bayes representation in \Cref{rem:surrogate-bayes-form} invoking standard measurable-selection conditions).
By contrast, the score-output certificates (\Cref{thm:score-certificate,cor:score-finite-library,thm:score-rademacher}) invoke the norm, Lipschitz, and evaluation-independence assumptions indexed in \texttt{A8}--\texttt{A11} below and \Cref{app:rad-assumptions}.

\begingroup
\small
\renewcommand{\arraystretch}{1.12}
\begin{longtable}{@{}p{0.07\textwidth}p{0.63\textwidth}p{0.24\textwidth}@{}}
\caption{Theorem-indexed assumptions (\texttt{A1}--\texttt{A17}) used across the main text and appendices.}
\label{tab:assumptions}\\
\toprule
Label & Assumption & Used in \\
\midrule
\endfirsthead
\toprule
Label & Assumption & Used in \\
\midrule
\endhead
\bottomrule
\endlastfoot
A1 & Deterministic hard-label wrapper $f_{g,h} = h \circ \fs \circ g$ with measurable $\fs: \XS \to [K_{\rm S}]$ and $g \in \Gfamily$ (with $\FMR$ pointwise measurable as above when uncountable). & \Cref{thm:T1-app,cor:T1-sample-app,thm:T1P-app,thm:T3P-app,thm:T2-app} \\
A2 & Bounded $0$--$1$ classification loss $\one\{f(x) \neq y\} \in \{0, 1\}$. & \Cref{thm:T1-app,cor:T1-sample-app,thm:T1P-app,thm:T3P-app,thm:T2-app} \\
A3 & i.i.d.\ target sample $D_{\rm T} = \{(X_i, Y_i)\}_{i=1}^{\nT} \sim \PT^{\nT}$. & \Cref{thm:T1-app,cor:T1-sample-app,thm:T1P-app,thm:T3P-app} \\
A4 & Measurable $\eta_n$-approximate ERM $\widehat f_n \in \FMR$ with optimization slack $\eta_n \ge 0$. & \Cref{thm:T1-app,cor:T1-sample-app,thm:T1P-app} \\
A5 & Measurable Bayes-optimal selectors $b_{\rm T}: \XT \to [K_{\rm T}]$ and $b_{\rm S}: \XS \to [K_{\rm S}]$. & \Cref{thm:T2C,thm:T2-app} \\
A6 & Nonempty discrete output-mapping class $\Hfamily \subseteq [K_{\rm T}]^{[K_{\rm S}]}$. & \Cref{prop:T2P,thm:T2C,thm:T2-app} \\
A7 & Finite-support binary construction defining $\mathfrak{F}^{\mathrm{bin}}_{d_{\Rreach}, d_{\Hfamily}}(\gamma)$ (\Cref{app:proof_T3P:construction}). & \Cref{con:binary-family,thm:T3P,thm:T3P-app} \\
A8 & Bounded source score map $s_{\rm S}: \XS \to [-B_s, B_s]^{K_{\rm S}}$ with $B_s \in [0, \infty)$. & \Cref{thm:score-certificate,thm:T2P-app,cor:C2prime-app,thm:rad-score-output-certificate} \\
A9 & Norm-bounded linear score-output mappings $\|W\|_F \le B_h$ ($B_h \in [0, \infty)$) for all $W \in \Wclass$. & \Cref{thm:score-certificate,thm:T2P-app,cor:C2prime-app,thm:rad-score-output-certificate} \\
A10 & Measurable surrogate $\ell \ge 0$, $L_\ell$-Lipschitz on $B(0, R_{U1})$, bounded by $B_\lambda < \infty$, and admitting multiclass regret transform $\psi_\ell^{-1}$ when calibrated. & \Cref{thm:score-certificate,thm:T2P-app,cor:C2prime-app,thm:rad-score-output-certificate} \\
A11 & Score-induced source classifier $\hat b_{\rm S}(u) = \lexargmax_{k \in [K_{\rm S}]} s_{{\rm S},k}(u)$ under deterministic lexicographic tie-breaking. & \Cref{thm:score-certificate,thm:T2P-app,cor:C2prime-app,thm:rad-score-output-certificate} \\
A12 & Additive input-transformation class $\Gfamily_p = \{g_\theta(x) = x + M\theta : \theta \in \Theta \subseteq \mathbb{R}^p\}$ with fixed $M \in \mathbb{R}^{d \times p}$. & \Cref{thm:T1P,cor:T1P-sample,thm:T1P-app,cor:C1prime-app,cor:relu-discharge-app} \\
A13 & Reachability arrangement partition of $\Theta$ into at most $(C_{\Rreach} m)^{c_{\Rreach} p}$ constant-source-label cells on every $x_{1:m} \in \XT^m$ (\Cref{cond:arrangement}). & \Cref{thm:T1P,cor:T1P-sample,thm:T1P-app} \\
A14 & Output-mapping Natarajan dimension bound $\Ndim(\Hfamily) \le d_{\Hfamily}$ (\Cref{cond:nat}). & \Cref{thm:T1P,cor:T1P-sample,thm:T1P-app} \\
A15 & Affine source-score subclass $s_{\rm S}(u) = Au + b$ with $\fs(u) = \lexargmax_{k \in [K_{\rm S}]} s_{{\rm S},k}(u)$. & \Cref{cor:T1P:affine,cor:C1prime-app} \\
A16 & Lower-bound sample-size and margin scaling $d_{\Rreach} + d_{\Hfamily} \le c_0 \nT$ and $\gamma = c_1 \sqrt{(d_{\Rreach} + d_{\Hfamily})/\nT}$ (with capped margin $\gamma = \min\{1/4,\allowbreak c_1\sqrt{(d_{\Rreach} + d_{\Hfamily})/\nT}\}$ in \Cref{cor:T3P-sample}). & \Cref{thm:T3P,cor:T3P-sample,thm:T3P-app} \\
A17 & Fixed finite-depth feedforward ReLU source-score network with hidden widths $q_1, \ldots, q_L$ and $\fs(u) = \lexargmax_{k \in [K_{\rm S}]} s_{{\rm S},k}(u)$. & \Cref{cor:T1P:relu,cor:relu-discharge-app} \\
\end{longtable}
\endgroup

\paragraph{Additional evaluation-independence conditions for empirical certificates.}
\Cref{cor:score-finite-library} (\Cref{cor:C2prime-app}) requires the finite candidate library $Q$ of size $M$ to be fixed independently of the certification samples $(D_{\rm T}, D_{\rm S})$, or selected on an independent holdout or cross-fitting fold prior to evaluation.
\Cref{thm:score-rademacher} (\Cref{thm:rad-score-output-certificate}) requires the candidate set $Q \subseteq \Gfamily \times \Wclass$, the mutually independent samples $(D_{\rm T}, D_{\rm S})$, and the induced scalar function classes $(\mathcal{L}_Q, \mathcal{Z}_{{\rm T}, Q}, \mathcal{Z}_{{\rm S}, Q})$ to satisfy the measurability and independence conditions detailed in \Cref{app:rad-assumptions}.

%% file: appendix/app_approximation_error.tex
\Cref{sec:T2} characterizes the population approximation error $a_{\mathrm{MR}}$ when the frozen source model exposes discrete labels in $[K_{\rm S}]$.
We now extend that approximation analysis to source models exposing real-valued logit (or score) vectors in $\mathbb{R}^{K_{\rm S}}$, where a learnable linear output mapping replaces discrete relabeling\footnote{Unlike the population approximation identities and held-out evaluation certificates established here, the discrete product factorization of \Cref{prop:factorization} (and hence the agnostic-PAC estimation bound of \Cref{thm:T1}) relies on the finite alphabet $[K_{\rm S}]^m$ and does not transfer automatically. \Cref{sec:discussion} outlines the metric-complexity route required for same-sample ERM estimation over continuous logits.}.

\Cref{app:scoreoutput:wrapper} defines the logit-exposing wrapper class together with its $0$--$1$ and surrogate approximation errors.
\Cref{app:scoreoutput:identity} establishes exact population decompositions at both interfaces (\Cref{prop:score-01-identity,prop:score-surrogate-identity}, proved in \Cref{app:proof-identities}), paralleling \Cref{prop:T2P}.
\Cref{app:scoreoutput:certificate} derives a pointwise population certificate from observable plug-in quantities (\Cref{thm:score-certificate}, proved in \Cref{app:score-certificate-proofs}), paralleling \Cref{thm:T2C}.
\Cref{app:scoreoutput:empirical} converts this population certificate into two finite-sample evaluation guarantees via an empirical plug-in certificate: a simultaneous Hoeffding bound over an evaluation-independent finite library (\Cref{cor:score-finite-library}, proved in \Cref{app:score-certificate-proofs}) and a uniform empirical Rademacher bound over a fixed candidate set (\Cref{thm:score-rademacher}, proved in \Cref{app:rademacher}).

\subsection{Logit-exposing wrapper architecture}
\label{app:scoreoutput:wrapper}

Replacing the discrete source classifier $\fs: \XS \to [K_{\rm S}]$ of \Cref{sec:prelim:wrapper} with a frozen measurable score map $s_{\rm S}: \XS \to \mathbb{R}^{K_{\rm S}}$ exposes a continuous logit vector at each transformed target input.
A learnable linear output mapping $W \in \Wclass \subseteq \mathbb{R}^{K_{\rm T} \times K_{\rm S}}$ projects $s_{\rm S}(g(x))$ to a target score vector $W s_{\rm S}(g(x)) \in \mathbb{R}^{K_{\rm T}}$, and the induced classifier selects a target label via the deterministic lexicographic rule $\lexargmax_{j \in [K]} t_j := \min \arg\max_{j \in [K]} t_j$ (\Cref{sec:T1P:discharge,app:assumptions}).

\begin{definition}[Source-out MR predictor]
\label{def:score-wrapper}
Given measurable input spaces $\XT$ and $\XS$ and finite label alphabets $[K_{\rm T}]$ and $[K_{\rm S}]$, for any fixed measurable score map $s_{\rm S}: \XS \to \mathbb{R}^{K_{\rm S}}$, any class $\Gfamily$ of measurable input transformations $g: \XT \to \XS$, and any linear output-mapping class $\Wclass \subseteq \mathbb{R}^{K_{\rm T} \times K_{\rm S}}$, the \emph{source-out MR predictor} associated with $(g, W) \in \Gfamily \times \Wclass$ is
\begin{equation*}
  f_{g, W}(x) \;:=\; \lexargmax_{y \in [K_{\rm T}]} \bigl(W\, s_{\rm S}(g(x))\bigr)_y, \qquad \text{for every } x \in \XT,
\end{equation*}
and the induced logit-exposing wrapper hypothesis class is
\begin{equation*}
  \mathcal{F}_{\mathrm{score}} \;:=\; \bigl\{\, f_{g, W} : g \in \Gfamily,\; W \in \Wclass \,\bigr\} \;\subseteq\; [K_{\rm T}]^{\XT}.
\end{equation*}
\end{definition}

For nonempty $\Gfamily$ and $\Wclass$ and a target distribution $\PT$ over $\XT \times [K_{\rm T}]$ with $0$--$1$ risk $\Risk(f) = \Prob_{(X,Y)\sim\PT}(f(X) \neq Y)$ and unrestricted Bayes risk $\RiskStar = \inf_f \Risk(f)$ over all measurable $f: \XT \to [K_{\rm T}]$, the $0$--$1$ approximation error of $\mathcal{F}_{\mathrm{score}}$ is
\begin{equation*}
  a_{\mathrm{score}, 0/1} \;:=\; \inf_{(g, W) \in \Gfamily \times \Wclass}\, \Risk(f_{g, W}) - \RiskStar.
\end{equation*}
Because practical optimization replaces the discontinuous $0$--$1$ loss with a measurable surrogate loss $\ell : [K_{\rm T}] \times \mathbb{R}^{K_{\rm T}} \to [0, \infty)$, we also compare the surrogate target risk $\Loss(g, W) := \E_{(X,Y)\sim\PT}\bigl[\ell\bigl(Y, W s_{\rm S}(g(X))\bigr)\bigr]$ against the unrestricted surrogate Bayes risk $\LossStar := \inf_q \E_{(X,Y)\sim\PT}[\ell(Y, q(X))]$ over all measurable $q : \XT \to \mathbb{R}^{K_{\rm T}}$, defining the surrogate approximation error
\begin{equation*}
  a_{\mathrm{score}, \ell} \;:=\; \inf_{(g, W) \in \Gfamily \times \Wclass}\, \Loss(g, W) - \LossStar.
\end{equation*}
When $\ell$ is multiclass classification-calibrated in the sense of \citet{zhang2004statistical} and \citet{tewari2007consistency} (including multiclass logistic loss and centered Lee--Lin--Wahba hinge losses~\citep{lee2004multicategory}), a nondecreasing regret transform $\psi_\ell^{-1}: [0, \infty) \to [0, \infty)$ converts any upper bound on the surrogate excess risk $\Loss(g, W) - \LossStar$ into an upper bound on the $0$--$1$ excess risk $\Risk(f_{g, W}) - \RiskStar$.

\subsection{Approximation identities}
\label{app:scoreoutput:identity}

To test whether exposing continuous logits alters the coupled two-stage structure of \Cref{prop:T2P}, fix $g \in \Gfamily$ and isolate the lowest target $0$--$1$ risk achievable from the induced logit vector $U_g := s_{\rm S}(g(X)) \in \mathbb{R}^{K_{\rm S}}$ on $(X, Y) \sim \PT$ before restricting the head to $\Wclass$.
For measurable $a: \mathbb{R}^{K_{\rm S}} \to [K_{\rm T}]$, we abbreviate $\Risk(a(U_g)) := \Risk(a \circ s_{\rm S} \circ g) = \Prob_{(X,Y)\sim\PT}(a(U_g) \neq Y)$.
On standard Borel spaces admitting regular conditional posteriors $\eta^g_y(u) := \Prob_{\PT}(Y = y \mid U_g = u)$ for $y \in [K_{\rm T}]$, the unrestricted logit-conditioned $0$--$1$ Bayes risk under $g$ is
\begin{equation*}
  R^\star_{\rm T}(U_g) \;:=\; \inf_{a : \mathbb{R}^{K_{\rm S}} \to [K_{\rm T}]\ \text{measurable}}\, \Risk\bigl(a(U_g)\bigr) \;=\; \E_{(X,Y)\sim\PT}\Bigl[1 - \max_{y \in [K_{\rm T}]} \eta^g_y(U_g)\Bigr],
\end{equation*}
which is attained by the measurable selector $b_g^\star(u) := \lexargmax_{y \in [K_{\rm T}]} \eta^g_y(u)$.
Telescoping each fixed-$g$ excess risk around $R^\star_{\rm T}(U_g)$ inside the nested infimum over $(g, W)$ (\cref{app:proof-prop-score-01} in \Cref{app:proof-identities}) yields the exact $0$--$1$ decomposition below.

\begin{proposition}[Score-output $0$--$1$ approximation identity]
\label{prop:score-01-identity}
Suppose $\XT \times [K_{\rm T}]$ is a standard Borel space.
For every target distribution $\PT$ over $\XT \times [K_{\rm T}]$, every fixed measurable score map $s_{\rm S}: \XS \to \mathbb{R}^{K_{\rm S}}$, every nonempty class $\Gfamily$ of measurable maps $\XT \to \XS$, and every nonempty linear output-mapping class $\Wclass \subseteq \mathbb{R}^{K_{\rm T} \times K_{\rm S}}$, let $U_g := s_{\rm S}(g(X))$ for $(X, Y) \sim \PT$ and $a_W(u) := \lexargmax_{y \in [K_{\rm T}]} (W u)_y$ for $u \in \mathbb{R}^{K_{\rm S}}$.
Then
\begin{equation*}
  a_{\mathrm{score}, 0/1}
  \;=\;
  \inf_{g \in \Gfamily}
  \Bigl\{
    \underbrace{R^\star_{\rm T}(U_g) - \RiskStar}_{\textup{logit-channel information loss}}
    \;+\;
    \inf_{W \in \Wclass}
    \underbrace{\bigl(\Risk(a_W(U_g)) - R^\star_{\rm T}(U_g)\bigr)}_{\textup{linear-mapping restriction}}
  \Bigr\},
\end{equation*}
where both underbraced summands are nonnegative for every $g \in \Gfamily$.
\end{proposition}

\Cref{prop:score-01-identity} parallels \Cref{prop:T2P} term for term at the continuous interface: $R^\star_{\rm T}(U_g) - \RiskStar$ measures target Bayes information lost when $X$ is compressed into $U_g$ (vanishing if and only if $R^\star_{\rm T}(U_g) = \RiskStar$, for instance when $s_{\rm S} \circ g$ is Borel-injective), whereas $\inf_{W \in \Wclass}\bigl(\Risk(a_W(U_g)) - R^\star_{\rm T}(U_g)\bigr)$ measures the $0$--$1$ excess risk of restricting the head to linear score projections $a_W$ rather than arbitrary measurable classifiers on $\mathbb{R}^{K_{\rm S}}$ (vanishing whenever some $W \in \Wclass$ satisfies $a_W(U_g) = b_g^\star(U_g)$ almost surely).
As in \Cref{prop:T2P}, both summands depend on the same $g$ inside the outer infimum.

To isolate the corresponding bottlenecks under a measurable surrogate loss $\ell$, define the unrestricted logit-conditioned surrogate Bayes risk at fixed $g \in \Gfamily$ by
\begin{equation*}
  L^\star_\ell(U_g) \;:=\; \inf_{q : \mathbb{R}^{K_{\rm S}} \to \mathbb{R}^{K_{\rm T}}\ \text{measurable}}\, \E_{(X,Y)\sim\PT}\bigl[\ell\bigl(Y, q(U_g)\bigr)\bigr].
\end{equation*}
Adding and subtracting the finite intermediate optimum $L^\star_\ell(U_g)$ at each $g \in \Gfamily$ (\cref{app:proof-prop-surrogate,rem:surrogate-bayes-form} in \Cref{app:proof-identities}) gives the exact surrogate decomposition.

\begin{proposition}[Score-output surrogate approximation identity]
\label{prop:score-surrogate-identity}
For every target distribution $\PT$ over $\XT \times [K_{\rm T}]$, every fixed measurable score map $s_{\rm S}: \XS \to \mathbb{R}^{K_{\rm S}}$, every nonempty class $\Gfamily$ of measurable maps $\XT \to \XS$, every nonempty linear output-mapping class $\Wclass \subseteq \mathbb{R}^{K_{\rm T} \times K_{\rm S}}$, and every measurable surrogate loss $\ell : [K_{\rm T}] \times \mathbb{R}^{K_{\rm T}} \to [0, \infty)$ satisfying $\LossStar < \infty$ and $\inf_{W \in \Wclass} \Loss(g, W) < \infty$ for every $g \in \Gfamily$, let $U_g := s_{\rm S}(g(X))$ for $(X, Y) \sim \PT$.
Then
\begin{equation*}
  a_{\mathrm{score}, \ell}
  \;=\;
  \inf_{g \in \Gfamily}
  \Bigl\{
    \underbrace{L^\star_\ell(U_g) - \LossStar}_{\textup{surrogate information loss}}
    \;+\;
    \inf_{W \in \Wclass}
    \underbrace{\bigl(\Loss(g, W) - L^\star_\ell(U_g)\bigr)}_{\textup{surrogate linear-mapping restriction}}
  \Bigr\},
\end{equation*}
where both underbraced summands are nonnegative for every $g \in \Gfamily$.
\end{proposition}

\Cref{prop:score-01-identity,prop:score-surrogate-identity} require no score boundedness or Lipschitz assumptions, and when $\Wclass$ contains an optimal measurable head $u \mapsto W u$ achieving $L^\star_\ell(U_g)$, surrogate linear-mapping restriction vanishes.
Evaluating $R^\star_{\rm T}(U_g)$, $\RiskStar$, $L^\star_\ell(U_g)$, and $\LossStar$ directly, however, requires unknown conditional posteriors and unrestricted Bayes regressors, motivating an observable upper bound on the surrogate excess risk $\Loss(g, W) - \LossStar$.

\subsection{Source-conditional surrogate certificate}
\label{app:scoreoutput:certificate}

Whereas the hard-label certificate of \Cref{thm:T2C} interpolates the unknown source Bayes selector $b_{\rm S}$, a logit-exposing source model directly furnishes its own deterministic score-induced classifier $\hat b_{\rm S}(u) := \lexargmax_{k \in [K_{\rm S}]} s_{{\rm S}, k}(u)$ and one-hot plug-in vector $c(u) := e_{\hat b_{\rm S}(u)} \in \mathbb{R}^{K_{\rm S}}$.
For every measurable pair $(g, W) \in \Gfamily \times \Wclass$, quantizing $s_{\rm S}$ to $c$ on $\PT$ defines the target plug-in surrogate mismatch $\lamscore(g, W)$, while comparing full logits $W s_{\rm S}(u)$ against $W c(u)$ defines the pointwise surrogate distortion $\zeta_W(u)$, its source-marginal expectation $\epsscore(W)$, and its pushforward discrepancy $\Deltascore(g, W)$:
\begin{equation*}
\begin{aligned}
  \lamscore(g, W)
    &\;:=\; \E_{(X, Y) \sim \PT}\bigl[\ell\bigl(Y,\, W c(g(X))\bigr)\bigr], \\
  \zeta_W(u)
    &\;:=\; \max_{y \in [K_{\rm T}]}\, \bigl|\ell\bigl(y, W s_{\rm S}(u)\bigr) - \ell\bigl(y, W c(u)\bigr)\bigr|, \\
  \epsscore(W)
    &\;:=\; \E_{U \sim P_{{\rm S}, X}}\bigl[\zeta_W(U)\bigr], \\
  \Deltascore(g, W)
    &\;:=\; \bigl|\E_{X \sim P_{{\rm T}, X}}\bigl[\zeta_W(g(X))\bigr] \;-\; \E_{U \sim P_{{\rm S}, X}}\bigl[\zeta_W(U)\bigr]\bigr|.
\end{aligned}
\end{equation*}
Summing these three nonnegative terms defines the population certificate $\tauscore(g, W) := \lamscore(g, W) + \epsscore(W) + \Deltascore(g, W)$.
Interpolating $\ell(Y, W c(g(X)))$, transferring $\zeta_W$ from $P_{{\rm S}, X}$ to $g_\# P_{{\rm T}, X}$, and using $\LossStar \ge 0$ (\cref{thm:T2P-app} in \Cref{app:score-certificate-proofs}) yields the pointwise bound below.

\begin{theorem}[Source-conditional surrogate certificate]
\label{thm:score-certificate}
For every target distribution $\PT$ over $\XT \times [K_{\rm T}]$, every source-input marginal $P_{{\rm S}, X}$ on $\XS$, and constants $B_s, B_h, L_\ell \in [0, \infty)$, suppose $s_{\rm S}: \XS \to [-B_s, B_s]^{K_{\rm S}}$ is measurable, every $W \in \Wclass$ satisfies $\|W\|_F \le B_h$, and the measurable surrogate loss $\ell : [K_{\rm T}] \times \mathbb{R}^{K_{\rm T}} \to [0, \infty)$ is $L_\ell$-Lipschitz in its second argument with respect to $\|\cdot\|_2$ on the closed Euclidean ball $B(0, R_{U1}) \subseteq \mathbb{R}^{K_{\rm T}}$ of radius $R_{U1} := B_h \max\{B_s \sqrt{K_{\rm S}}, 1\}$ (\texttt{A8}--\texttt{A11} in \Cref{tab:assumptions}).
Then for every measurable pair $(g, W) \in \Gfamily \times \Wclass$,
\begin{equation*}
  \Loss(g, W) - \LossStar
  \;\le\;
  \tauscore(g, W)
  \;=\;
  \lamscore(g, W) + \epsscore(W) + \Deltascore(g, W).
\end{equation*}
Consequently, whenever $\ell$ is multiclass classification-calibrated with nondecreasing regret transform $\psi_\ell^{-1}: [0, \infty) \to [0, \infty)$,
\begin{equation*}
  \Risk(f_{g, W}) - \RiskStar
  \;\le\;
  \psi_\ell^{-1}\!\bigl(\tauscore(g, W)\bigr)
  \;=\;
  \psi_\ell^{-1}\!\bigl(\lamscore(g, W) + \epsscore(W) + \Deltascore(g, W)\bigr).
\end{equation*}
\end{theorem}

\Cref{thm:score-certificate} upper-bounds the surrogate excess risk $\Loss(g, W) - \LossStar$ and the $0$--$1$ excess risk $\Risk(f_{g, W}) - \RiskStar$ via the population certificate $\tauscore(g, W)$: $\lamscore(g, W)$ pays for target surrogate loss after quantizing $s_{\rm S}$ to its one-hot plug-in $c$, $\epsscore(W)$ pays for the deviation between full logits and $c$ under $W$ on $P_{{\rm S}, X}$, and $\Deltascore(g, W)$ pays for the shift of that distortion between $g_\# P_{{\rm T}, X}$ and $P_{{\rm S}, X}$.
When $s_{\rm S}(u) = c(u)$ is already one-hot across $\XS$, $\zeta_W \equiv 0$ so $\epsscore(W) = \Deltascore(g, W) = 0$, reducing $\tauscore(g, W)$ to $\lamscore(g, W)$; when $g_\# P_{{\rm T}, X} = P_{{\rm S}, X}$, $\Deltascore(g, W) = 0$.
Taking the infimum over nonempty $\Gfamily \times \Wclass$ gives $a_{\mathrm{score}, \ell} \le \inf_{(g, W) \in \Gfamily \times \Wclass} \tauscore(g, W)$.
Like \Cref{thm:T2C}, $\tauscore(g, W)$ is a one-directional upper certificate because the proof lower-bounds $\LossStar$ by $0$ and evaluates $\zeta_W$ pointwise at $W$ rather than via a symmetric supremum over $\Wclass$.

\subsection{Empirical certificates}
\label{app:scoreoutput:empirical}

Because $\lamscore(g, W)$, $\epsscore(W)$, and $\Deltascore(g, W)$ depend only on the known maps $(s_{\rm S}, c, \ell)$, they can be estimated directly from labeled target examples and unlabeled source inputs without auxiliary posterior estimation.
We establish two finite-sample guarantees for the resulting empirical plug-in certificate under evaluation-independent candidate selection.

\textbf{Finite-library empirical certificate.}
Let $Q = \{(g_r, W_r)\}_{r=1}^M \subseteq \Gfamily \times \Wclass$ ($M \ge 1$) be a finite library of measurable candidate pairs fixed independently of the certification samples (for example by preregistration or selection on a disjoint training split).
Given mutually independent i.i.d.\ certification samples $D_{\rm T} = \{(X_i, Y_i)\}_{i=1}^{\nT} \sim \PT^{\nT}$ and $D_{\rm S} = \{U_j\}_{j=1}^{n_{\rm S}} \sim P_{{\rm S}, X}^{n_{\rm S}}$ drawn independently of $Q$, define for each $(g, W) \in Q$ the empirical plug-in certificate $\widehat{\tauscore}(g, W)$ and its constituent sample means by
\begin{equation*}
\begin{aligned}
  \widehat{\lamscore}(g, W)
    &:= \frac{1}{\nT} \sum_{i=1}^{\nT} \ell\bigl(Y_i, W c(g(X_i))\bigr), \\
  \widehat{T}(g, W)
    &:= \frac{1}{\nT} \sum_{i=1}^{\nT} \zeta_W(g(X_i)), \quad
  \widehat{S}(W)
    := \frac{1}{n_{\rm S}} \sum_{j=1}^{n_{\rm S}} \zeta_W(U_j), \\
  \widehat{\tauscore}(g, W)
    &:= \widehat{\lamscore}(g, W) + \widehat{S}(W) + \bigl|\widehat{T}(g, W) - \widehat{S}(W)\bigr|.
\end{aligned}
\end{equation*}
Decomposing $\tauscore(g, W) - \widehat{\tauscore}(g, W)$ deterministically and applying Hoeffding's inequality across the $3M$ scalar sample means (\cref{cor:C2prime-app} in \Cref{app:score-certificate-proofs}) yields the simultaneous bound below.

\begin{corollary}[Finite-library empirical certificate]
\label{cor:score-finite-library}
Under the assumptions of \Cref{thm:score-certificate}, set $B_\zeta := L_\ell B_h(B_s\sqrt{K_{\rm S}} + 1)$ and $B_\lambda := \sup_{y \in [K_{\rm T}],\, q \in B(0, R_{U1})} \ell(y, q) < \infty$.
For every integer $M \ge 1$, every finite candidate library $Q = \{(g_r, W_r)\}_{r=1}^M \subseteq \Gfamily \times \Wclass$ of measurable pairs, all sample sizes $\nT, n_{\rm S} \ge 1$, and every $\delta \in (0, 1)$, if $D_{\rm T} \sim \PT^{\nT}$ and $D_{\rm S} \sim P_{{\rm S}, X}^{n_{\rm S}}$ are mutually independent i.i.d.\ samples drawn independently of $Q$, then with probability at least $1 - \delta$ over $(D_{\rm T}, D_{\rm S})$, simultaneously for all $(g, W) \in Q$,
\begin{equation*}
  \Loss(g, W) - \LossStar
  \;\le\;
  \widehat{\tauscore}(g, W)
  + B_\lambda \sqrt{\frac{\log(6 M / \delta)}{2 \nT}}
  + B_\zeta \sqrt{\frac{\log(6 M / \delta)}{2 \nT}}
  + 2 B_\zeta \sqrt{\frac{\log(6 M / \delta)}{2 n_{\rm S}}}.
\end{equation*}
Consequently, whenever $\ell$ is multiclass classification-calibrated with nondecreasing regret transform $\psi_\ell^{-1}: [0, \infty) \to [0, \infty)$, on the same $(1 - \delta)$-probability event, simultaneously for all $(g, W) \in Q$,
\begin{equation*}
  \Risk(f_{g, W}) - \RiskStar
  \;\le\;
  \psi_\ell^{-1}\!\left(
    \widehat{\tauscore}(g, W)
    + (B_\lambda + B_\zeta)\sqrt{\frac{\log(6 M / \delta)}{2 \nT}}
    + 2 B_\zeta \sqrt{\frac{\log(6 M / \delta)}{2 n_{\rm S}}}
  \right).
\end{equation*}
\end{corollary}

In \Cref{cor:score-finite-library}, the factor $2$ on the source-sample concentration term arises because $\widehat{S}(W)$ enters both the source-distortion estimate and the empirical discrepancy $|\widehat{T}(g, W) - \widehat{S}(W)|$ (\eqref{eq:C2prime-det-slack} in \Cref{app:score-certificate-proofs}).
At $M = 1$, \Cref{cor:score-finite-library} certifies a single held-out wrapper learned on an independent training split, whereas selecting $Q$ on $(D_{\rm T}, D_{\rm S})$ without sample splitting invalidates the independence hypothesis.

\textbf{Uniform empirical certificate over a fixed candidate set.}
When the candidate set $Q \subseteq \Gfamily \times \Wclass$ is large or infinite, the $\sqrt{\log(6M/\delta)}$ union-bound term in \Cref{cor:score-finite-library} becomes vacuous.
To replace finite-library counting with empirical capacity control while keeping $Q$ independent of $(D_{\rm T}, D_{\rm S})$, write $\widehat{\mathfrak{R}}_{Z_{1:n}}(\mathcal{F}) := n^{-1}\, \E_\sigma\bigl[\sup_{f \in \mathcal{F}} \sum_{i=1}^n \sigma_i f(Z_i) \bigm| Z_{1:n}\bigr]$ for the empirical Rademacher complexity of a scalar class $\mathcal{F}$ on $Z_{1:n}$ with independent uniform signs $\sigma_1, \ldots, \sigma_n \in \{-1, +1\}$, and define the three induced scalar function classes
\begin{equation*}
\begin{aligned}
  \mathcal{L}_Q &\;:=\; \bigl\{(x, y) \mapsto \ell\bigl(y, W c(g(x))\bigr) : (g, W) \in Q\bigr\}, \\
  \mathcal{Z}_{{\rm T}, Q} &\;:=\; \bigl\{x \mapsto \zeta_W(g(x)) : (g, W) \in Q\bigr\}, \qquad
  \mathcal{Z}_{{\rm S}, Q} \;:=\; \bigl\{u \mapsto \zeta_W(u) : (g, W) \in Q\bigr\}.
\end{aligned}
\end{equation*}
Applying two-sided empirical Rademacher uniform-deviation bounds~\citep{bartlett2002rademacher} across $\mathcal{L}_Q$ on $D_{\rm T}$, $\mathcal{Z}_{{\rm T}, Q}$ on the target-input tuple $X_{1:\nT} := (X_1, \ldots, X_{\nT})$ of $D_{\rm T}$, and $\mathcal{Z}_{{\rm S}, Q}$ on $D_{\rm S} = \{U_j\}_{j=1}^{n_{\rm S}}$ (\cref{thm:rad-score-output-certificate} in \Cref{app:rademacher}) yields the uniform guarantee below.

\begin{theorem}[Uniform empirical certificate over a fixed candidate set]
\label{thm:score-rademacher}
Under the assumptions of \Cref{thm:score-certificate}, let $B_\lambda, B_\zeta$ be defined as in \Cref{cor:score-finite-library}, and let $Q \subseteq \Gfamily \times \Wclass$ be a nonempty set of measurable candidate pairs whose induced scalar classes $\mathcal{L}_Q$, $\mathcal{Z}_{{\rm T}, Q}$, and $\mathcal{Z}_{{\rm S}, Q}$ satisfy the measurability conditions of \Cref{app:rad-assumptions}.
For all sample sizes $\nT, n_{\rm S} \ge 1$ and every $\delta \in (0, 1)$, if $D_{\rm T} = \{(X_i, Y_i)\}_{i=1}^{\nT} \sim \PT^{\nT}$ (with input tuple $X_{1:\nT}$) and $D_{\rm S} = \{U_j\}_{j=1}^{n_{\rm S}} \sim P_{{\rm S}, X}^{n_{\rm S}}$ are mutually independent i.i.d.\ samples drawn independently of $Q$, then with probability at least $1 - \delta$ over $(D_{\rm T}, D_{\rm S})$, simultaneously for all $(g, W) \in Q$,
\begin{equation*}
  \Loss(g, W) - \LossStar
  \;\le\;
  \widehat{\tauscore}(g, W) + \Gamma_Q,
\end{equation*}
where the uniform concentration slack $\Gamma_Q$ is
\begin{align*}
  \Gamma_Q
  \;:=\;& 2 \widehat{\mathfrak{R}}_{D_{\rm T}}(\mathcal{L}_Q)
    + 3 B_\lambda \sqrt{\frac{\log(12 / \delta)}{2 \nT}} \\
  &+ 2 \widehat{\mathfrak{R}}_{X_{1:\nT}}(\mathcal{Z}_{{\rm T}, Q})
    + 3 B_\zeta \sqrt{\frac{\log(12 / \delta)}{2 \nT}} \\
  &+ 4 \widehat{\mathfrak{R}}_{D_{\rm S}}(\mathcal{Z}_{{\rm S}, Q})
    + 6 B_\zeta \sqrt{\frac{\log(12 / \delta)}{2 n_{\rm S}}}.
\end{align*}
Consequently, whenever $\ell$ is multiclass classification-calibrated with nondecreasing regret transform $\psi_\ell^{-1}: [0, \infty) \to [0, \infty)$, on the same $(1 - \delta)$-probability event, simultaneously for all $(g, W) \in Q$,
\begin{equation*}
  \Risk(f_{g, W}) - \RiskStar
  \;\le\;
  \psi_\ell^{-1}\!\bigl(\widehat{\tauscore}(g, W) + \Gamma_Q\bigr).
\end{equation*}
\end{theorem}

\Cref{thm:score-rademacher} bounds the surrogate excess risk $\Loss(g, W) - \LossStar$ and calibrated $0$--$1$ excess risk $\Risk(f_{g, W}) - \RiskStar$ via the empirical plug-in certificate $\widehat{\tauscore}(g, W)$ plus empirical Rademacher complexities of $\mathcal{L}_Q$, $\mathcal{Z}_{{\rm T}, Q}$, and $\mathcal{Z}_{{\rm S}, Q}$; when $|Q| = 1$, all three empirical Rademacher terms vanish identically, recovering a single-candidate concentration bound up to constants.
Because $Q$ is fixed independently of $(D_{\rm T}, D_{\rm S})$, \Cref{thm:score-rademacher} certifies held-out candidate sets rather than same-sample ERM over $\Gfamily \times \Wclass$; extending to in-sample ERM over continuous logits requires bounding the complexity of the composed vector class $\{s_{\rm S} \circ g : g \in \Gfamily\}$ via vector contraction~\citep{maurer2016vector}, as outlined in \Cref{sec:discussion}.

%% file: appendix/proof_T1.tex
\label{app:proof_T1}

\textbf{Proof roadmap.}
This section establishes the finite-sample estimation guarantees of \Cref{sec:T1} together with the restricted head-growth lemma invoked later in \Cref{app:proof_T1P}.
We first prove an elementary fixed-point inversion lemma (\Cref{lem:log-inversion}) that converts logarithmic sample-size inequalities into explicit polynomial sample-complexity thresholds, and then factorize the sample-wise growth of $\FMR$ into reachability and conditional-relabeling components (\Cref{prop:factorization}).
Inserting this product bound into a two-sided growth-function uniform-deviation inequality yields the distribution-free agnostic-PAC estimation bound (\Cref{thm:T1-app}, restating \Cref{thm:T1}), which combines with \Cref{lem:log-inversion} to establish polynomial-growth sample complexity (\Cref{cor:T1-sample-app}, restating \Cref{cor:T1-sample}).
Finally, we prove dominance of conditional relabeling over the structure-oblivious proxy (\Cref{prop:no-worse}) and establish the reachable-alphabet Sauer--Shelah bound on $\PihatH(m)$ (\Cref{lem:restricted-head}), whose coarse specialization (\Cref{rem:restricted-head-recovery}) supplies the output-capacity bound in \Cref{thm:T1P-app}.

\subsection{Auxiliary fixed-point inversion lemma}

\begin{lemma}[Logarithmic fixed-point inversion]
\label{lem:log-inversion}
Let $a \ge 1$, $b \ge 1$, and $c \ge 0$.
For every real number $n$ satisfying
\begin{equation*}
  n \;\ge\; 4a\log(2ab) + 2c,
\end{equation*}
we have $n \ge a\log(bn) + c$.
\end{lemma}

\begin{proof}
Define $q(n) := n - a\log(bn) - c$ on $(0, \infty)$.
For every $n \ge a$, the derivative satisfies $q'(n) = 1 - a/n \ge 0$, so $q$ is nondecreasing on $[a, \infty)$.
Set $L := \log(2ab)$ and $n_0 := 4aL + 2c$.
Since $a \ge 1$ and $b \ge 1$, we have $2ab \ge 2$, which implies $L \ge \log 2 > 1/2$ and, using $4\log 2 > 1$ and $c \ge 0$, gives $n_0 \ge 4a\log 2 > a$.
By monotonicity of $q$ on $[a, \infty)$, it therefore suffices to show $q(n_0) \ge 0$, or equivalently $a\log(bn_0) + c \le n_0$.

Expanding $\log(bn_0)$ at $n_0 = 4aL + 2c$ yields
\begin{align*}
  \log(bn_0)
  &= \log(4abL + 2bc) \\
  &= \log(4abL) + \log\!\left(1 + \frac{c}{2aL}\right) \\
  &= L + \log(2L) + \log\!\left(1 + \frac{c}{2aL}\right)
    &&\text{($\log(2ab) = L$)} \\
  &\le L + \log(2L) + \frac{c}{2aL},
    &&\text{($\log(1 + u) \le u$)}
\end{align*}
where the final step uses $c/(2aL) \ge 0$.
Moreover, for every $L > 0$, truncating the exponential series at the quadratic term gives
\begin{align*}
  e^L
  &\ge 1 + L + \frac{L^2}{2}
    &&\text{(since $L > 0$)} \\
  &= 2L + \frac{1}{2}\bigl((L - 1)^2 + 1\bigr) \\
  &> 2L,
    &&\text{($(L - 1)^2 + 1 \ge 1$)}
\end{align*}
so taking logarithms gives $\log(2L) < L$.
Combining this bound with $1/(2L) < 1$ (since $L > 1/2$), $a \ge 1$, and $c \ge 0$ yields
\begin{align*}
  a\log(bn_0) + c
  &\le a\left(L + \log(2L) + \frac{c}{2aL}\right) + c \\
  &\le 2aL + \frac{c}{2L} + c
    &&\text{(since $\log(2L) < L$)} \\
  &\le 2aL + 2c
    &&\text{($1/(2L) < 1$, $c \ge 0$)} \\
  &\le 4aL + 2c
    &&\text{(since $aL > 0$)} \\
  &= n_0.
\end{align*}
Hence $q(n_0) \ge 0$, and monotonicity of $q$ on $[a, \infty)$ implies $n \ge a\log(bn) + c$ for every $n \ge n_0$.
\end{proof}

\begin{remark}[Role of the inversion constants]
The numerical factors $4$ and $2$ in \Cref{lem:log-inversion} are chosen for algebraic simplicity rather than tightness, converting the implicit logarithmic condition of \Cref{cor:T1-sample-app} into an explicit $\widetilde O(\cdot)$ sample-complexity threshold.
\end{remark}

\subsection{Proof of \Cref{prop:factorization}}

\begin{proof}
Fix an integer $m \ge 1$ and a target input tuple $x_{1:m} = (x_1, \ldots, x_m) \in \XT^m$.
For each reachable source-label predictor $r = \fs \circ g \in \Rreach$, write $z_r := (r(x_1), \ldots, r(x_m)) \in [K_{\rm S}]^m$, so that $\ReaClass(x_{1:m}) = \{z_r : r \in \Rreach\}$ by \Cref{def:patterns}.
Every wrapper predictor $f \in \FMR$ takes the form $f = h \circ r$ for some $r = \fs \circ g \in \Rreach$ and $h \in \Hfamily$, and evaluating $f$ coordinatewise on $x_{1:m}$ places its target-label vector in $\RelClass(z_r)$:
\begin{equation*}
  (f(x_1), \ldots, f(x_m)) \;=\; \bigl(h(r(x_1)), \ldots, h(r(x_m))\bigr) \;\in\; \RelClass(z_r).
\end{equation*}
Thus $\{(f(x_1), \ldots, f(x_m)) : f \in \FMR\} \subseteq \bigcup_{z \in \ReaClass(x_{1:m})} \RelClass(z)$, and counting realized target-label patterns gives
\begin{align*}
  \bigl|\{(f(x_1), \ldots, f(x_m)) : f \in \FMR\}\bigr|
  &\le \Bigl|\bigcup_{z \in \ReaClass(x_{1:m})} \RelClass(z)\Bigr| \\
  &\le \sum_{z \in \ReaClass(x_{1:m})} |\RelClass(z)| \\
  &\le |\ReaClass(x_{1:m})| \sup_{z \in \ReaClass(x_{1:m})} |\RelClass(z)| \\
  &\le \PihatR(m)\,\PihatH(m),
\end{align*}
where the second inequality applies the finite union bound and the fourth invokes \Cref{def:growth}.
Taking the supremum over $x_{1:m} \in \XT^m$ yields $\Pi_{\FMR}(m) \le \PihatR(m)\,\PihatH(m)$.
\end{proof}

\subsection{Proof of \Cref{thm:T1}}

Before proving \Cref{thm:T1}, we record the formal definition of agnostic-PAC learnability referenced in \Cref{sec:T1:bound}.
The theorem controls the class-relative estimation error $\EstRisk(\widehat f_n) = \Risk(\widehat f_n) - \RiskClsOptimum$, while the population approximation error $a_{\mathrm{MR}} = \RiskClsOptimum - \RiskStar$ enters only when converting estimation guarantees into target Bayes excess-risk bounds via \Cref{def:errors}.

\begin{definition}[Agnostic-PAC learnability~\citep{kearns1992toward}]
\label{def:pac}
A hypothesis class $\FMR \subseteq [K_{\rm T}]^{\XT}$ is \emph{agnostically PAC learnable} by $\eta_n$-approximate ERM ($\eta_n \ge 0$) if there exists a sample-complexity function $n_0 : (0,1)^2 \to \mathbb N$ such that, for every target distribution $\PT$ on $\XT \times [K_{\rm T}]$, every $\varepsilon, \delta \in (0,1)$, and every $\nT \ge n_0(\varepsilon, \delta)$, with probability at least $1 - \delta$ over an i.i.d.\ target sample of size $\nT$ drawn from $\PT$, every measurable $\eta_n$-approximate ERM $\widehat f_n \in \FMR$ satisfies $\EstRisk(\widehat f_n) \le \eta_n + \varepsilon$.
\end{definition}

\begin{theorem}[Factorized agnostic-PAC bound; restatement of \Cref{thm:T1}]
\label{thm:T1-app}
Let $\fs: \XS \to [K_{\rm S}]$ be a fixed measurable source classifier, $\Gfamily$ a nonempty class of measurable maps $\XT \to \XS$, and $\Hfamily \subseteq [K_{\rm T}]^{[K_{\rm S}]}$ a nonempty output-mapping class, inducing $\FMR = \{h \circ \fs \circ g : g \in \Gfamily,\, h \in \Hfamily\}$.
For every target distribution $\PT$ over $\XT \times [K_{\rm T}]$, every $\nT \ge 1$, every $\delta \in (0,1)$, and every $\eta_n \ge 0$, with probability at least $1 - \delta$ over an i.i.d.\ sample of size $\nT$ from $\PT$, every measurable $\eta_n$-approximate ERM $\widehat f_n \in \FMR$ satisfies
\begin{equation*}
  \EstRisk(\widehat f_n) \;\le\; \eta_n + 4\sqrt{6}\,\sqrt{\frac{\log\PihatR(2\nT) + \log\PihatH(2\nT) + \log(1/\delta)}{\nT}}.
\end{equation*}
\end{theorem}

\begin{proof}
Because $\Gfamily$ and $\Hfamily$ are nonempty and $\PT$ is a probability distribution on $\XT \times [K_{\rm T}]$, both $\FMR$ and $\XT$ are nonempty.
Define the induced binary $0$--$1$ loss class on $\XT \times [K_{\rm T}]$ by
\begin{equation*}
  \mathcal L_{\mathrm{MR}} \;:=\; \bigl\{(x, y) \mapsto \one\{f(x) \neq y\} : f \in \FMR\bigr\} \;\subseteq\; \{0, 1\}^{\XT \times [K_{\rm T}]}.
\end{equation*}
Fix any integer $m \ge 1$ and any labeled tuple $((x_1, y_1), \ldots, (x_m, y_m)) \in (\XT \times [K_{\rm T}])^m$.
Once $(y_1, \ldots, y_m)$ is fixed, the map $(u_1, \ldots, u_m) \mapsto (\one\{u_1 \neq y_1\}, \ldots, \one\{u_m \neq y_m\})$ is deterministic, so any two predictors $f, f' \in \FMR$ that agree on $x_{1:m}$ induce identical loss vectors on $((x_i, y_i))_{i=1}^m$.
Taking the supremum over labeled tuples of size $m$ and applying \Cref{prop:factorization} yields, for every $m \ge 1$,
\begin{equation}
\label{eq:T1-loss-growth}
  \Pi_{\mathcal L_{\mathrm{MR}}}(m) \;\le\; \Pi_{\FMR}(m) \;\le\; \PihatR(m)\,\PihatH(m).
\end{equation}
Set $G_n := \PihatR(2\nT)\,\PihatH(2\nT)$.
Since $\FMR \neq \varnothing$ and $\XT \neq \varnothing$, \eqref{eq:T1-loss-growth} gives $1 \le \Pi_{\mathcal L_{\mathrm{MR}}}(2\nT) \le G_n$.

Applying the two-sided growth-function uniform-deviation inequality of \citet[Corollary~3.9 and Eq.~(3.23)]{mohri2018foundations} to the $\{0, 1\}$-valued class $\mathcal L_{\mathrm{MR}}$ on the i.i.d.\ sample $\{(X_i, Y_i)\}_{i=1}^{\nT} \sim \PT^{\nT}$ (under the standard pointwise-measurability convention of \Cref{app:assumptions}) and bounding $\Pi_{\mathcal L_{\mathrm{MR}}}(2\nT) \le G_n$ via \eqref{eq:T1-loss-growth} gives an event of probability at least $1 - \delta$ on which
\begin{equation}
\label{eq:T1-uniform-dev}
  \sup_{f \in \FMR}\bigl|\Risk(f) - \widehat R_n(f)\bigr|
  \;\le\; \sqrt{\frac{8\bigl\{\log G_n + \log 4 + \log(1/\delta)\bigr\}}{\nT}},
\end{equation}
where $\widehat R_n(f) = \nT^{-1}\sum_{i=1}^{\nT}\one\{f(X_i) \neq Y_i\}$.
Fix an arbitrary $\nu > 0$ and choose a deterministic reference predictor $f_\nu \in \FMR$ satisfying $\Risk(f_\nu) \le \RiskClsOptimum + \nu$ by definition of $\RiskClsOptimum = \inf_{f \in \FMR}\Risk(f)$.
Every measurable $\eta_n$-approximate ERM $\widehat f_n \in \FMR$ satisfies $\widehat R_n(\widehat f_n) \le \inf_{f \in \FMR}\widehat R_n(f) + \eta_n \le \widehat R_n(f_\nu) + \eta_n$.
Telescoping $\EstRisk(\widehat f_n)$ into four differences gives
\begin{align*}
  \EstRisk(\widehat f_n)
  &= \Risk(\widehat f_n) - \RiskClsOptimum
    &&\text{(by \Cref{def:errors})} \\
  &= \bigl(\Risk(\widehat f_n) - \widehat R_n(\widehat f_n)\bigr)
     + \bigl(\widehat R_n(\widehat f_n) - \widehat R_n(f_\nu)\bigr) \\
  &\quad + \bigl(\widehat R_n(f_\nu) - \Risk(f_\nu)\bigr)
     + \bigl(\Risk(f_\nu) - \RiskClsOptimum\bigr) \\
  &\le 2\sup_{f \in \FMR}\bigl|\Risk(f) - \widehat R_n(f)\bigr| \\
  &\quad + \bigl(\widehat R_n(\widehat f_n) - \widehat R_n(f_\nu)\bigr)
     + \bigl(\Risk(f_\nu) - \RiskClsOptimum\bigr)
    &&\text{(two-sided supremum)} \\
  &\le 2\sup_{f \in \FMR}\bigl|\Risk(f) - \widehat R_n(f)\bigr| + \eta_n + \nu.
    &&\text{($\eta_n$-ERM and $f_\nu$)}
\end{align*}
Letting $\nu \downarrow 0$ and substituting \eqref{eq:T1-uniform-dev} with $2\sqrt{8} = 4\sqrt{2}$ yields, on the same $(1 - \delta)$-probability event,
\begin{equation}
\label{eq:T1-before-log-absorb}
  \EstRisk(\widehat f_n)
  \;\le\; \eta_n + 4\sqrt{2}\,\sqrt{\frac{\log G_n + \log 4 + \log(1/\delta)}{\nT}}.
\end{equation}
To absorb the additive $\log 4$ term, we split on the integer value of $G_n \ge 1$.

\textbf{Case 1 ($G_n = 1$).}
Monotonicity of sample-wise growth functions and \eqref{eq:T1-loss-growth} yield $1 \le \Pi_{\FMR}(1) \le \Pi_{\FMR}(2\nT) \le G_n = 1$, so $\Pi_{\FMR}(1) = 1$.
Every predictor in $\FMR$ therefore takes a common value at each $x \in \XT$, forcing $\Risk(f) = \RiskClsOptimum$ for all $f \in \FMR$ and hence $\EstRisk(\widehat f_n) = 0$.
Since $\eta_n \ge 0$, $\log G_n = \log\PihatR(2\nT) + \log\PihatH(2\nT) = 0$, and $\log(1/\delta) > 0$, the claimed bound holds deterministically.

\textbf{Case 2 ($G_n \ge 2$).}
Because $\log 4 = 2\log 2 \le 2\log G_n$ and $\log(1/\delta) > 0$, we obtain
\begin{align*}
  \log G_n + \log 4 + \log(1/\delta)
  &\le 3\log G_n + \log(1/\delta)
    &&\text{(since $G_n \ge 2$)} \\
  &\le 3\bigl\{\log G_n + \log(1/\delta)\bigr\}.
    &&\text{(since $\log(1/\delta) > 0$)}
\end{align*}
Substituting this inequality into \eqref{eq:T1-before-log-absorb}, using $4\sqrt{2}\cdot\sqrt{3} = 4\sqrt{6}$, and expanding $\log G_n = \log\PihatR(2\nT) + \log\PihatH(2\nT)$ completes the proof.
\end{proof}

\begin{remark}[Target Bayes excess-risk consequence]
\label{rem:T1-bayes-excess-consequence}
Combining \Cref{thm:T1-app} with the exact identity $\Risk(\widehat f_n) - \RiskStar = \EstRisk(\widehat f_n) + a_{\mathrm{MR}}$ (\Cref{def:errors}) gives, on the same $(1-\delta)$-probability event,
\begin{equation*}
  \Risk(\widehat f_n) - \RiskStar \;\le\; a_{\mathrm{MR}} + \eta_n + 4\sqrt{6}\,\sqrt{\frac{\log\PihatR(2\nT) + \log\PihatH(2\nT) + \log(1/\delta)}{\nT}}.
\end{equation*}
\end{remark}

\subsection{Proof of \Cref{cor:T1-sample}}

\begin{corollary}[Polynomial-growth sample complexity; restatement of \Cref{cor:T1-sample}]
\label{cor:T1-sample-app}
Under the wrapper setup of \Cref{thm:T1-app}, suppose there exist constants $d_{\Rreach}, d_{\Hfamily} \ge 0$ and $C_{\Rreach}, C_{\Hfamily} \ge 1$ such that $\log \PihatR(m) \le d_{\Rreach} \log(C_{\Rreach} m)$ and $\log \PihatH(m) \le d_{\Hfamily} \log(C_{\Hfamily} m)$ for every $m \ge 1$.
Then for every target distribution $\PT$ over $\XT \times [K_{\rm T}]$, every $\varepsilon, \delta \in (0,1)$, and every $\eta_n \ge 0$, there exists $n_0 = \widetilde O\bigl((d_{\Rreach} + d_{\Hfamily} + \log(1/\delta))/\varepsilon^2\bigr)$ such that whenever $\nT \ge n_0$, every measurable $\eta_n$-approximate ERM $\widehat f_n \in \FMR$ satisfies $\EstRisk(\widehat f_n) \le \eta_n + \varepsilon$ with probability at least $1 - \delta$ over $\PT^{\nT}$.
\end{corollary}

\begin{proof}
Fix a target distribution $\PT$ over $\XT \times [K_{\rm T}]$, $\varepsilon, \delta \in (0,1)$, and $\eta_n \ge 0$.
By \Cref{thm:T1-app}, it suffices to exhibit an integer $n_0 \ge 1$ such that the statistical term in \Cref{thm:T1-app} is at most $\varepsilon$ for all $\nT \ge n_0$.
Evaluating the assumed polynomial growth bounds at $m = 2\nT$ gives
\begin{equation*}
  \log\PihatR(2\nT) + \log\PihatH(2\nT) + \log(1/\delta)
  \;\le\; d_{\Rreach}\log(2C_{\Rreach}\nT) + d_{\Hfamily}\log(2C_{\Hfamily}\nT) + \log(1/\delta).
\end{equation*}
Since $(4\sqrt{6})^2 = 96$, the statistical term in \Cref{thm:T1-app} is at most $\varepsilon$ whenever
\begin{equation}
\label{eq:T1-sample-fixed-point}
  \nT \;\ge\; \frac{96}{\varepsilon^2}\Bigl\{d_{\Rreach}\log(2C_{\Rreach}\nT) + d_{\Hfamily}\log(2C_{\Hfamily}\nT) + \log(1/\delta)\Bigr\}.
\end{equation}
Set $d := d_{\Rreach} + d_{\Hfamily} \ge 0$ and $C_* := \max\{C_{\Rreach}, C_{\Hfamily}\} \ge 1$.
If $d = 0$, then $d_{\Rreach} = d_{\Hfamily} = 0$, and \eqref{eq:T1-sample-fixed-point} reduces directly to $\nT \ge 96\log(1/\delta)/\varepsilon^2$, which holds for all $\nT \ge n_0 := \bigl\lceil 96\log(1/\delta)/\varepsilon^2 \bigr\rceil = O\bigl(\log(1/\delta)/\varepsilon^2\bigr)$.

Assume henceforth that $d > 0$, and define
\begin{equation*}
  a \;:=\; \max\!\left\{1,\, \frac{96d}{\varepsilon^2}\right\} \;\ge\; 1,
  \qquad
  b \;:=\; 2C_* \;\ge\; 2,
  \qquad
  c \;:=\; \frac{96\log(1/\delta)}{\varepsilon^2} \;>\; 0.
\end{equation*}
For every $\nT \ge 1$, since $b\nT \ge 2$ implies $\log(b\nT) > 0$, the right-hand side of \eqref{eq:T1-sample-fixed-point} satisfies
\begin{align*}
  &\frac{96}{\varepsilon^2}\Bigl\{d_{\Rreach}\log(2C_{\Rreach}\nT) + d_{\Hfamily}\log(2C_{\Hfamily}\nT) + \log(1/\delta)\Bigr\} \\
  &\quad\le \frac{96}{\varepsilon^2}\bigl\{d\log(2C_*\nT) + \log(1/\delta)\bigr\}
    \quad\text{(since $C_{\Rreach}, C_{\Hfamily} \le C_*$)} \\
  &\quad= \frac{96d}{\varepsilon^2}\log(b\nT) + c \\
  &\quad\le a\log(b\nT) + c.
    \quad\text{(since $a \ge 96d/\varepsilon^2$ and $\log(b\nT) > 0$)}
\end{align*}
Consequently, every integer $\nT \ge 1$ satisfying $\nT \ge a\log(b\nT) + c$ also satisfies \eqref{eq:T1-sample-fixed-point}.
Invoking \Cref{lem:log-inversion} shows that $n_0 := \bigl\lceil 4a\log(2ab) + 2c \bigr\rceil$ suffices, with $d_{\Rreach}$ and $d_{\Hfamily}$ entering linearly while $C_{\Rreach}, C_{\Hfamily}$ appear only inside $\log(2ab)$, giving $n_0 = \widetilde O\bigl((d_{\Rreach} + d_{\Hfamily} + \log(1/\delta))/\varepsilon^2\bigr)$.
Moreover, the explicit formula $n_0 = \bigl\lceil 4a\log(2ab) + 2c \bigr\rceil$ (and $\bigl\lceil 96\log(1/\delta)/\varepsilon^2 \bigr\rceil$ when $d = 0$) depends only on $(d_{\Rreach},\allowbreak d_{\Hfamily},\allowbreak C_{\Rreach},\allowbreak C_{\Hfamily},\allowbreak \varepsilon,\allowbreak \delta)$ and in particular not on $\PT$ or $\eta_n$, establishing agnostic-PAC learnability under \Cref{def:pac}.
\end{proof}

\begin{remark}[Bayes-excess form under polynomial growth]
\label{rem:T1-sample-bayes-excess}
Combining \Cref{cor:T1-sample-app} with \Cref{def:errors} yields $\Risk(\widehat f_n) - \RiskStar \le a_{\mathrm{MR}} + \eta_n + \varepsilon$ with probability at least $1 - \delta$ whenever $\nT \ge n_0$, which is the form invoked when pairing the estimation rate with a population approximation certificate.
\end{remark}

\subsection{Proof of \Cref{prop:no-worse} and restricted head-growth lemma}

\begin{proof}[Proof of \Cref{prop:no-worse}]
Fix an integer $m \ge 1$.
For every target tuple $x_{1:m} \in \XT^m$, \Cref{def:patterns} gives $\ReaClass(x_{1:m}) \subseteq [K_{\rm S}]^m$, so maximizing $|\RelClass(z)|$ over reachable source-label patterns cannot exceed maximizing over $[K_{\rm S}]^m$:
\begin{equation*}
  \sup_{z \in \ReaClass(x_{1:m})} |\RelClass(z)|
  \;\le\; \sup_{z \in [K_{\rm S}]^m} |\RelClass(z)|
  \;=\; \PihatHall(m).
\end{equation*}
Taking the supremum over $x_{1:m} \in \XT^m$ gives $\PihatH(m) \le \PihatHall(m)$.
Combining \Cref{prop:factorization} with $\PihatR(m) \ge 0$ yields
\begin{equation*}
  \Pi_{\FMR}(m) \;\le\; \PihatR(m)\,\PihatH(m) \;\le\; \PihatR(m)\,\PihatHall(m). \qedhere
\end{equation*}
\end{proof}

We conclude this section with the reachable-alphabet Sauer--Shelah bound on $\PihatH(m)$, which restricts output-side counting to the set of source labels realized on a target sample and supplies the head-capacity bound invoked in \Cref{app:proof_T1P}.
Recall from \Cref{app:assumptions} that for $z = (z_1, \ldots, z_m) \in [K_{\rm S}]^m$, $\reachset(z) = \{z_1, \ldots, z_m\} \subseteq [K_{\rm S}]$ denotes the reached source-label set of cardinality $|z|_{\#} = |\reachset(z)|$, with sample-wise maxima $\tR(m) = \sup_{x_{1:m}\in\XT^m}\sup_{z\in\ReaClass(x_{1:m})} |\reachset(z)|$ and $\dHR(m) = \sup_{x_{1:m}\in\XT^m}\sup_{z\in\ReaClass(x_{1:m})} \Ndim(\Hfamily|_{\reachset(z)})$.

\begin{lemma}[Restricted head growth on the reachable source alphabet]
\label{lem:restricted-head}
Under the wrapper setup of \Cref{thm:T1-app}, for every integer $m \ge 1$, the conditional-relabeling growth satisfies
\begin{equation}
\label{eq:restricted-head}
  \PihatH(m) \;\le\; \sum_{i=0}^{\dHR(m) \wedge \tR(m)} \binom{\tR(m)}{i}\binom{K_{\rm T}}{2}^{i},
\end{equation}
with the convention $\binom{K_{\rm T}}{2}^0 := 1$ when $K_{\rm T} = 1$.
\end{lemma}

\begin{proof}
Fix an integer $m \ge 1$.
If $K_{\rm T} = 1$, then every $h \in \Hfamily \subseteq \{1\}^{[K_{\rm S}]}$ is identically $1$, so $|\RelClass(z)| \le 1$ on every $z \in \ReaClass(x_{1:m})$ and hence $\PihatH(m) \le 1$.
Moreover, no coordinate admits two distinct target labels when $K_{\rm T} = 1$, giving $\dHR(m) = 0$ and $\binom{\tR(m)}{0}\binom{1}{2}^0 = 1$ under $\binom{K_{\rm T}}{2}^0 := 1$, which establishes \eqref{eq:restricted-head}.

Assume henceforth that $K_{\rm T} \ge 2$.
Fix any target tuple $x_{1:m} \in \XT^m$ and any reachable source-label vector $z = (z_1, \ldots, z_m) \in \ReaClass(x_{1:m})$, and set $A_z := \reachset(z) = \{z_1, \ldots, z_m\} \subseteq [K_{\rm S}]$ and $d_z := \Ndim(\Hfamily|_{A_z})$.
Two maps $h, h' \in \Hfamily$ satisfy $(h(z_1), \ldots, h(z_m)) = (h'(z_1), \ldots, h'(z_m))$ if and only if $h(k) = h'(k)$ for every $k \in A_z$, because $A_z$ consists of the exact coordinates appearing in $z$.
Thus the evaluation map $h|_{A_z} \mapsto (h(z_1), \ldots, h(z_m))$ is a bijection from $\Hfamily|_{A_z} := \{h|_{A_z} : h \in \Hfamily\}$ onto $\RelClass(z)$, giving $|\RelClass(z)| = |\Hfamily|_{A_z}|$.
Applying the multivalued Sauer--Shelah lemma~\citep[Theorem~9]{ben1992characterizations} to $\Hfamily|_{A_z} \subseteq [K_{\rm T}]^{A_z}$ on the finite domain $A_z$, together with $|A_z| \le \tR(m)$ and $d_z \le \dHR(m)$, yields
\begin{align*}
  |\RelClass(z)|
  &= |\Hfamily|_{A_z}|
    &&\text{(restriction bijection)} \\
  &\le \sum_{i=0}^{d_z \wedge |A_z|} \binom{|A_z|}{i}\binom{K_{\rm T}}{2}^{i}
    &&\text{(multivalued Sauer--Shelah)} \\
  &\le \sum_{i=0}^{d_z \wedge |A_z|} \binom{\tR(m)}{i}\binom{K_{\rm T}}{2}^{i}
    &&\text{(since $|A_z| \le \tR(m)$)} \\
  &\le \sum_{i=0}^{\dHR(m) \wedge \tR(m)} \binom{\tR(m)}{i}\binom{K_{\rm T}}{2}^{i},
    &&\text{(since $d_z \le \dHR(m)$)}
\end{align*}
where the last inequality extends a sum of nonnegative terms from $d_z \wedge |A_z|$ to $\dHR(m) \wedge \tR(m)$.
Taking the supremum over $z \in \ReaClass(x_{1:m})$ and then over $x_{1:m} \in \XT^m$ proves \eqref{eq:restricted-head}.
\end{proof}

\begin{remark}[Recovery of the coarse Natarajan head bound]
\label{rem:restricted-head-recovery}
Because every $z \in [K_{\rm S}]^m$ contains at most $m$ distinct symbols and domain restriction cannot increase Natarajan dimension, we always have $\tR(m) \le m$ and $\dHR(m) \le \Ndim(\Hfamily)$.
Whenever $\Ndim(\Hfamily) \le d_{\Hfamily}$, integrality of $\Ndim(\Hfamily) \in \mathbb{N}_0$ gives $\dHR(m) \le \lfloor d_{\Hfamily} \rfloor$, and \Cref{lem:restricted-head} therefore implies
\begin{equation*}
  \PihatH(m) \;\le\; \sum_{i=0}^{\lfloor d_{\Hfamily} \rfloor \wedge m} \binom{m}{i}\binom{K_{\rm T}}{2}^{i},
\end{equation*}
which is the source-free output-capacity bound used in \Cref{thm:T1P-app}.
\end{remark}

%% file: appendix/proof_T1P_C1prime.tex
\label{app:proof_T1P}

\textbf{Proof roadmap.}
This section proves the explicit PAC rate and sample complexity for additive input transformations (\Cref{thm:T1P,cor:T1P-sample}) and discharges the reachability-arrangement condition (\Cref{cond:arrangement}) for affine and fixed finite-depth ReLU source-score maps (\Cref{cor:T1P:affine,cor:T1P:relu}).
In \Cref{app:proof_T1P:main}, \Cref{thm:T1P-app} combines the cell-partition bound of \Cref{cond:arrangement} on $\PihatR(m)$ with the restricted Sauer--Shelah bound of \Cref{lem:restricted-head} (\Cref{rem:restricted-head-recovery}) under \Cref{cond:nat} on $\PihatH(m)$ and substitutes both into \Cref{thm:T1-app}, after which \Cref{cor:T1-sample-app} yields \Cref{cor:T1P-sample}.
In \Cref{app:proof_T1P:discharges}, we establish the affine hyperplane weak-sign-pattern recursion on $\mathbb{R}^p$ (\Cref{lem:weak-sign}), apply it with the lexicographic weak-sign characterization \eqref{eq:lexargmax-weak-sign} to the $m\binom{K_{\rm S}}{2}$ pairwise score differences of an affine score map (\Cref{cor:C1prime-app}, proving \Cref{cor:T1P:affine}), and propagate the refinement inductively across $L$ hidden ReLU layers before the output-layer score comparisons (\Cref{cor:relu-discharge-app}, proving \Cref{cor:T1P:relu}).

\subsection{Proof of \Cref{thm:T1P} and \Cref{cor:T1P-sample}}
\label{app:proof_T1P:main}

\begin{theorem}[Explicit PAC bound for additive MR; restatement of \Cref{thm:T1P}]
\label{thm:T1P-app}
For $\XT \subseteq \mathbb{R}^d$, $\XS = \mathbb{R}^d$, and the additive input-transformation class $\Gfamily_p = \{g_\theta(x) = x + M\theta : \theta \in \Theta \subseteq \mathbb{R}^p\}$ with fixed $M \in \mathbb{R}^{d \times p}$, assume \Cref{cond:arrangement,cond:nat} and set $d_m := d_{\Hfamily} \wedge m$ (with $B_m := 0$ when $d_m = 0$ or $K_{\rm T} = 1$).
For every integer $m \ge 1$,
\begin{equation*}
  \log\PihatR(m) \;\le\; \underbrace{c_{\Rreach}\,p\log(C_{\Rreach}\,m)}_{=:\,A_m},
  \qquad
  \log\PihatH(m) \;\le\; \underbrace{d_m\log\!\left(\frac{e\binom{K_{\rm T}}{2}m}{d_m}\right)}_{=:\,B_m}.
\end{equation*}
Consequently, for every target distribution $\PT$, every $\nT \ge 1$, every $\delta \in (0,1)$, and every $\eta_n \ge 0$, with probability at least $1 - \delta$ over $\PT^{\nT}$, every measurable $\eta_n$-approximate ERM $\widehat f_n \in \FMR$ satisfies
\begin{equation*}
  \EstRisk(\widehat f_n) \;\le\; \eta_n + 4\sqrt{6}\,\sqrt{\frac{A_{2\nT} + B_{2\nT} + \log(1/\delta)}{\nT}}.
\end{equation*}
\end{theorem}

\begin{proof}
Fix an integer $m \ge 1$.

\textbf{Bounding $\log\PihatR(m)$ ($A_m$).}
Fix an arbitrary target tuple $x_{1:m} \in \XT^m$.
For each parameter $\theta \in \Theta$, write the induced source-label vector as
\begin{equation*}
  z_\theta(x_{1:m}) \;:=\; \bigl(\fs(g_\theta(x_1)), \ldots, \fs(g_\theta(x_m))\bigr) \;\in\; [K_{\rm S}]^m,
\end{equation*}
so that $\ReaClass(x_{1:m}) = \{z_\theta(x_{1:m}) : \theta \in \Theta\}$.
By \Cref{cond:arrangement}, $\Theta$ admits a partition into at most $(C_{\Rreach}\,m)^{c_{\Rreach}\,p}$ cells on each of which $\theta \mapsto z_\theta(x_{1:m})$ is constant.
Because at most one source-label vector is realized per cell, $|\ReaClass(x_{1:m})| \le (C_{\Rreach}\,m)^{c_{\Rreach}\,p}$.
Taking the supremum over $x_{1:m} \in \XT^m$ gives $\PihatR(m) \le (C_{\Rreach}\,m)^{c_{\Rreach}\,p}$, and taking logarithms yields $\log\PihatR(m) \le c_{\Rreach}\,p\log(C_{\Rreach}\,m) = A_m$.

\textbf{Bounding $\log\PihatH(m)$ ($B_m$).}
Under \Cref{cond:nat}, $\Ndim(\Hfamily) \le d_{\Hfamily} < \infty$, and since $\Ndim(\Hfamily) \in \mathbb{N}_0$, we have $d_{\Hfamily} \ge 0$ and $\dHR(m) \le \Ndim(\Hfamily) \le \lfloor d_{\Hfamily} \rfloor$, while $\tR(m) \le m$.
If $K_{\rm T} = 1$ or $d_m = 0$, then \Cref{lem:restricted-head} (\Cref{rem:restricted-head-recovery}) gives $\PihatH(m) = 1$, so $\log\PihatH(m) = 0 = B_m$.
If $K_{\rm T} \ge 2$ and $0 < d_m < 1$, then $m \ge 1$ forces $0 < d_{\Hfamily} < 1$ and hence $\Ndim(\Hfamily) \le \lfloor d_{\Hfamily} \rfloor = 0$, so \Cref{lem:restricted-head} gives $\PihatH(m) = 1$ and $\log\PihatH(m) = 0 < d_m\log\bigl(e\binom{K_{\rm T}}{2}m/d_m\bigr) = B_m$ (since $e\binom{K_{\rm T}}{2}m/d_m > e > 1$).
Assume henceforth that $K_{\rm T} \ge 2$ and $1 \le d_m \le m$, and set $k_m := \lfloor d_m \rfloor \in \{1, \ldots, m\}$.
Then $\dHR(m) \wedge m \le \lfloor d_{\Hfamily} \rfloor \wedge m = k_m \le d_m$, while $\binom{K_{\rm T}}{2} \ge 1$ gives $\binom{K_{\rm T}}{2}^i \le \binom{K_{\rm T}}{2}^{d_m}$ and $m \ge d_m$ gives $(m/d_m)^{d_m}(d_m/m)^i = (m/d_m)^{d_m - i} \ge 1$ for every $0 \le i \le k_m$.
Starting from the restricted Sauer--Shelah bound (\Cref{lem:restricted-head,rem:restricted-head-recovery}), we obtain
\begin{align*}
  \PihatH(m)
  &\le \sum_{i=0}^{k_m} \binom{m}{i}\binom{K_{\rm T}}{2}^{i} \\
  &\le \binom{K_{\rm T}}{2}^{d_m} \sum_{i=0}^{k_m} \binom{m}{i}
    &&\text{($\tbinom{K_{\rm T}}{2} \ge 1,\; i \le d_m$)} \\
  &\le \binom{K_{\rm T}}{2}^{d_m} \left(\frac{m}{d_m}\right)^{d_m} \sum_{i=0}^{k_m} \binom{m}{i}\left(\frac{d_m}{m}\right)^{i}
    &&\text{($(m/d_m)^{d_m - i} \ge 1$)} \\
  &\le \binom{K_{\rm T}}{2}^{d_m} \left(\frac{m}{d_m}\right)^{d_m} \left(1 + \frac{d_m}{m}\right)^{m}
    &&\text{(extending sum to $m$)} \\
  &\le \left(\frac{e\binom{K_{\rm T}}{2}m}{d_m}\right)^{d_m},
    &&\text{($1 + u \le e^u$)}
\end{align*}
and taking logarithms yields $\log\PihatH(m) \le d_m\log\bigl(e\binom{K_{\rm T}}{2}m/d_m\bigr) = B_m$.

Evaluating both growth bounds at $m = 2\nT$ gives $\log\PihatR(2\nT) \le A_{2\nT}$ and $\log\PihatH(2\nT) \le B_{2\nT}$.
Since $u \mapsto \sqrt{u}$ is nondecreasing on $[0, \infty)$, substituting both inequalities into \Cref{thm:T1-app} yields the stated finite-sample bound on $\EstRisk(\widehat f_n)$.
\end{proof}

\begin{proof}[Proof of \Cref{cor:T1P-sample}]
Under \Cref{cond:arrangement,cond:nat}, \Cref{thm:T1P-app} gives $\log\PihatR(m) \le A_m = (c_{\Rreach}\,p)\log(C_{\Rreach}\,m)$ for every integer $m \ge 1$, with $c_{\Rreach}\,p \ge 0$ and $C_{\Rreach} \ge 1$.
Define $C_{\Hfamily} := \max\bigl\{1,\, e\binom{K_{\rm T}}{2}\bigr\} \ge 1$.
If $K_{\rm T} = 1$ or $0 \le d_{\Hfamily} < 1$, then $\Ndim(\Hfamily) \in \mathbb{N}_0$ forces $\Ndim(\Hfamily) = 0$ whenever $0 \le d_{\Hfamily} < 1$, so \Cref{lem:restricted-head} gives $\PihatH(m) = 1$ and $\log\PihatH(m) = 0 \le d_{\Hfamily}\log(C_{\Hfamily}\,m)$ for every $m \ge 1$ (since $d_{\Hfamily} \ge 0$ and $C_{\Hfamily}\,m \ge 1$).
Otherwise, $d_{\Hfamily} \ge 1$ and $K_{\rm T} \ge 2$, so for every $m \ge 1$ we have $1 \le d_m \le d_{\Hfamily}$, $C_{\Hfamily} = e\binom{K_{\rm T}}{2}$, and $\log\bigl(e\binom{K_{\rm T}}{2}m\bigr) \ge 1 > 0$; \Cref{thm:T1P-app} then yields
\begin{align*}
  \log\PihatH(m)
  &\le B_m
    \;=\; d_m\log\!\left(\frac{e\binom{K_{\rm T}}{2}m}{d_m}\right) \\
  &\le d_m\log\!\left(e\binom{K_{\rm T}}{2}m\right)
    &&\text{($d_m \ge 1$)} \\
  &\le d_{\Hfamily}\log\!\left(e\binom{K_{\rm T}}{2}m\right)
    \;=\; d_{\Hfamily}\log(C_{\Hfamily}\,m).
    &&\text{($d_m \le d_{\Hfamily}$)}
\end{align*}
In both cases, the polynomial-growth hypotheses of \Cref{cor:T1-sample-app} hold with exponents $(c_{\Rreach}\,p, d_{\Hfamily})$ and scale constants $(C_{\Rreach}, C_{\Hfamily})$.
Invoking \Cref{cor:T1-sample-app} gives $n_0 = \widetilde O\bigl((c_{\Rreach}\,p + d_{\Hfamily} + \log(1/\delta))/\varepsilon^2\bigr)$ such that whenever $\nT \ge n_0$, with probability at least $1 - \delta$ over $\PT^{\nT}$, every measurable $\eta_n$-approximate ERM $\widehat f_n \in \FMR$ satisfies $\EstRisk(\widehat f_n) \le \eta_n + \varepsilon$, and hence $\Risk(\widehat f_n) - \RiskStar = \EstRisk(\widehat f_n) + a_{\mathrm{MR}} \le a_{\mathrm{MR}} + \eta_n + \varepsilon$ (\Cref{def:errors}).
\end{proof}

\subsection{Weak affine sign-pattern bound and proofs of \Cref{cor:T1P:affine,cor:T1P:relu}}
\label{app:proof_T1P:discharges}

\begin{lemma}[Weak affine sign-pattern count]
\label{lem:weak-sign}
For $p, N \in \mathbb{N}_0$, define
\begin{equation*}
  \Phip(N) \;:=\; \sum_{j=0}^{\min\{p,N\}} 2^j \binom{N}{j}, \qquad \Phip(0) \;=\; 1.
\end{equation*}
Let $\ell_1, \ldots, \ell_N : \mathbb{R}^p \to \mathbb{R}$ be affine functions, and define the three-valued weak sign $\operatorname{sgn} : \mathbb{R} \to \{-1, 0, +1\}$ by
\begin{equation*}
\operatorname{sgn}(t) \;:=\;
\begin{cases}
+1, & t > 0,\\
0,  & t = 0,\\
-1, & t < 0.
\end{cases}
\end{equation*}
Then for every subset $\Theta \subseteq \mathbb{R}^p$, the number of distinct weak sign vectors $\bigl(\operatorname{sgn}(\ell_1(\theta)), \ldots, \operatorname{sgn}(\ell_N(\theta))\bigr)$ realized as $\theta$ ranges over $\Theta$ is at most $\Phip(N)$.
Moreover, $\Phip(N) \le (1 + 2N)^p$ for every $p \ge 1$, while $\Phi_0(N) = 1$, so $\log\Phip(N) \le p\log(1 + 2N)$ for all $p, N \in \mathbb{N}_0$.
\end{lemma}

\begin{proof}
Because restricting $\theta$ from $\mathbb{R}^p$ to $\Theta \subseteq \mathbb{R}^p$ can only decrease the set of realized sign vectors, it suffices to take $\Theta = \mathbb{R}^p$.
For each $\sigma \in \{-1, 0, +1\}^N$, define its weak sign cell
\begin{equation*}
  C_\sigma \;:=\; \bigl\{\theta \in \mathbb{R}^p : \operatorname{sgn}(\ell_j(\theta)) = \sigma_j \text{ for all } j \in [N]\bigr\}.
\end{equation*}
The cells $\{C_\sigma : \sigma \in \{-1, 0, +1\}^N\}$ form a pairwise disjoint partition of $\mathbb{R}^p$, and each nonempty $C_\sigma$ is convex as an intersection of open affine halfspaces ($\sigma_j = \pm 1$) and affine hyperplanes ($\sigma_j = 0$).
For any $N$-tuple $(\ell_1, \ldots, \ell_N)$ of affine functions on $\mathbb{R}^p$, let $C(\ell_1, \ldots, \ell_N) := \bigl|\{\sigma \in \{-1, 0, +1\}^N : C_\sigma \neq \varnothing\}\bigr|$ denote the number of nonempty weak sign cells it generates on $\mathbb{R}^p$, and define the worst-case cell count
\begin{equation*}
  F_p(N) \;:=\; \sup_{(\ell'_1, \ldots, \ell'_N)} C(\ell'_1, \ldots, \ell'_N),
\end{equation*}
where the supremum (attained as a maximum since $C(\ell'_1, \ldots, \ell'_N) \le 3^N$) ranges over all $N$-tuples of affine functions on $\mathbb{R}^p$.
In the boundary cases, $F_p(0) = 1 = \Phip(0)$ (the single empty tuple) and $F_0(N) = 1 = \Phi_0(N)$ (since $\mathbb{R}^0$ is a singleton).

Now assume $p, N \ge 1$, fix an arbitrary $N$-tuple $(\ell_1, \ldots, \ell_N)$ of affine functions on $\mathbb{R}^p$, and consider adding $\ell_N$ to $(\ell_1, \ldots, \ell_{N-1})$.
If $\ell_N$ is constant on $\mathbb{R}^p$, its weak sign takes a single value everywhere, so each nonempty cell of $(\ell_1, \ldots, \ell_{N-1})$ produces a single nonempty cell of $(\ell_1, \ldots, \ell_N)$ and
\begin{equation*}
  C(\ell_1, \ldots, \ell_N) \;=\; C(\ell_1, \ldots, \ell_{N-1}) \;\le\; F_p(N-1) \;\le\; F_p(N-1) + 2F_{p-1}(N-1).
\end{equation*}
Otherwise, $\ell_N$ is nonconstant, and its zero set $H := \{\theta \in \mathbb{R}^p : \ell_N(\theta) = 0\}$ is an affine hyperplane isomorphic to $\mathbb{R}^{p-1}$.
Each nonempty cell $C_\tau$ ($\tau \in \{-1, 0, +1\}^{N-1}$) of $(\ell_1, \ldots, \ell_{N-1})$ falls into one of three mutually exclusive cases relative to $H$:
\begin{enumerate}
  \item \emph{Non-intersecting ($C_\tau \cap H = \varnothing$).} Because $C_\tau$ is convex (hence connected) and $\ell_N$ is continuous and nonzero on $C_\tau$, the Intermediate Value Theorem forces $\operatorname{sgn}(\ell_N(\theta))$ to take a single value ($+1$ or $-1$) throughout $C_\tau$; thus $C_\tau$ contributes $1$ nonempty cell after $\ell_N$ is added.
  \item \emph{Contained ($C_\tau \subseteq H$).} Here $\operatorname{sgn}(\ell_N(\theta)) = 0$ on $C_\tau$, so $C_\tau$ again contributes $1$ nonempty cell.
  \item \emph{Crossing ($C_\tau \cap H \neq \varnothing$ and $C_\tau \not\subseteq H$).} Splitting $C_\tau$ according to $\operatorname{sgn}(\ell_N(\theta)) \in \{-1, 0, +1\}$ partitions $C_\tau$ into at most $3$ nonempty subcells (at most one each for $\ell_N > 0$, $\ell_N = 0$, and $\ell_N < 0$), adding at most $2$ nonempty cells beyond the $1$ counted in $C(\ell_1, \ldots, \ell_{N-1})$.
\end{enumerate}
Let $\mathcal T_{\mathrm{cross}}$ denote the set of indices $\tau$ in the crossing case.
For every $\tau \in \mathcal T_{\mathrm{cross}}$, the trace $C_\tau \cap H$ is nonempty and equals the weak sign cell $\{\theta \in H : \operatorname{sgn}(\ell_j|_H(\theta)) = \tau_j \text{ for all } j \in [N-1]\}$ of the $(N-1)$ restricted affine functions $\ell_1|_H, \ldots, \ell_{N-1}|_H$ on $H$.
Moreover, the map $\tau \mapsto C_\tau \cap H$ on $\mathcal T_{\mathrm{cross}}$ is injective: if $\theta_0 \in (C_\tau \cap H) \cap (C_{\tau'} \cap H)$, then $\tau_j = \operatorname{sgn}(\ell_j(\theta_0)) = \tau'_j$ for every $j \in [N-1]$, so $\tau = \tau'$.
Identifying $H$ affinely with $\mathbb{R}^{p-1}$ bounds $|\mathcal T_{\mathrm{cross}}| \le F_{p-1}(N-1)$, and summing over all nonempty cells of $(\ell_1, \ldots, \ell_{N-1})$ gives
\begin{equation*}
  C(\ell_1, \ldots, \ell_N) \;\le\; C(\ell_1, \ldots, \ell_{N-1}) + 2|\mathcal T_{\mathrm{cross}}| \;\le\; F_p(N-1) + 2F_{p-1}(N-1).
\end{equation*}
In both cases, $C(\ell_1, \ldots, \ell_N) \le F_p(N-1) + 2F_{p-1}(N-1)$; taking the supremum over all $N$-tuples $(\ell_1, \ldots, \ell_N)$ of affine functions on $\mathbb{R}^p$ yields the recursion
\begin{equation*}
  F_p(N) \;\le\; F_p(N-1) + 2F_{p-1}(N-1).
\end{equation*}

For all $p, N \ge 1$, Pascal's identity $\binom{N-1}{j} + \binom{N-1}{j-1} = \binom{N}{j}$ shows that $\Phip$ satisfies the matching recursion with equality:
\begin{align*}
  \Phip(N-1) + 2\Phi_{p-1}(N-1)
  &= \sum_{j=0}^{\min\{p,N-1\}} 2^j \binom{N-1}{j} + \sum_{j=1}^{\min\{p,N\}} 2^j \binom{N-1}{j-1} \\
  &= \sum_{j=0}^{\min\{p,N\}} 2^j \binom{N}{j} \;=\; \Phip(N),
\end{align*}
where when $N - 1 < p$, the $j = N$ summand in the first sum vanishes via $\binom{N-1}{N} = 0$.
Since $F_p(N) = \Phip(N) = 1$ on the boundary $p = 0$ or $N = 0$, induction on $p + N$ gives $C(\ell_1, \ldots, \ell_N) \le F_p(N) \le \Phip(N)$ for all $p, N \in \mathbb{N}_0$.

Finally, if $p = 0$, then $\Phi_0(N) = 1$ and $\log\Phi_0(N) = 0$.
For $p \ge 1$, if $N = 0$ then $\Phip(0) = 1 = (1 + 2\cdot 0)^p$; if $N \ge 1$, then for every $0 \le j \le \min\{p, N\}$,
\begin{align*}
  \binom{N}{j}
  &= \frac{1}{j!}\prod_{r=0}^{j-1}(N - r) \\
  &\le \frac{N^j}{j!}
    &&\text{($0 < N - r \le N$)} \\
  &\le N^j
    &&\text{($j! \ge 1$)} \\
  &\le \binom{p}{j} N^j,
    &&\text{($\tbinom{p}{j} \ge 1,\; j \le p$)}
\end{align*}
with the empty product at $j = 0$ equal to $1$.
Summing over $j$ gives
\begin{align*}
  \Phip(N)
  &= \sum_{j=0}^{\min\{p,N\}} 2^j \binom{N}{j} \\
  &\le \sum_{j=0}^{\min\{p,N\}} \binom{p}{j} (2N)^j \\
  &\le \sum_{j=0}^{p} \binom{p}{j} (2N)^j
    &&\text{(extending sum to $p$)} \\
  &= (1 + 2N)^p,
    &&\text{(binomial theorem)}
\end{align*}
and taking logarithms yields $\log\Phip(N) \le p\log(1 + 2N)$.
\end{proof}

\begin{corollary}[Affine source-score maps; restatement of \Cref{cor:T1P:affine}]
\label{cor:C1prime-app}
For $\Gfamily_p$, if $\fs(u) = \lexargmax_{k \in [K_{\rm S}]} s_{{\rm S},k}(u)$ with $s_{\rm S}(u) = Au + b$ ($A \in \mathbb{R}^{K_{\rm S} \times d}$, $b \in \mathbb{R}^{K_{\rm S}}$), then for every $m \ge 1$,
\begin{equation*}
  \log \PihatR(m) \;\le\; \underbrace{\log \Phip\!\bigl(m \binom{K_{\rm S}}{2}\bigr)}_{=:\,\CAff(m)} \;\le\; p \log \bigl(1 + 2 m \binom{K_{\rm S}}{2}\bigr).
\end{equation*}
\end{corollary}

\begin{proof}
Under deterministic lexicographic tie-breaking $\lexargmax_{j \in [K_{\rm S}]} t_j = \min\arg\max_{j \in [K_{\rm S}]} t_j$, any score vector $t = (t_1, \ldots, t_{K_{\rm S}}) \in \mathbb{R}^{K_{\rm S}}$ ($K_{\rm S} \ge 1$) selects class $k \in [K_{\rm S}]$ if and only if $t_k$ strictly exceeds every lower-indexed coordinate and weakly exceeds every higher-indexed coordinate:
\begin{equation}\label{eq:lexargmax-weak-sign}
  \lexargmax_{j \in [K_{\rm S}]} t_j = k
  \quad\Longleftrightarrow\quad
  \begin{cases}
    \operatorname{sgn}(t_r - t_k) = -1 & \text{for all } 1 \le r < k,\\
    \operatorname{sgn}(t_k - t_\ell) \in \{0, +1\} & \text{for all } k < \ell \le K_{\rm S},
  \end{cases}
\end{equation}
where both index ranges are vacuous when $K_{\rm S} = 1$.

Fix $m \ge 1$ and an arbitrary target tuple $x_{1:m} \in \XT^m$.
If $K_{\rm S} = 1$, then $\binom{K_{\rm S}}{2} = 0$ and $\fs \equiv 1$, so $\Theta$ itself forms a single constant-label cell, $|\ReaClass(x_{1:m})| \le 1 = \Phip(0)$, $\PihatR(m) \le 1 = \Phip(0)$, and $\log\PihatR(m) \le 0 = \CAff(m) = p\log(1 + 0)$ for all $p \in \mathbb{N}_0$.
Assume henceforth $K_{\rm S} \ge 2$, and denote the $k$-th row of $A$ by $A_k^\top \in \mathbb{R}^{1 \times d}$.
For each sample index $i \in [m]$ and ordered label pair $1 \le k < \ell \le K_{\rm S}$, the pairwise score difference
\begin{equation*}
  \delta_{i,k\ell}(\theta)
  \;:=\; s_{{\rm S},k}\bigl(g_\theta(x_i)\bigr) - s_{{\rm S},\ell}\bigl(g_\theta(x_i)\bigr)
  \;=\; (A_k - A_\ell)^\top M\theta + (A_k - A_\ell)^\top x_i + (b_k - b_\ell)
\end{equation*}
is an affine function of $\theta \in \mathbb{R}^p$.
Applying \Cref{lem:weak-sign} to the $N := m\binom{K_{\rm S}}{2}$ affine functions $\{\delta_{i,k\ell} : i \in [m],\, 1 \le k < \ell \le K_{\rm S}\}$ partitions $\Theta$ into at most $\Phip\bigl(m\binom{K_{\rm S}}{2}\bigr)$ nonempty weak sign cells on each of which every $\operatorname{sgn}(\delta_{i,k\ell}(\theta))$ is constant.
Applying \eqref{eq:lexargmax-weak-sign} to $t = s_{\rm S}(g_\theta(x_i))$ shows that the source-label vector $\bigl(\fs(g_\theta(x_1)), \ldots, \fs(g_\theta(x_m))\bigr)$ is constant on each cell, discharging \Cref{cond:arrangement} with $c_{\Rreach} = 1$ and $C_{\Rreach} = 1 + 2\binom{K_{\rm S}}{2}$ (since $1 + 2m\binom{K_{\rm S}}{2} \le \bigl(1 + 2\binom{K_{\rm S}}{2}\bigr)m$ for $m \ge 1$).
Hence $|\ReaClass(x_{1:m})| \le \Phip\bigl(m\binom{K_{\rm S}}{2}\bigr)$, and taking the supremum over $x_{1:m} \in \XT^m$ gives $\PihatR(m) \le \Phip\bigl(m\binom{K_{\rm S}}{2}\bigr)$; taking logarithms and applying $\log\Phip(N) \le p\log(1 + 2N)$ (\Cref{lem:weak-sign}) for all $p \in \mathbb{N}_0$ yields $\log\PihatR(m) \le \CAff(m) \le p\log\bigl(1 + 2m\binom{K_{\rm S}}{2}\bigr)$.
\end{proof}

For the fixed-depth ReLU discharge (\Cref{cor:T1P:relu}), consider a fixed feedforward ReLU source-score network $s_{\rm S} : \mathbb{R}^d \to \mathbb{R}^{K_{\rm S}}$ with $L \ge 1$ hidden layers of widths $q_1, \ldots, q_L \ge 1$, defined recursively for $u \in \mathbb{R}^d$ by
\begin{equation*}
  z_0(u) \;:=\; u,
  \qquad
  a_\ell(u) \;:=\; A_\ell z_{\ell-1}(u) + b_\ell,
  \qquad
  z_\ell(u) \;:=\; \sigma\bigl(a_\ell(u)\bigr) \quad (\ell = 1, \ldots, L),
\end{equation*}
where $\sigma(t) := \max\{t, 0\}$ acts coordinatewise, layer $\ell$ has fixed weights $A_\ell \in \mathbb{R}^{q_\ell \times q_{\ell-1}}$ ($q_0 := d$) and biases $b_\ell \in \mathbb{R}^{q_\ell}$, and the output score map is $s_{\rm S}(u) := A_{L+1} z_L(u) + b_{L+1}$ with $A_{L+1} \in \mathbb{R}^{K_{\rm S} \times q_L}$ and $b_{L+1} \in \mathbb{R}^{K_{\rm S}}$ ($K_{\rm S} \ge 1$).

\begin{corollary}[Fixed finite-depth ReLU source-score networks; restatement of \Cref{cor:T1P:relu}]
\label{cor:relu-discharge-app}
For $\Gfamily_p$, if $\fs(u) = \lexargmax_{k \in [K_{\rm S}]} s_{{\rm S},k}(u)$ where $s_{\rm S}$ is a fixed feedforward ReLU network with $L \ge 1$ hidden layers of widths $q_1, \ldots, q_L$, then for every $m \ge 1$,
\begin{equation*}
  \log \PihatR(m) \;\le\; \underbrace{\log \Phip\!\bigl(m \binom{K_{\rm S}}{2}\bigr) + \sum_{\ell=1}^L \log \Phip(m\, q_\ell)}_{=:\,\CRelu(m)} \;=\; O\!\bigl(p(L+1) \log m\bigr).
\end{equation*}
\end{corollary}

\begin{proof}
Fix $m \ge 1$ and an arbitrary target tuple $x_{1:m} \in \XT^m$, and write $u_i(\theta) := g_\theta(x_i) = x_i + M\theta$ for $i \in [m]$.
We construct finite partitions $\mathcal P_0, \mathcal P_1, \ldots, \mathcal P_L$ of $\Theta$ by induction on $\ell \in \{0, \ldots, L\}$ maintaining two invariants:
(a) $|\mathcal P_\ell| \le \prod_{s=1}^{\ell} \Phip(m\,q_s)$ (with the empty product equal to $1$), and
(b) on every cell $C \in \mathcal P_\ell$ and for every $i \in [m]$, the hidden representation $\theta \mapsto z_\ell(u_i(\theta)) \in \mathbb{R}^{q_\ell}$ is the restriction to $C$ of an affine map on $\mathbb{R}^p$.

\textbf{Base case ($\ell = 0$).}
Set $\mathcal P_0 := \{\Theta\}$ (assuming $\Theta \neq \varnothing$; otherwise $\PihatR(m) = 0$ trivially).
Then $|\mathcal P_0| = 1$, matching the empty product in (a), and $z_0(u_i(\theta)) = x_i + M\theta$ is globally affine on $\mathbb{R}^p$ for every $i \in [m]$, establishing (b).

\textbf{Inductive step ($\ell - 1 \to \ell$).}
For $1 \le \ell \le L$, assume $\mathcal P_{\ell-1}$ satisfies (a) and (b), and fix a cell $C \in \mathcal P_{\ell-1}$.
By invariant (b), for each $i \in [m]$ there exist $U_{\ell-1}^{(i)} \in \mathbb{R}^{q_{\ell-1} \times p}$ and $v_{\ell-1}^{(i)} \in \mathbb{R}^{q_{\ell-1}}$ such that $z_{\ell-1}(u_i(\theta)) = U_{\ell-1}^{(i)}\theta + v_{\ell-1}^{(i)}$ for all $\theta \in C$.
Consequently, each preactivation coordinate $(i, j) \in [m] \times [q_\ell]$ agrees on $C$ with the globally defined affine function
\begin{equation*}
  \widetilde a_{\ell,j}^{(i)}(\theta) \;:=\; (A_\ell)_{j,\cdot}\bigl(U_{\ell-1}^{(i)}\theta + v_{\ell-1}^{(i)}\bigr) + (b_\ell)_j \qquad (\theta \in \mathbb{R}^p).
\end{equation*}
Applying \Cref{lem:weak-sign} to the family $\{\widetilde a_{\ell,j}^{(i)} : i \in [m],\, j \in [q_\ell]\}$ of $m\,q_\ell$ affine functions over $C \subseteq \mathbb{R}^p$ partitions $C$ into at most $\Phip(m\,q_\ell)$ nonempty subcells on each of which every weak sign $\operatorname{sgn}\bigl(\widetilde a_{\ell,j}^{(i)}(\theta)\bigr) \in \{-1, 0, +1\}$ is constant.
On any such subcell $C'$, for every $(i, j) \in [m] \times [q_\ell]$, the post-activation coordinate $z_{\ell,j}(u_i(\theta)) = \max\bigl\{\widetilde a_{\ell,j}^{(i)}(\theta), 0\bigr\}$ equals $\widetilde a_{\ell,j}^{(i)}(\theta)$ throughout $C'$ when the constant sign is $+1$ and equals $0$ throughout $C'$ when the constant sign is $-1$ or $0$; in all three cases it agrees on $C'$ with an affine map on $\mathbb{R}^p$.
Collecting the refined subcells $C'$ across all $C \in \mathcal P_{\ell-1}$ yields a partition $\mathcal P_\ell$ of $\Theta$ satisfying $|\mathcal P_\ell| \le |\mathcal P_{\ell-1}|\,\Phip(m\,q_\ell) \le \prod_{s=1}^{\ell} \Phip(m\,q_s)$, completing the induction.

At the output layer, fix any terminal hidden-layer cell $C \in \mathcal P_L$.
By invariant (b) at $\ell = L$, the top hidden representation $\theta \mapsto z_L(u_i(\theta))$ agrees on $C$ with an affine map on $\mathbb{R}^p$ for every $i \in [m]$, and therefore so do every output score coordinate $s_{{\rm S},k}(u_i(\theta)) = (A_{L+1})_{k,\cdot} z_L(u_i(\theta)) + (b_{L+1})_k$ and every pairwise score difference
\begin{equation*}
  \Delta_{i,kr}(\theta) \;:=\; s_{{\rm S},k}\bigl(u_i(\theta)\bigr) - s_{{\rm S},r}\bigl(u_i(\theta)\bigr), \qquad i \in [m],\; 1 \le k < r \le K_{\rm S}.
\end{equation*}
Applying \Cref{lem:weak-sign} to these $m\binom{K_{\rm S}}{2}$ affine extensions over $C$ (with $\Phip(0) = 1$ when $K_{\rm S} = 1$) refines $C$ into at most $\Phip\bigl(m\binom{K_{\rm S}}{2}\bigr)$ nonempty subcells on each of which all output pairwise weak signs are constant.
By the lexicographic weak-sign characterization \eqref{eq:lexargmax-weak-sign} applied to $t = s_{\rm S}(u_i(\theta))$, fixed pairwise weak signs uniquely fix $\fs(u_i(\theta))$ for every $i \in [m]$.
Taking the union of refined subcells across all $C \in \mathcal P_L$ partitions $\Theta$ into at most
\begin{equation*}
  \Phip\!\left(m\binom{K_{\rm S}}{2}\right) |\mathcal P_L|
  \;\le\; \Phip\!\left(m\binom{K_{\rm S}}{2}\right) \prod_{\ell=1}^{L} \Phip(m\,q_\ell)
\end{equation*}
constant-source-label cells, so $|\ReaClass(x_{1:m})|$ satisfies the same bound and taking the supremum over $x_{1:m} \in \XT^m$ gives $\PihatR(m) \le \Phip\bigl(m\binom{K_{\rm S}}{2}\bigr)\prod_{\ell=1}^{L}\Phip(m\,q_\ell)$.
Setting $C_{\Rreach} := 1 + 2\max\bigl\{\binom{K_{\rm S}}{2}, q_1, \ldots, q_L\bigr\}$, taking logarithms, and applying $\log\Phip(N) \le p\log(1 + 2N)$ (\Cref{lem:weak-sign}) yields
\begin{align*}
  \log\PihatR(m)
  &\le \log\Phip\!\left(m\binom{K_{\rm S}}{2}\right) + \sum_{\ell=1}^{L} \log\Phip(m\,q_\ell)
    \;=\; \CRelu(m) \\
  &\le p\log\!\left(1 + 2m\binom{K_{\rm S}}{2}\right) + \sum_{\ell=1}^{L} p\log(1 + 2m\,q_\ell)
    &&\text{(\Cref{lem:weak-sign})} \\
  &\le p(L + 1)\log(C_{\Rreach}\,m)
    \;=\; O\bigl(p(L + 1)\log m\bigr),
    &&\text{($1 + 2mq \le C_{\Rreach}\,m$)}
\end{align*}
which also shows that the cell count satisfies $\Phip\bigl(m\binom{K_{\rm S}}{2}\bigr)\prod_{\ell=1}^{L}\Phip(m\,q_\ell) \le (C_{\Rreach}\,m)^{(L+1)p}$, discharging \Cref{cond:arrangement} with $c_{\Rreach} = L + 1$ and $C_{\Rreach} = 1 + 2\max\bigl\{\binom{K_{\rm S}}{2}, q_1, \ldots, q_L\bigr\}$.
\end{proof}

%% file: appendix/proof_T3P.tex
\label{app:proof_T3P}

\textbf{Proof roadmap.}
This section constructs the explicit binary reprogramming family of \Cref{con:binary-family} and proves the matching minimax estimation lower bound of \Cref{thm:T3P} together with its sample-complexity consequence (\Cref{cor:T3P-sample}).
In \Cref{app:proof_T3P:construction}, we give a single self-contained finite-support construction of $(\XT, \XS, \fs, \Gfamily, \Hfamily, \mathfrak{F}^{\mathrm{bin}}_{d_{\Rreach}, d_{\Hfamily}}(\gamma))$ and verify parameter injectivity, exact reachability and conditional-relabeling growth ($\PihatR(m) = 2^{d_{\Rreach}}$ and $\PihatH(m) = 2^{d_{\Hfamily}}$), full binary hypercube realizability on $x_{1:m}$, and zero approximation error ($a_{\mathrm{MR}} = 0$).
In \Cref{app:proof_T3P:lower}, we first isolate the neighboring product-law KL divergence bound and hypercube Assouad reduction in \Cref{lem:binary-neighbor-kl}, then prove \Cref{thm:T3P-app} (restating \Cref{thm:T3P}) and \Cref{cor:T3P-sample} directly on the construction of \Cref{app:proof_T3P:construction}, and finally record the matching specialization of \Cref{thm:T1} in \Cref{rem:T3P-upper-specialization}.

\subsection{Full construction of the binary MR family (\Cref{con:binary-family})}
\label{app:proof_T3P:construction}

We construct the canonical finite-support objects claimed in \Cref{con:binary-family} and verify each combinatorial and population property explicitly.

\paragraph{Parameters and label-set identification.}
Fix integers $d_{\Rreach}, d_{\Hfamily} \ge 1$, and set $m := d_{\Rreach} + d_{\Hfamily}$, $K_{\rm T} := 2$, and $K_{\rm S} := d_{\Hfamily} + 2$.
For indexing convenience, we identify the binary target label space $[K_{\rm T}]$ with $\{0, 1\}$ and the source label space $[K_{\rm S}]$ with $\{0, 1, \ldots, d_{\Hfamily} + 1\}$ via the shift bijection $y \leftrightarrow y - 1$.

\paragraph{Target and source input spaces.}
Let $\XT := \{x_1, \ldots, x_m\}$ be a set of $m$ distinct points equipped with the discrete $\sigma$-algebra, and denote the canonical support tuple by $x_{1:m} := (x_1, \ldots, x_m) \in \XT^m$.
Let
\begin{equation*}
  \XS \;:=\; \{u^0_j, u^1_j : 1 \le j \le d_{\Rreach}\} \;\cup\; \{v_k : 1 \le k \le d_{\Hfamily}\}
\end{equation*}
be a set of $2d_{\Rreach} + d_{\Hfamily}$ pairwise distinct points, also equipped with the discrete $\sigma$-algebra.
The paired points $(u^0_j, u^1_j)$ carry the $d_{\Rreach}$ input-side reachability bits, while the singletons $v_k$ carry the $d_{\Hfamily}$ output-side relabeling bits.

\paragraph{Frozen source classifier.}
Define $\fs : \XS \to [K_{\rm S}] \cong \{0, 1, \ldots, d_{\Hfamily} + 1\}$ by
\begin{equation*}
  \fs(u^0_j) \;:=\; 0,
  \qquad
  \fs(u^1_j) \;:=\; 1,
  \qquad
  \fs(v_k) \;:=\; k + 1,
  \qquad 1 \le j \le d_{\Rreach},\; 1 \le k \le d_{\Hfamily}.
\end{equation*}
Labels $0$ and $1$ are shared across all $1 \le j \le d_{\Rreach}$, whereas each $v_k$ receives its dedicated private source label $k + 1 \in \{2, \ldots, d_{\Hfamily} + 1\}$.
Measurability is immediate under the discrete $\sigma$-algebra.

\paragraph{Input-transformation and output-mapping classes.}
For each bit vector $\alpha = (\alpha_1, \ldots, \alpha_{d_{\Rreach}}) \in \{0, 1\}^{d_{\Rreach}}$, define $g_\alpha : \XT \to \XS$ by
\begin{equation*}
  g_\alpha(x_j) \;:=\; u^{\alpha_j}_j \quad (1 \le j \le d_{\Rreach}),
  \qquad
  g_\alpha(x_{d_{\Rreach} + k}) \;:=\; v_k \quad (1 \le k \le d_{\Hfamily}),
\end{equation*}
and set $\Gfamily := \{g_\alpha : \alpha \in \{0, 1\}^{d_{\Rreach}}\}$.
For each bit vector $\beta = (\beta_1, \ldots, \beta_{d_{\Hfamily}}) \in \{0, 1\}^{d_{\Hfamily}}$, define $h_\beta : \{0, 1, \ldots, d_{\Hfamily} + 1\} \to \{0, 1\}$ by
\begin{equation*}
  h_\beta(0) \;:=\; 0,
  \qquad
  h_\beta(1) \;:=\; 1,
  \qquad
  h_\beta(k + 1) \;:=\; \beta_k \quad (1 \le k \le d_{\Hfamily}),
\end{equation*}
and set $\Hfamily := \{h_\beta : \beta \in \{0, 1\}^{d_{\Hfamily}}\}$.
Both parameterizations are injective: if $\alpha \neq \alpha'$, then $\alpha_j \neq \alpha'_j$ for some $j \le d_{\Rreach}$ so $g_\alpha(x_j) = u^{\alpha_j}_j \neq u^{\alpha'_j}_j = g_{\alpha'}(x_j)$; if $\beta \neq \beta'$, then $\beta_k \neq \beta'_k$ for some $k \le d_{\Hfamily}$ so $h_\beta(k + 1) = \beta_k \neq \beta'_k = h_{\beta'}(k + 1)$.
Hence $|\Gfamily| = 2^{d_{\Rreach}}$ and $|\Hfamily| = 2^{d_{\Hfamily}}$.

\paragraph{Induced wrapper class and hypercube bijection on $x_{1:m}$.}
Write $f_{\alpha,\beta} := h_\beta \circ \fs \circ g_\alpha$ and $\FMR = \{f_{\alpha,\beta} : \alpha \in \{0,1\}^{d_{\Rreach}},\, \beta \in \{0,1\}^{d_{\Hfamily}}\}$.
On the first $d_{\Rreach}$ support points ($1 \le j \le d_{\Rreach}$),
\begin{equation*}
  f_{\alpha,\beta}(x_j) \;=\; h_\beta\bigl(\fs(u^{\alpha_j}_j)\bigr) \;=\; h_\beta(\alpha_j) \;=\; \alpha_j,
\end{equation*}
because $\alpha_j \in \{0, 1\}$ and $h_\beta$ fixes $0$ and $1$.
On the remaining $d_{\Hfamily}$ support points ($1 \le k \le d_{\Hfamily}$),
\begin{equation*}
  f_{\alpha,\beta}(x_{d_{\Rreach}+k}) \;=\; h_\beta\bigl(\fs(v_k)\bigr) \;=\; h_\beta(k + 1) \;=\; \beta_k.
\end{equation*}
Thus, for every $\omega = (\alpha, \beta) \in \{0, 1\}^{d_{\Rreach}} \times \{0, 1\}^{d_{\Hfamily}} = \{0, 1\}^m$, the wrapper $f_\omega := f_{\alpha,\beta} \in \FMR$ satisfies
\begin{equation*}
  \bigl(f_\omega(x_1), \ldots, f_\omega(x_m)\bigr) \;=\; (\alpha_1, \ldots, \alpha_{d_{\Rreach}}, \beta_1, \ldots, \beta_{d_{\Hfamily}}) \;=\; \omega,
\end{equation*}
establishing a bijection $(\alpha, \beta) \mapsto (f_{\alpha,\beta}(x_1), \ldots, f_{\alpha,\beta}(x_m))$ from $\{0,1\}^{d_{\Rreach}} \times \{0,1\}^{d_{\Hfamily}}$ onto $\{0,1\}^m$.

\paragraph{Exact reachability and conditional-relabeling growth.}
Because $|\ReaClass(x'_{1:m'})| \le |\Rreach| \le |\Gfamily| = 2^{d_{\Rreach}}$ and $|\RelClass(z')| \le |\Hfamily| = 2^{d_{\Hfamily}}$ for every $m' \ge 1$, every $x'_{1:m'} \in \XT^{m'}$, and every $z' \in [K_{\rm S}]^{m'}$, we have
\begin{equation*}
  \PihatR(m') \;\le\; 2^{d_{\Rreach}}
  \qquad\text{and}\qquad
  \PihatH(m') \;\le\; 2^{d_{\Hfamily}}
  \qquad \text{for every } m' \ge 1.
\end{equation*}
On the canonical support $x_{1:m}$ ($m' = m$), each $r_\alpha := \fs \circ g_\alpha$ induces the source-label vector
\begin{equation*}
  z_\alpha \;:=\; \bigl(r_\alpha(x_1), \ldots, r_\alpha(x_m)\bigr) \;=\; (\alpha_1, \ldots, \alpha_{d_{\Rreach}}, 2, 3, \ldots, d_{\Hfamily} + 1).
\end{equation*}
Distinct $\alpha \in \{0,1\}^{d_{\Rreach}}$ yield distinct $z_\alpha$, so $|\ReaClass(x_{1:m})| = 2^{d_{\Rreach}}$ and hence $\PihatR(m) = 2^{d_{\Rreach}}$ (and $\PihatR(m') = 2^{d_{\Rreach}}$ for all $m' \ge d_{\Rreach}$).
Moreover, for every $z_\alpha \in \ReaClass(x_{1:m})$, relabeling $z_\alpha$ by $h_\beta \in \Hfamily$ yields $(h_\beta(z_{\alpha,1}), \ldots, h_\beta(z_{\alpha,m})) = (\alpha_1, \ldots, \alpha_{d_{\Rreach}}, \beta_1, \ldots, \beta_{d_{\Hfamily}})$, which takes $2^{d_{\Hfamily}}$ distinct values as $\beta$ ranges over $\{0,1\}^{d_{\Hfamily}}$; hence $|\RelClass(z_\alpha)| = 2^{d_{\Hfamily}}$, proving $\PihatH(m) = 2^{d_{\Hfamily}}$ (and $\PihatH(m') = 2^{d_{\Hfamily}}$ for all $m' \ge d_{\Hfamily}$).

\paragraph{Target distribution family and zero approximation error.}
Fix $\gamma \in (0, 1/2]$.
For each $\omega \in \{0, 1\}^m$, define the target probability measure $\PT^\omega$ on $\XT \times \{0, 1\}$ by
\begin{equation*}
  \PT^\omega(X = x_i) = \frac{1}{m}, \;
  \PT^\omega(Y = \omega_i \mid X = x_i) = \frac{1 + \gamma}{2}, \;
  \PT^\omega(Y = 1 - \omega_i \mid X = x_i) = \frac{1 - \gamma}{2},
\end{equation*}
for all $i \in [m]$,
and collect these $2^m$ distributions into
\begin{equation*}
  \mathfrak{F}^{\mathrm{bin}}_{d_{\Rreach}, d_{\Hfamily}}(\gamma) \;:=\; \bigl\{\PT^\omega : \omega \in \{0, 1\}^m\bigr\}.
\end{equation*}
Since $\gamma > 0$, $(1 + \gamma)/2 > (1 - \gamma)/2$, so the unique Bayes-optimal classifier on $\XT$ under $\PT^\omega$ is $b_{\rm T}^\omega(x_i) = \omega_i$ ($i \in [m]$), achieving Bayes risk $\RiskStar(\PT^\omega) = (1 - \gamma)/2$.
Because $f_\omega \in \FMR$ satisfies $f_\omega(x_i) = \omega_i = b_{\rm T}^\omega(x_i)$ for all $i \in [m]$, we have $\Risk(f_\omega) = \RiskStar(\PT^\omega)$ and hence $\RiskClsOptimum(\PT^\omega) = \RiskStar(\PT^\omega)$, i.e.,
\begin{equation*}
  a_{\mathrm{MR}}(\PT^\omega) \;=\; \RiskClsOptimum(\PT^\omega) - \RiskStar(\PT^\omega) \;=\; 0 \qquad \text{for every } \PT^\omega \in \mathfrak{F}^{\mathrm{bin}}_{d_{\Rreach}, d_{\Hfamily}}(\gamma).
\end{equation*}
(If an ambient Euclidean domain is desired, embedding $\XT$ and $\XS$ into Euclidean spaces and extending $\fs \equiv 0$ outside $\XS$ preserves measurability without altering any distribution in $\mathfrak{F}^{\mathrm{bin}}_{d_{\Rreach}, d_{\Hfamily}}(\gamma)$, since $\PT^\omega$ is supported on $\XT$.)
This completes the verification of \Cref{con:binary-family}.

\subsection{Proof of \Cref{thm:T3P} and \Cref{cor:T3P-sample}}
\label{app:proof_T3P:lower}

We first establish the pairwise product-measure KL bound and hypercube Assouad reduction used in the proof of \Cref{thm:T3P}.
Write $\rho_{\mathrm{H}}(\omega, \omega') := \sum_{i=1}^{m} \one\{\omega_i \neq \omega'_i\}$ for Hamming distance on $\{0, 1\}^m$, $\mathrm{TV}(P, Q) := \sup_A |P(A) - Q(A)|$ for total variation distance over measurable events $A$, and $\mathrm{KL}(P \,\|\, Q) := \int \log(\mathrm dP/\mathrm dQ)\,\mathrm dP$ for Kullback--Leibler divergence.

\begin{lemma}[Neighboring-law KL bound and hypercube Assouad reduction]
\label{lem:binary-neighbor-kl}
Let $\mathfrak{F}^{\mathrm{bin}}_{d_{\Rreach}, d_{\Hfamily}}(\gamma) = \{\PT^\omega : \omega \in \{0, 1\}^m\}$ be the family of \Cref{app:proof_T3P:construction} with $m = d_{\Rreach} + d_{\Hfamily} \ge 2$ and $\gamma \in (0, 1/2]$, and let $\nT \ge 1$.
For any $\omega, \omega' \in \{0, 1\}^m$ satisfying $\rho_{\mathrm{H}}(\omega, \omega') = 1$, the $\nT$-fold product laws satisfy
\begin{equation*}
  \mathrm{KL}\!\left((\PT^\omega)^{\otimes \nT} \,\middle\|\, (\PT^{\omega'})^{\otimes \nT}\right)
  \;=\; \frac{\nT \gamma}{m}\log\!\left(\frac{1 + \gamma}{1 - \gamma}\right)
  \;\le\; \frac{4\nT \gamma^2}{m}.
\end{equation*}
Consequently, if $4\nT \gamma^2 / m \le \bar\alpha$ for some $\bar\alpha > 0$, then every measurable estimator $\widehat\omega : (\XT \times \{0, 1\})^{\nT} \to \{0, 1\}^m$ satisfies
\begin{equation*}
  \sup_{\omega \in \{0, 1\}^m} \E_{(\PT^\omega)^{\otimes \nT}}\bigl[\rho_{\mathrm{H}}(\widehat\omega, \omega)\bigr]
  \;\ge\; \frac{m}{2}\max\!\left\{\frac{1}{2}e^{-\bar\alpha},\, 1 - \sqrt{\frac{\bar\alpha}{2}}\right\}
  \;\ge\; \frac{m}{4}e^{-\bar\alpha}.
\end{equation*}
\end{lemma}

\begin{proof}
Suppose $\omega, \omega' \in \{0, 1\}^m$ differ only at coordinate $i \in [m]$.
Under $\PT^\omega$ and $\PT^{\omega'}$, the marginal law of $X$ is uniform on $\XT$, and for every $j \neq i$ the conditional laws $\PT^\omega(Y = \cdot \mid X = x_j)$ and $\PT^{\omega'}(Y = \cdot \mid X = x_j)$ coincide; at $X = x_i$ (which carries marginal mass $1/m$), the Bernoulli parameters $(1+\gamma)/2$ and $(1-\gamma)/2$ are swapped.
Using additivity of KL divergence across $\nT$ independent draws and $1 - t^2 \ge 1 - \gamma^2 \ge 3/4 > 1/2$ for all $t \in [0, \gamma] \subseteq [0, 1/2]$, direct summation over $\XT \times \{0, 1\}$ gives
\begin{align*}
  \mathrm{KL}\!\left((\PT^\omega)^{\otimes \nT} \,\middle\|\, (\PT^{\omega'})^{\otimes \nT}\right)
  &= \nT\,\mathrm{KL}(\PT^\omega \,\|\, \PT^{\omega'}) \\
  &= \frac{\nT}{m}\left\{\frac{1 + \gamma}{2}\log\!\left(\frac{1 + \gamma}{1 - \gamma}\right) + \frac{1 - \gamma}{2}\log\!\left(\frac{1 - \gamma}{1 + \gamma}\right)\right\} \\
  &= \frac{\nT \gamma}{m}\log\!\left(\frac{1 + \gamma}{1 - \gamma}\right)
    \;=\; \frac{2\nT \gamma}{m}\int_0^\gamma \frac{\mathrm dt}{1 - t^2} \\
  &\le \frac{2\nT \gamma}{m}\int_0^\gamma 2\,\mathrm dt
    \;=\; \frac{4\nT \gamma^2}{m},
\end{align*}
where the first equality uses KL tensorization across independent draws and the inequality uses $(1 - t^2)^{-1} \le 2$ on $[0, \gamma]$.

Now suppose $4\nT \gamma^2 / m \le \bar\alpha$ with $\bar\alpha > 0$.
Write $P_\omega := (\PT^\omega)^{\otimes \nT}$ with probability $\Prob_\omega$ and expectation $\E_\omega$, and for each $\omega \in \{0, 1\}^m$ and $i \in [m]$ let $\omega^{\oplus i} \in \{0, 1\}^m$ denote the neighbor obtained by flipping coordinate $i$ of $\omega$, so $\rho_{\mathrm{H}}(\omega, \omega^{\oplus i}) = 1$ and $\mathrm{KL}(P_\omega \,\|\, P_{\omega^{\oplus i}}) \le \bar\alpha$.
For any measurable estimator $\widehat\omega : (\XT \times \{0, 1\})^{\nT} \to \{0, 1\}^m$, averaging over the uniform prior on $\{0, 1\}^m$ gives
\begin{align*}
  \sup_{\omega \in \{0, 1\}^m} \E_\omega\bigl[\rho_{\mathrm{H}}(\widehat\omega, \omega)\bigr]
  &\ge \frac{1}{2^m}\sum_{\omega \in \{0, 1\}^m} \E_\omega\!\left[\sum_{i=1}^{m} \one\{\widehat\omega_i \neq \omega_i\}\right]
    &&\text{(uniform prior)} \\
  &= \sum_{i=1}^{m} \frac{1}{2^m}\sum_{\omega \in \{0, 1\}^m} \Prob_\omega(\widehat\omega_i \neq \omega_i)
    &&\text{(linearity)} \\
  &= \frac{1}{2}\sum_{i=1}^{m} \frac{1}{2^{m-1}}\sum_{\omega :\, \omega_i = 0} \Bigl(\Prob_\omega(\widehat\omega_i = 1) + \Prob_{\omega^{\oplus i}}(\widehat\omega_i = 0)\Bigr)
    &&\text{(pairing $\omega, \omega^{\oplus i}$)} \\
  &\ge \frac{1}{2}\sum_{i=1}^{m} \frac{1}{2^{m-1}}\sum_{\omega :\, \omega_i = 0} \bigl(1 - \mathrm{TV}(P_\omega, P_{\omega^{\oplus i}})\bigr)
    &&\text{(Le Cam testing)} \\
  &\ge \frac{1}{2}\sum_{i=1}^{m} \frac{1}{2^{m-1}}\sum_{\omega :\, \omega_i = 0} \max\!\left\{\frac{1}{2}e^{-\bar\alpha},\, 1 - \sqrt{\frac{\bar\alpha}{2}}\right\}
    &&\text{(KL $\le \bar\alpha$)} \\
  &= \frac{m}{2}\max\!\left\{\frac{1}{2}e^{-\bar\alpha},\, 1 - \sqrt{\frac{\bar\alpha}{2}}\right\}
    \;\ge\; \frac{m}{4}e^{-\bar\alpha},
\end{align*}
where the fourth relation applies the two-point testing identity $\Prob_\omega(\widehat\omega_i = 1) + \Prob_{\omega^{\oplus i}}(\widehat\omega_i = 0) = 1 - \bigl(P_{\omega^{\oplus i}}(\widehat\omega_i = 1) - P_\omega(\widehat\omega_i = 1)\bigr) \ge 1 - \mathrm{TV}(P_\omega, P_{\omega^{\oplus i}})$, and the fifth applies the Bretagnolle--Huber inequality $1 - \mathrm{TV}(P, Q) \ge \frac{1}{2}e^{-\mathrm{KL}(P \,\|\, Q)}$ together with Pinsker's inequality $\mathrm{TV}(P, Q) \le \sqrt{\mathrm{KL}(P \,\|\, Q)/2}$~\citep[Lemmas~2.5 and~2.6, Theorem~2.12]{tsybakov2009introduction} under $\mathrm{KL}(P_\omega \,\|\, P_{\omega^{\oplus i}}) \le \bar\alpha$ and $|\{\omega \in \{0, 1\}^m : \omega_i = 0\}| = 2^{m-1}$.
\end{proof}

\begin{theorem}[Matching minimax lower bound; restatement of \Cref{thm:T3P}]
\label{thm:T3P-app}
With $c_0 = 1$, $c_1 = 1/4$, and $c_2 = e^{-1/4}/16$, for all integers $d_{\Rreach}, d_{\Hfamily} \ge 1$ and $\nT \ge 1$ satisfying $d_{\Rreach} + d_{\Hfamily} \le c_0 \nT$, on the binary reprogramming family $\mathfrak{F}^{\mathrm{bin}}_{d_{\Rreach}, d_{\Hfamily}}(\gamma)$ (\Cref{con:binary-family,app:proof_T3P:construction}) with margin $\gamma = c_1\sqrt{(d_{\Rreach} + d_{\Hfamily})/\nT}$, every learning rule $\widehat f_n$ mapping $\nT$ i.i.d.\ target samples to a measurable predictor $\XT \to [K_{\rm T}]$ satisfies
\begin{equation*}
  \sup_{\PT \in \mathfrak{F}^{\mathrm{bin}}_{d_{\Rreach}, d_{\Hfamily}}(\gamma)} \E_{\PT^{\nT}}\bigl[\,\EstRisk(\widehat f_n)\,\bigr]
  \;\ge\; c_2 \sqrt{\frac{d_{\Rreach} + d_{\Hfamily}}{\nT}}.
\end{equation*}
\end{theorem}

\begin{proof}
Let $m := d_{\Rreach} + d_{\Hfamily} \ge 2$ and work on the family $\mathfrak{F}^{\mathrm{bin}}_{d_{\Rreach}, d_{\Hfamily}}(\gamma) = \{\PT^\omega : \omega \in \{0, 1\}^m\}$ of \Cref{app:proof_T3P:construction}, writing $\Risk^{\omega}(f) := \PT^\omega(f(X) \neq Y)$ for the target $0$--$1$ risk under $\PT^\omega$.
By \Cref{app:proof_T3P:construction}, $a_{\mathrm{MR}}(\PT^\omega) = 0$ and $\RiskClsOptimum(\PT^\omega) = \RiskStar(\PT^\omega) = (1 - \gamma)/2$ for every $\omega \in \{0, 1\}^m$.
Conditioned on the realized $\nT$-sample, any learning rule fixes a measurable predictor $\widehat f_n : \XT \to \{0, 1\}$ (whether or not $\widehat f_n$ takes values in $\FMR$) and induces the bit vector $\widehat\omega = (\widehat\omega_1, \ldots, \widehat\omega_m) \in \{0, 1\}^m$ defined by $\widehat\omega_i := \widehat f_n(x_i)$ for $i \in [m]$.
At each support point $x_i$, $\PT^\omega\bigl(\widehat f_n(x_i) \neq Y \mid X = x_i\bigr)$ equals $(1 - \gamma)/2$ when $\widehat\omega_i = \omega_i$ and $(1 + \gamma)/2 = (1 - \gamma)/2 + \gamma$ when $\widehat\omega_i \neq \omega_i$.
Disintegrating $\Risk^{\omega}(\widehat f_n)$ over the uniform marginal $\PT^\omega(X = x_i) = 1/m$ gives
\begin{align}
  \EstRisk(\widehat f_n)
  &= \Risk^{\omega}(\widehat f_n) - \RiskClsOptimum(\PT^\omega)
    &&\text{(\Cref{def:errors})} \nonumber \\
  &= \Risk^{\omega}(\widehat f_n) - \RiskStar(\PT^\omega)
    &&\text{($a_{\mathrm{MR}}(\PT^\omega) = 0$)} \nonumber \\
  &= \frac{1}{m}\sum_{i=1}^{m} \PT^\omega\bigl(\widehat f_n(x_i) \neq Y \;\big|\; X = x_i\bigr) - \frac{1 - \gamma}{2}
    &&\text{(uniform marginal)} \nonumber \\
  &= \frac{1}{m}\sum_{i=1}^{m} \left(\frac{1 - \gamma}{2} + \gamma\,\one\{\widehat\omega_i \neq \omega_i\}\right) - \frac{1 - \gamma}{2} \nonumber \\
  &= \frac{1}{m}\sum_{i=1}^{m} \gamma\,\one\{\widehat\omega_i \neq \omega_i\}
    \;=\; \frac{\gamma}{m}\,\rho_{\mathrm{H}}(\widehat\omega, \omega).
  \label{eq:lower-risk-hamming-expanded}
\end{align}

Under $m \le c_0 \nT = \nT$ and $\gamma = c_1\sqrt{m/\nT} = \frac{1}{4}\sqrt{m/\nT}$, we have $\gamma \in (0, 1/4] \subset (0, 1/2]$ and
\begin{equation*}
  \frac{4\nT \gamma^2}{m} \;=\; \frac{4\nT}{m}\cdot \frac{m}{16\nT} \;=\; \frac{1}{4}.
\end{equation*}
Applying \Cref{lem:binary-neighbor-kl} with $\bar\alpha := 1/4$ to the measurable estimator $\widehat\omega$ induced by $\widehat f_n$, taking expectation over $(\PT^\omega)^{\otimes \nT}$ in \eqref{eq:lower-risk-hamming-expanded}, and taking the supremum over $\omega \in \{0, 1\}^m$ yields
\begin{align*}
  \sup_{\PT \in \mathfrak{F}^{\mathrm{bin}}_{d_{\Rreach}, d_{\Hfamily}}(\gamma)} \E_{\PT^{\nT}}\bigl[\,\EstRisk(\widehat f_n)\,\bigr]
  &= \frac{\gamma}{m}\sup_{\omega \in \{0, 1\}^m} \E_{(\PT^\omega)^{\otimes \nT}}\bigl[\rho_{\mathrm{H}}(\widehat\omega, \omega)\bigr]
    &&\text{(\eqref{eq:lower-risk-hamming-expanded})} \\
  &\ge \frac{\gamma}{m}\cdot \frac{m}{4}e^{-1/4}
    &&\text{(\Cref{lem:binary-neighbor-kl})} \\
  &= \frac{e^{-1/4}}{16}\sqrt{\frac{d_{\Rreach} + d_{\Hfamily}}{\nT}},
    &&\text{($\gamma = \tfrac{1}{4}\sqrt{m/\nT}$)}
\end{align*}
establishing the claim with $c_2 = e^{-1/4}/16$.
\end{proof}

\begin{proof}[Proof of \Cref{cor:T3P-sample}]
Fix integers $d_{\Rreach}, d_{\Hfamily} \ge 1$, set $m := d_{\Rreach} + d_{\Hfamily} \ge 2$, fix $\varepsilon \in (0, c_2/2)$, and let $\widehat f_n$ be any learning rule on $\nT \ge 1$ i.i.d.\ target samples achieving $\sup_{\PT \in \mathfrak{F}^{\mathrm{bin}}_{d_{\Rreach}, d_{\Hfamily}}(\gamma)} \E_{\PT^{\nT}}\bigl[\EstRisk(\widehat f_n)\bigr] \le \varepsilon$ at margin $\gamma = \min\bigl\{1/4,\, c_1\sqrt{m/\nT}\bigr\}$ with $c_1 = 1/4$.

\textbf{Case 1 ($\nT \ge d_{\Rreach} + d_{\Hfamily}$).}
Here $\nT \ge m$, so $\sqrt{m/\nT} \le 1$ and the margin evaluates to
\begin{equation*}
  \gamma \;=\; \min\!\left\{\frac{1}{4},\, \frac{1}{4}\sqrt{\frac{m}{\nT}}\right\} \;=\; \frac{1}{4}\sqrt{\frac{m}{\nT}} \;=\; c_1\sqrt{\frac{m}{\nT}}.
\end{equation*}
Since $m \le \nT = c_0\nT$, \Cref{thm:T3P-app} applies directly and gives
\begin{equation*}
  \varepsilon
  \;\ge\; \sup_{\PT \in \mathfrak{F}^{\mathrm{bin}}_{d_{\Rreach}, d_{\Hfamily}}(\gamma)} \E_{\PT^{\nT}}\bigl[\,\EstRisk(\widehat f_n)\,\bigr]
  \;\ge\; c_2\sqrt{\frac{m}{\nT}},
\end{equation*}
which rearranges to $\nT \ge c_2^2 m / \varepsilon^2 = c_2^2(d_{\Rreach} + d_{\Hfamily})/\varepsilon^2$.

\textbf{Case 2 ($1 \le \nT < d_{\Rreach} + d_{\Hfamily}$).}
Here $1 \le \nT < m$, so $\sqrt{m/\nT} > 1$ and the margin evaluates to
\begin{equation*}
  \gamma \;=\; \min\!\left\{\frac{1}{4},\, \frac{1}{4}\sqrt{\frac{m}{\nT}}\right\} \;=\; \frac{1}{4} \;\in\; (0, 1/2],
\end{equation*}
for which $4\nT\gamma^2 / m = \nT / (4m) < 1/4 =: \bar\alpha$.
Combining \eqref{eq:lower-risk-hamming-expanded} with \Cref{lem:binary-neighbor-kl} at $\gamma = 1/4$ and $\bar\alpha = 1/4$ gives
\begin{align*}
  \sup_{\PT \in \mathfrak{F}^{\mathrm{bin}}_{d_{\Rreach}, d_{\Hfamily}}(1/4)} \E_{\PT^{\nT}}\bigl[\,\EstRisk(\widehat f_n)\,\bigr]
  &= \frac{1/4}{m}\sup_{\omega \in \{0, 1\}^m} \E_{(\PT^\omega)^{\otimes \nT}}\bigl[\rho_{\mathrm{H}}(\widehat\omega, \omega)\bigr]
    &&\text{(\eqref{eq:lower-risk-hamming-expanded})} \\
  &\ge \frac{1}{4m}\cdot\frac{m}{4}e^{-1/4}
    \;=\; \frac{e^{-1/4}}{16}
    \;=\; c_2,
    &&\text{(\Cref{lem:binary-neighbor-kl})}
\end{align*}
contradicting $\sup_{\PT \in \mathfrak{F}^{\mathrm{bin}}_{d_{\Rreach}, d_{\Hfamily}}(\gamma)} \E_{\PT^{\nT}}\bigl[\EstRisk(\widehat f_n)\bigr] \le \varepsilon < c_2 / 2 < c_2$.
Thus Case~2 cannot occur, and $\nT \ge c_2^2(d_{\Rreach} + d_{\Hfamily})/\varepsilon^2 = \Omega\bigl((d_{\Rreach} + d_{\Hfamily})/\varepsilon^2\bigr)$ is necessary.
\end{proof}

\begin{remark}[Specialization of \Cref{thm:T1} on $\mathfrak{F}^{\mathrm{bin}}_{d_{\Rreach}, d_{\Hfamily}}(\gamma)$ and scope of \Cref{thm:T3P}]
\label{rem:T3P-upper-specialization}
On the binary family of \Cref{app:proof_T3P:construction}, we proved $a_{\mathrm{MR}} = 0$, $\log\PihatR(2\nT) \le d_{\Rreach}\log 2$, and $\log\PihatH(2\nT) \le d_{\Hfamily}\log 2$.
Substituting these bounds into \Cref{thm:T1-app} (\Cref{rem:T1-bayes-excess-consequence}) shows that for every $\PT \in \mathfrak{F}^{\mathrm{bin}}_{d_{\Rreach}, d_{\Hfamily}}(\gamma)$, every $\nT \ge 1$, every $\delta \in (0,1)$, and every $\eta_n \ge 0$, with probability at least $1 - \delta$ over $\PT^{\nT}$, every measurable $\eta_n$-approximate ERM $\widehat f_n \in \FMR$ satisfies
\begin{equation*}
  \Risk(\widehat f_n) - \RiskStar \;=\; \EstRisk(\widehat f_n) \;\le\; \eta_n + 4\sqrt{6}\,\sqrt{\frac{(d_{\Rreach} + d_{\Hfamily})\log 2 + \log(1/\delta)}{\nT}},
\end{equation*}
matching the $\sqrt{(d_{\Rreach} + d_{\Hfamily})/\nT}$ scaling of \Cref{thm:T3P-app} up to absolute constants and the high-probability confidence term.
As noted in \Cref{sec:T3P}, \Cref{thm:T3P-app} establishes existential rate tightness on $\mathfrak{F}^{\mathrm{bin}}_{d_{\Rreach}, d_{\Hfamily}}(\gamma)$, showing that neither $d_{\Rreach}$ nor $d_{\Hfamily}$ can be omitted in general, rather than a lower bound holding for every fixed source classifier $\fs$.
\end{remark}

%% file: appendix/proof_identities.tex
\label{app:proof-identities}

\textbf{Proof roadmap.}
This section proves the three exact population approximation identities of the paper: the hard-label $0$--$1$ decomposition (\Cref{prop:T2P} in \Cref{app:proof-prop-hard-identity}), the score-output $0$--$1$ decomposition (\Cref{prop:score-01-identity} in \Cref{app:proof-prop-score-01}), and the score-output surrogate decomposition (\Cref{prop:score-surrogate-identity} in \Cref{app:proof-prop-surrogate}).
Each proof rewrites the wrapper infimum as a nested optimization over the input transformation $g \in \Gfamily$ and the output mapping ($h \in \Hfamily$ or $W \in \Wclass$), characterizes the unrestricted channel optimum conditioned on the induced representation ($Z_g = \fs(g(X))$ or $U_g = s_{\rm S}(g(X))$), and telescopes the fixed-$g$ excess risk around that intermediate reference while verifying pointwise nonnegativity of both summands.
None of the three identities requires bounded scores, Lipschitz losses, classification calibration, or finite-sample concentration.

\subsection{Proof of \Cref{prop:T2P}}
\label{app:proof-prop-hard-identity}

\begin{proof}
Fix $g \in \Gfamily$ and write $Z_g := \fs(g(X)) \in [K_{\rm S}]$ for the induced source-label random variable under $(X, Y) \sim \PT$.
Because $\fs: \XS \to [K_{\rm S}]$ and $g: \XT \to \XS$ are measurable, $Z_g$ is measurable.
For each $(k, y) \in [K_{\rm S}] \times [K_{\rm T}]$, set $p_g(k, y) := \Prob_{(X,Y)\sim\PT}(Z_g = k,\, Y = y)$.
For any relabeling $h \in [K_{\rm T}]^{[K_{\rm S}]}$, writing $\Risk(h(Z_g)) := \Risk(h \circ \fs \circ g)$ and partitioning $\{h(Z_g) = Y\}$ across the disjoint level sets $\{Z_g = k\}_{k=1}^{K_{\rm S}}$ gives
\begin{align*}
  \Risk\bigl(h(Z_g)\bigr)
  &= \Prob_{(X,Y)\sim\PT}\bigl(h(Z_g) \neq Y\bigr) \\
  &= 1 - \Prob_{(X,Y)\sim\PT}\bigl(h(Z_g) = Y\bigr) \\
  &= 1 - \sum_{k=1}^{K_{\rm S}} \Prob_{(X,Y)\sim\PT}\bigl(Z_g = k,\, Y = h(k)\bigr)
    &&\text{(disjoint level sets)} \\
  &= 1 - \sum_{k=1}^{K_{\rm S}} p_g\bigl(k, h(k)\bigr).
\end{align*}
Since $\Gfamily$ and $\Hfamily \subseteq [K_{\rm T}]^{[K_{\rm S}]}$ are nonempty and $\FMR = \{h \circ \fs \circ g : g \in \Gfamily,\, h \in \Hfamily\}$, setting $R^{\ast}_{\Hfamily}(g) := \inf_{h \in \Hfamily} \Risk\bigl(h(Z_g)\bigr)$ factors the wrapper approximation error as
\begin{align*}
  a_{\mathrm{MR}}
  &= \RiskClsOptimum - \RiskStar
    &&\text{(by \Cref{def:errors})} \\
  &= \inf_{f \in \FMR} \Risk(f) - \RiskStar \\
  &= \inf_{g \in \Gfamily} \inf_{h \in \Hfamily} \Risk\bigl(h(Z_g)\bigr) - \RiskStar
    &&\text{(nested infimum)} \\
  &= \inf_{g \in \Gfamily}\bigl\{R^{\ast}_{\Hfamily}(g) - \RiskStar\bigr\}.
    &&\text{($\Gfamily \neq \varnothing$)}
\end{align*}

Now consider the unrestricted relabeling class $\Hfamily_{\mathrm{all}} := [K_{\rm T}]^{[K_{\rm S}]}$, whose coordinates $h(1), \ldots, h(K_{\rm S}) \in [K_{\rm T}]$ vary independently, and denote the active support of $Z_g$ by $\mathcal{Z}_g := \{k \in [K_{\rm S}] : \PT(Z_g = k) > 0\}$.
For every zero-probability atom $k \in [K_{\rm S}] \setminus \mathcal{Z}_g$, monotonicity of probability gives $0 \le p_g(k, y) \le \PT(Z_g = k) = 0$ for all $y \in [K_{\rm T}]$, so $\max_{y \in [K_{\rm T}]} p_g(k, y) = 0$; meanwhile, $\PT(Z_g \notin \mathcal{Z}_g) = \sum_{k \notin \mathcal{Z}_g} \PT(Z_g = k) = 0$, so any version of the conditional probability $\PT(Y = y \mid Z_g)$ agrees with $p_g(k, y)/\PT(Z_g = k)$ on $\{Z_g = k\}$ for every $k \in \mathcal{Z}_g$ and integrates to $0$ over $\{Z_g \notin \mathcal{Z}_g\}$.
Evaluating $R^{\ast}_{\mathrm{all}}(g) := \inf_{h \in \Hfamily_{\mathrm{all}}} \Risk\bigl(h(Z_g)\bigr)$ therefore yields
\begin{align*}
  R^{\ast}_{\mathrm{all}}(g)
  &= \inf_{h \in \Hfamily_{\mathrm{all}}}\!\left\{1 - \sum_{k=1}^{K_{\rm S}} p_g\bigl(k, h(k)\bigr)\right\} \\
  &= 1 - \sup_{h \in \Hfamily_{\mathrm{all}}} \sum_{k=1}^{K_{\rm S}} p_g\bigl(k, h(k)\bigr) \\
  &= 1 - \sum_{k=1}^{K_{\rm S}} \max_{y \in [K_{\rm T}]} p_g(k, y) \\
  &= 1 - \sum_{k \in \mathcal{Z}_g} \max_{y \in [K_{\rm T}]} p_g(k, y) \\
  &= 1 - \sum_{k \in \mathcal{Z}_g} \PT(Z_g = k) \max_{y \in [K_{\rm T}]} \PT(Y = y \mid Z_g = k) \\
  &= \E_{\PT}\!\left[1 - \max_{y \in [K_{\rm T}]} \PT(Y = y \mid Z_g)\right],
\end{align*}
where the third equality decouples the supremum across $k \in [K_{\rm S}]$, the fourth drops null atoms $k \notin \mathcal{Z}_g$, the fifth factors $p_g(k, y) = \PT(Z_g = k)\,\PT(Y = y \mid Z_g = k)$ on $\mathcal{Z}_g$, and the sixth rewrites the sum over $\mathcal{Z}_g$ as an expectation under $\PT(Z_g \in \mathcal{Z}_g) = 1$.

Because $R^{\ast}_{\Hfamily}(g), R^{\ast}_{\mathrm{all}}(g), \RiskStar \in [0, 1]$ are finite, adding and subtracting $R^{\ast}_{\mathrm{all}}(g)$ inside the outer infimum over $g \in \Gfamily$ gives
\begin{equation*}
  a_{\mathrm{MR}}
  \;=\; \inf_{g \in \Gfamily}\Bigl\{\bigl(R^{\ast}_{\mathrm{all}}(g) - \RiskStar\bigr) + \bigl(R^{\ast}_{\Hfamily}(g) - R^{\ast}_{\mathrm{all}}(g)\bigr)\Bigr\}.
\end{equation*}
Finally, for every $g \in \Gfamily$, each composition $h \circ \fs \circ g$ ($h \in \Hfamily_{\mathrm{all}}$) is a measurable classifier $\XT \to [K_{\rm T}]$ and $\Hfamily \subseteq \Hfamily_{\mathrm{all}}$, so monotonicity of the infimum gives
\begin{align*}
  R^{\ast}_{\mathrm{all}}(g) - \RiskStar
  &= \inf_{h \in \Hfamily_{\mathrm{all}}} \Risk(h \circ \fs \circ g) - \inf_{f : \XT \to [K_{\rm T}]\ \text{measurable}} \Risk(f) \;\ge\; 0, \\
  R^{\ast}_{\Hfamily}(g) - R^{\ast}_{\mathrm{all}}(g)
  &= \inf_{h \in \Hfamily} \Risk\bigl(h(Z_g)\bigr) - \inf_{h \in \Hfamily_{\mathrm{all}}} \Risk\bigl(h(Z_g)\bigr) \;\ge\; 0,
\end{align*}
establishing \eqref{eq:T2P}.
\end{proof}

\subsection{Proof of \Cref{prop:score-01-identity}}
\label{app:proof-prop-score-01}

\begin{proof}
Recall from \Cref{def:score-wrapper} that $\mathcal{F}_{\mathrm{score}} = \{f_{g,W} = a_W \circ s_{\rm S} \circ g : g \in \Gfamily,\, W \in \Wclass\}$, where $a_W(u) = \lexargmax_{y \in [K_{\rm T}]} (Wu)_y$ for $u \in \mathbb{R}^{K_{\rm S}}$.
For each $W \in \Wclass$ and $k \in [K_{\rm T}]$, the level set
\begin{align*}
  \bigl\{u \in \mathbb{R}^{K_{\rm S}} : a_W(u) = k\bigr\}
  \;=\;& \bigcap_{r < k}\bigl\{u \in \mathbb{R}^{K_{\rm S}} : (Wu)_k > (Wu)_r\bigr\} \\
       &\cap \bigcap_{r > k}\bigl\{u \in \mathbb{R}^{K_{\rm S}} : (Wu)_k \ge (Wu)_r\bigr\}
\end{align*}
is a finite intersection of open and closed half-spaces, so $a_W$ is Borel-measurable; since $s_{\rm S}$ and $g$ are measurable, $f_{g,W} = a_W(U_g)$ with $U_g := s_{\rm S}(g(X))$ is measurable.

Fix $g \in \Gfamily$.
Because $\XT \times [K_{\rm T}]$ is standard Borel and $U_g$ takes values in $\mathbb{R}^{K_{\rm S}}$, regular conditional probabilities $\eta^g_y(u) := \Prob_{\PT}(Y = y \mid U_g = u)$ exist as Borel-measurable functions of $u \in \mathbb{R}^{K_{\rm S}}$ for each $y \in [K_{\rm T}]$.
Define the lexicographically tie-broken channel selector $b_g^\star(u) := \lexargmax_{y \in [K_{\rm T}]} \eta^g_y(u)$ for $u \in \mathbb{R}^{K_{\rm S}}$.
For each $k \in [K_{\rm T}]$, $\{u \in \mathbb{R}^{K_{\rm S}} : b_g^\star(u) = k\} = \bigcap_{r < k}\{u : \eta^g_k(u) > \eta^g_r(u)\} \cap \bigcap_{r > k}\{u : \eta^g_k(u) \ge \eta^g_r(u)\}$ is a finite intersection of Borel sets, so $b_g^\star : \mathbb{R}^{K_{\rm S}} \to [K_{\rm T}]$ is Borel-measurable and satisfies $\eta^g_{b_g^\star(u)}(u) = \max_{y \in [K_{\rm T}]} \eta^g_y(u)$ for every $u \in \mathbb{R}^{K_{\rm S}}$.
For any Borel-measurable map $a : \mathbb{R}^{K_{\rm S}} \to [K_{\rm T}]$, conditioning on $U_g$ gives
\begin{align*}
  \Risk\bigl(a(U_g)\bigr)
  &= \Prob_{\PT}\bigl(Y \neq a(U_g)\bigr) \\
  &= \E_{\PT}\bigl[\Prob_{\PT}\bigl(Y \neq a(U_g) \mid U_g\bigr)\bigr]
    &&\text{(tower property)} \\
  &= \E_{\PT}\bigl[1 - \eta^g_{a(U_g)}(U_g)\bigr]
    &&\text{($\sigma(U_g)$-measurability)} \\
  &\ge \E_{\PT}\!\left[1 - \max_{y \in [K_{\rm T}]} \eta^g_y(U_g)\right]
    &&\text{($\eta^g_{a(u)}(u) \le \max_y \eta^g_y(u)$)} \\
  &= \E_{\PT}\bigl[1 - \eta^g_{b_g^\star(U_g)}(U_g)\bigr]
    &&\text{(definition of $b_g^\star$)} \\
  &= \Risk\bigl(b_g^\star(U_g)\bigr).
\end{align*}
Since $b_g^\star$ itself is Borel-measurable from $\mathbb{R}^{K_{\rm S}}$ to $[K_{\rm T}]$, the infimum over measurable $a : \mathbb{R}^{K_{\rm S}} \to [K_{\rm T}]$ is attained at $b_g^\star$, giving $R^\star_{\rm T}(U_g) = \Risk\bigl(b_g^\star(U_g)\bigr) = \E_{\PT}\bigl[1 - \max_{y \in [K_{\rm T}]} \eta^g_y(U_g)\bigr]$.

Because all $0$--$1$ risks lie in $[0, 1]$ and $\Gfamily, \Wclass$ are nonempty, factoring the joint infimum over $(g, W) \in \Gfamily \times \Wclass$ and adding and subtracting $R^\star_{\rm T}(U_g)$ (which is independent of $W$) yields
\begin{align*}
  a_{\mathrm{score}, 0/1}
  &= \inf_{(g, W) \in \Gfamily \times \Wclass} \Risk(f_{g, W}) - \RiskStar \\
  &= \inf_{g \in \Gfamily}\Bigl\{\inf_{W \in \Wclass} \Risk\bigl(a_W(U_g)\bigr) - \RiskStar\Bigr\}
    &&\text{($\Gfamily, \Wclass \neq \varnothing$)} \\
  &= \inf_{g \in \Gfamily}\Bigl\{\bigl(R^\star_{\rm T}(U_g) - \RiskStar\bigr) + \Bigl(\inf_{W \in \Wclass} \Risk\bigl(a_W(U_g)\bigr) - R^\star_{\rm T}(U_g)\Bigr)\Bigr\} \\
  &= \inf_{g \in \Gfamily}\Bigl\{\bigl(R^\star_{\rm T}(U_g) - \RiskStar\bigr) + \inf_{W \in \Wclass}\bigl(\Risk\bigl(a_W(U_g)\bigr) - R^\star_{\rm T}(U_g)\bigr)\Bigr\}.
    &&\text{($\Wclass \neq \varnothing$)}
\end{align*}
Moreover, for every $g \in \Gfamily$, $b_g^\star \circ s_{\rm S} \circ g : \XT \to [K_{\rm T}]$ is measurable and $a_W : \mathbb{R}^{K_{\rm S}} \to [K_{\rm T}]$ is Borel-measurable for each $W \in \Wclass$, whence
\begin{align*}
  R^\star_{\rm T}(U_g) - \RiskStar
  &= \Risk(b_g^\star \circ s_{\rm S} \circ g) - \inf_{f : \XT \to [K_{\rm T}]\ \text{measurable}} \Risk(f) \;\ge\; 0, \\
  \inf_{W \in \Wclass}\bigl(\Risk\bigl(a_W(U_g)\bigr) - R^\star_{\rm T}(U_g)\bigr)
  &= \inf_{W \in \Wclass} \E_{\PT}\bigl[\eta^g_{b_g^\star(U_g)}(U_g) - \eta^g_{a_W(U_g)}(U_g)\bigr] \;\ge\; 0,
\end{align*}
where the second inequality uses $\eta^g_{b_g^\star(u)}(u) - \eta^g_{a_W(u)}(u) \ge 0$ pointwise on $\mathbb{R}^{K_{\rm S}}$.
\end{proof}

\subsection{Proof of \Cref{prop:score-surrogate-identity}}
\label{app:proof-prop-surrogate}

\begin{proof}
For each $g \in \Gfamily$, write $U_g := s_{\rm S}(g(X))$ and set $A_\ell(g) := \inf_{W \in \Wclass} \Loss(g, W)$.
By hypothesis, $\Gfamily$ and $\Wclass$ are nonempty, $\ell \ge 0$, $\LossStar < \infty$, and $A_\ell(g) < \infty$ for every $g \in \Gfamily$.
Fix $g \in \Gfamily$; comparing infima across nested classes of measurable predictors gives
\begin{align*}
  0
  &\le \inf_{q_0 : \XT \to \mathbb{R}^{K_{\rm T}}\ \text{measurable}} \E_{\PT}\bigl[\ell\bigl(Y, q_0(X)\bigr)\bigr] \;=\; \LossStar \\
  &\le \inf_{q : \mathbb{R}^{K_{\rm S}} \to \mathbb{R}^{K_{\rm T}}\ \text{measurable}} \E_{\PT}\bigl[\ell\bigl(Y, q(s_{\rm S}(g(X)))\bigr)\bigr] \;=\; L^\star_\ell(U_g) \\
  &\le \inf_{W \in \Wclass} \E_{\PT}\bigl[\ell\bigl(Y, W s_{\rm S}(g(X))\bigr)\bigr] \;=\; A_\ell(g) \;<\; \infty,
\end{align*}
where the first inequality uses $\ell \ge 0$, the second holds because $q \circ s_{\rm S} \circ g : \XT \to \mathbb{R}^{K_{\rm T}}$ is measurable whenever $q : \mathbb{R}^{K_{\rm S}} \to \mathbb{R}^{K_{\rm T}}$ is Borel-measurable, and the third holds because every linear map $u \mapsto W u$ ($W \in \Wclass$) is continuous and hence Borel-measurable.

Because $\LossStar, L^\star_\ell(U_g), A_\ell(g) \in [0, \infty)$ are all finite, adding and subtracting $L^\star_\ell(U_g)$ introduces no $\infty - \infty$ indeterminacy:
\begin{align*}
  a_{\mathrm{score}, \ell}
  &= \inf_{(g, W) \in \Gfamily \times \Wclass} \Loss(g, W) - \LossStar \\
  &= \inf_{g \in \Gfamily}\bigl\{A_\ell(g) - \LossStar\bigr\}
    &&\text{($\Gfamily, \Wclass \neq \varnothing$)} \\
  &= \inf_{g \in \Gfamily}\Bigl\{\bigl(L^\star_\ell(U_g) - \LossStar\bigr) + \bigl(A_\ell(g) - L^\star_\ell(U_g)\bigr)\Bigr\}
    &&\text{($L^\star_\ell(U_g) < \infty$)} \\
  &= \inf_{g \in \Gfamily}\Bigl\{\bigl(L^\star_\ell(U_g) - \LossStar\bigr) + \inf_{W \in \Wclass}\bigl(\Loss(g, W) - L^\star_\ell(U_g)\bigr)\Bigr\}.
    &&\text{($\Wclass \neq \varnothing$)}
\end{align*}
Finally, the ordering $0 \le \LossStar \le L^\star_\ell(U_g) \le A_\ell(g) < \infty$ directly certifies $L^\star_\ell(U_g) - \LossStar \ge 0$ and $\inf_{W \in \Wclass}\bigl(\Loss(g, W) - L^\star_\ell(U_g)\bigr) = A_\ell(g) - L^\star_\ell(U_g) \ge 0$ for every $g \in \Gfamily$.
\end{proof}

\begin{remark}[Pointwise conditional surrogate Bayes form]
\label{rem:surrogate-bayes-form}
If, in addition to the assumptions of \Cref{prop:score-surrogate-identity}, regular conditional posteriors $\eta^g_y(u) = \Prob_{\PT}(Y = y \mid U_g = u)$ exist and $\ell$ satisfies measurable-selection or normal-integrand conditions permitting the interchange of expectation and pointwise infimum~\citep{zhang2004statistical, tewari2007consistency}, then $L^\star_\ell(U_g)$ admits the pointwise representation
\begin{equation*}
  L^\star_\ell(U_g) \;=\; \E_{\PT}\!\left[\inf_{v \in \mathbb{R}^{K_{\rm T}}} \sum_{y=1}^{K_{\rm T}} \eta^g_y(U_g)\,\ell(y, v)\right].
\end{equation*}
This formula interprets $L^\star_\ell(U_g)$ geometrically as a conditional surrogate Bayes risk, while the proof of \Cref{prop:score-surrogate-identity} holds without invoking measurable selection.
\end{remark}

%% file: appendix/proof_T2.tex
\label{app:proof_T2}

\textbf{Proof roadmap.}
This section proves the population source-conditional upper certificate for the hard-label approximation error (\Cref{thm:T2-app}, restating \Cref{thm:T2C}), records its finite-sample Bayes excess-risk consequence when combined with \Cref{thm:T1-app} (\Cref{cor:T2-combined-pac}), and pinpoints the two inequalities responsible for the one-directional boundary discussed in \Cref{rem:one-way-certificate} (\Cref{rem:T2-converse-obstructions}).
For each candidate wrapper $(g, h) \in \Gfamily \times \Hfamily$, the proof of \Cref{thm:T2-app} upper-bounds target Bayes excess risk by disagreement with the target Bayes selector $b_{\rm T}$, interpolates the relabeled source Bayes selector $h \circ b_{\rm S} \circ g$ via the indicator triangle inequality, and transfers the resulting source-classifier disagreement indicator $\phi_h$ from $P_{{\rm S},X}$ to the pushforward target marginal $g_\# P_{{\rm T},X}$.

Recall from \Cref{sec:T2:certificate} that for measurable Bayes-optimal classifiers $b_{\rm T} : \XT \to [K_{\rm T}]$ and $b_{\rm S} : \XS \to [K_{\rm S}]$ and any measurable pair $(g, h) \in \Gfamily \times \Hfamily$,
\begin{equation*}
\begin{aligned}
  \lamhard(g, h) &:= \Prob_{X \sim P_{{\rm T},X}}\!\bigl(b_{\rm T}(X) \neq h(b_{\rm S}(g(X)))\bigr), \\
  \epshard(h)    &:= \Prob_{U \sim P_{{\rm S},X}}\!\bigl(h(\fs(U)) \neq h(b_{\rm S}(U))\bigr), \\
  \phi_h(u)      &:= \one\bigl\{h(\fs(u)) \neq h(b_{\rm S}(u))\bigr\}, \\
  \Deltahard(g, h) &:= \Bigl|\E_{X \sim P_{{\rm T},X}}\bigl[\phi_h(g(X))\bigr] - \E_{U \sim P_{{\rm S},X}}\bigl[\phi_h(U)\bigr]\Bigr|
                   \;=\; \Bigl|\E_{g_\# P_{{\rm T},X}}[\phi_h] - \E_{P_{{\rm S},X}}[\phi_h]\Bigr|,
\end{aligned}
\end{equation*}
and write the population certificate value as
\begin{equation*}
  \tauhard \;:=\; \inf_{(g,h) \in \Gfamily \times \Hfamily}\bigl\{\lamhard(g, h) + \epshard(h) + \Deltahard(g, h)\bigr\}.
\end{equation*}

\begin{theorem}[Population upper certificate for $a_{\mathrm{MR}}$; restatement of \Cref{thm:T2C}]
\label{thm:T2-app}
Suppose $\PT$ and $\PS$ admit measurable Bayes-optimal classifiers $b_{\rm T} : \XT \to [K_{\rm T}]$ and $b_{\rm S} : \XS \to [K_{\rm S}]$.
Then for every fixed measurable $\fs : \XS \to [K_{\rm S}]$, every nonempty class $\Gfamily$ of measurable maps $\XT \to \XS$, and every nonempty $\Hfamily \subseteq [K_{\rm T}]^{[K_{\rm S}]}$,
\begin{equation*}
  a_{\mathrm{MR}} \;\le\; \tauhard \;=\; \inf_{(g,h) \in \Gfamily \times \Hfamily}\bigl\{\lamhard(g, h) + \epshard(h) + \Deltahard(g, h)\bigr\}.
\end{equation*}
\end{theorem}

\begin{proof}
Fix any pair $(g, h) \in \Gfamily \times \Hfamily$, and define the induced hard-label wrapper predictor and its relabeled source-Bayes comparator on $\XT$ by
\begin{equation*}
  f_{g,h}(x) \;:=\; h\bigl(\fs(g(x))\bigr),
  \qquad
  m_{g,h}(x) \;:=\; h\bigl(b_{\rm S}(g(x))\bigr),
  \qquad x \in \XT.
\end{equation*}
For $(X, Y) \sim \PT$ and each $y \in [K_{\rm T}]$, let $\eta_y(X) := \E_{\PT}[\one\{Y = y\} \mid X] \in [0, 1]$ denote a version of the conditional class posterior, so that for any measurable classifier $f : \XT \to [K_{\rm T}]$, $\E_{\PT}[\one\{f(X) \neq Y\} \mid X] = 1 - \eta_{f(X)}(X)$ almost surely.
Because $0 \le \eta_y(X) \le 1$ almost surely for all $y \in [K_{\rm T}]$, the difference $\eta_{b_{\rm T}(X)}(X) - \eta_{f(X)}(X)$ vanishes on $\{f(X) = b_{\rm T}(X)\}$ and is at most $1$ almost surely on $\{f(X) \neq b_{\rm T}(X)\}$, giving $\eta_{b_{\rm T}(X)}(X) - \eta_{f(X)}(X) \le \one\{f(X) \neq b_{\rm T}(X)\}$ almost surely.
Using $\Risk(b_{\rm T}) = \RiskStar$ and conditioning on $X$ therefore yields
\begin{align}
  \Risk(f) - \RiskStar
  &= \Risk(f) - \Risk(b_{\rm T}) \nonumber \\
  &= \E_{X \sim P_{{\rm T},X}}\bigl[1 - \eta_{f(X)}(X)\bigr] - \E_{X \sim P_{{\rm T},X}}\bigl[1 - \eta_{b_{\rm T}(X)}(X)\bigr] \nonumber \\
  &= \E_{X \sim P_{{\rm T},X}}\bigl[\eta_{b_{\rm T}(X)}(X) - \eta_{f(X)}(X)\bigr] \nonumber \\
  &\le \E_{X \sim P_{{\rm T},X}}\bigl[\one\{f(X) \neq b_{\rm T}(X)\}\bigr] \nonumber \\
  &= \Prob_{X \sim P_{{\rm T},X}}\bigl(f(X) \neq b_{\rm T}(X)\bigr),
  \label{eq:bayes-disagreement-hard-expanded}
\end{align}
where the second equality applies the tower property of conditional expectation.

Next, for every $x \in \XT$, if $b_{\rm T}(x) = m_{g,h}(x)$ and $m_{g,h}(x) = f_{g,h}(x)$, then $f_{g,h}(x) = b_{\rm T}(x)$; taking the contrapositive gives the pointwise indicator bound $\one\{f_{g,h}(x) \neq b_{\rm T}(x)\} \le \one\{b_{\rm T}(x) \neq m_{g,h}(x)\} + \one\{f_{g,h}(x) \neq m_{g,h}(x)\}$.
Noting that $\one\{f_{g,h}(x) \neq m_{g,h}(x)\} = \one\{h(\fs(g(x))) \neq h(b_{\rm S}(g(x)))\} = \phi_h(g(x))$ and applying \eqref{eq:bayes-disagreement-hard-expanded} to $f = f_{g,h}$ yields
\begin{align}
  \Risk(f_{g,h}) - \RiskStar
  &\le \Prob_{X \sim P_{{\rm T},X}}\bigl(f_{g,h}(X) \neq b_{\rm T}(X)\bigr) \nonumber \\
  &\le \E_{X \sim P_{{\rm T},X}}\bigl[\one\{b_{\rm T}(X) \neq m_{g,h}(X)\}\bigr]
     + \E_{X \sim P_{{\rm T},X}}\bigl[\one\{f_{g,h}(X) \neq m_{g,h}(X)\}\bigr] \nonumber \\
  &= \lamhard(g, h) + \E_{X \sim P_{{\rm T},X}}\bigl[\phi_h(g(X))\bigr],
  \label{eq:hard-interp-bound}
\end{align}
where the first inequality invokes \eqref{eq:bayes-disagreement-hard-expanded}, the second applies the pointwise indicator triangle bound, and the equality substitutes the definitions of $\lamhard$ and $\phi_h$.

To transfer the second term from $g_\# P_{{\rm T},X}$ to $P_{{\rm S},X}$, apply the pushforward change-of-variables identity together with $t \le |t|$:
\begin{align}
  \E_{X \sim P_{{\rm T},X}}\bigl[\phi_h(g(X))\bigr]
  &= \E_{g_\# P_{{\rm T},X}}[\phi_h] \nonumber \\
  &= \E_{P_{{\rm S},X}}[\phi_h] + \Bigl(\E_{g_\# P_{{\rm T},X}}[\phi_h] - \E_{P_{{\rm S},X}}[\phi_h]\Bigr) \nonumber \\
  &\le \E_{P_{{\rm S},X}}[\phi_h] + \Bigl|\E_{g_\# P_{{\rm T},X}}[\phi_h] - \E_{P_{{\rm S},X}}[\phi_h]\Bigr| \nonumber \\
  &= \epshard(h) + \Deltahard(g, h),
  \label{eq:pushforward-transfer-hard}
\end{align}
where the final equality substitutes $\E_{P_{{\rm S},X}}[\phi_h] = \epshard(h)$ and $|\E_{g_\# P_{{\rm T},X}}[\phi_h] - \E_{P_{{\rm S},X}}[\phi_h]| = \Deltahard(g, h)$.
Substituting \eqref{eq:pushforward-transfer-hard} into \eqref{eq:hard-interp-bound} gives $\Risk(f_{g,h}) - \RiskStar \le \lamhard(g, h) + \epshard(h) + \Deltahard(g, h)$ for every $(g, h) \in \Gfamily \times \Hfamily$.
Taking the infimum over the nonempty product $\Gfamily \times \Hfamily$ yields
\begin{align*}
  a_{\mathrm{MR}}
  &= \RiskClsOptimum - \RiskStar
    &&\text{(by \Cref{def:errors})} \\
  &= \inf_{(g,h) \in \Gfamily \times \Hfamily}\bigl\{\Risk(f_{g,h}) - \RiskStar\bigr\}
    &&\text{($\Gfamily, \Hfamily \neq \varnothing$)} \\
  &\le \inf_{(g,h) \in \Gfamily \times \Hfamily}\bigl\{\lamhard(g, h) + \epshard(h) + \Deltahard(g, h)\bigr\} \;=\; \tauhard,
\end{align*}
which establishes $a_{\mathrm{MR}} \le \tauhard$.
\end{proof}

\begin{corollary}[Combined finite-sample Bayes excess-risk certificate]
\label{cor:T2-combined-pac}
Under the hypotheses of \Cref{thm:T1-app,thm:T2-app}, for every $\nT \ge 1$, every $\delta \in (0, 1)$, and every $\eta_n \ge 0$, with probability at least $1 - \delta$ over an i.i.d.\ target sample of size $\nT$ drawn from $\PT$, every measurable $\eta_n$-approximate ERM $\widehat f_n \in \FMR$ satisfies
\begin{equation*}
  \Risk(\widehat f_n) - \RiskStar
  \;\le\; \tauhard + \eta_n + 4\sqrt{6}\,\sqrt{\frac{\log\PihatR(2\nT) + \log\PihatH(2\nT) + \log(1/\delta)}{\nT}}.
\end{equation*}
\end{corollary}

\begin{proof}
On the $(1 - \delta)$-probability event of \Cref{thm:T1-app} (\Cref{rem:T1-bayes-excess-consequence}), every measurable $\eta_n$-approximate ERM $\widehat f_n \in \FMR$ satisfies
\begin{align*}
  \Risk(\widehat f_n) - \RiskStar
  &= \EstRisk(\widehat f_n) + a_{\mathrm{MR}} \\
  &\le a_{\mathrm{MR}} + \eta_n + 4\sqrt{6}\,\sqrt{\frac{\log\PihatR(2\nT) + \log\PihatH(2\nT) + \log(1/\delta)}{\nT}} \\
  &\le \tauhard + \eta_n + 4\sqrt{6}\,\sqrt{\frac{\log\PihatR(2\nT) + \log\PihatH(2\nT) + \log(1/\delta)}{\nT}},
\end{align*}
where the equality applies \Cref{def:errors}, the first inequality invokes \Cref{thm:T1-app}, and the second substitutes $a_{\mathrm{MR}} \le \tauhard$ from \Cref{thm:T2-app}.
\end{proof}

\begin{remark}[Proof-level origin of the two converse obstructions in \Cref{rem:one-way-certificate}]
\label{rem:T2-converse-obstructions}
The proof of \Cref{thm:T2-app} localizes the two inequalities that prevent a converse lower bound in \Cref{rem:one-way-certificate}.
First, \eqref{eq:bayes-disagreement-hard-expanded} bounds the pointwise posterior regret $\eta_{b_{\rm T}(X)}(X) - \eta_{f_{g,h}(X)}(X)$ from above by the binary disagreement indicator $\one\{f_{g,h}(X) \neq b_{\rm T}(X)\}$.
Whenever the target posterior $\eta_{\cdot}(X)$ is nearly uniform across top competing classes, $f_{g,h}$ can disagree with $b_{\rm T}$ with probability bounded away from zero while incurring arbitrarily small $0$--$1$ excess risk, so reversing \eqref{eq:bayes-disagreement-hard-expanded} requires an additional low-noise margin condition on $\PT$.
Second, \eqref{eq:pushforward-transfer-hard} transfers $\phi_h$ between $g_\# P_{{\rm T},X}$ and $P_{{\rm S},X}$ through the pointwise pushforward discrepancy $\Deltahard(g, h) = \bigl|\E_{g_\# P_{{\rm T},X}}[\phi_h] - \E_{P_{{\rm S},X}}[\phi_h]\bigr|$ evaluated at the single tested head $h$.
Evaluating $\phi_h$ at a fixed $h$ makes $\Deltahard(g, h)$ estimable from finite unlabeled samples once $b_{\rm S}$ is plugged in, whereas symmetric domain-adaptation divergences~\citep{ben2010theory} take a supremum over $\Hfamily \Delta \Hfamily$ and sacrifice direct pointwise estimability.
\end{remark}

%% file: appendix/proof_T2P_C2prime.tex
\label{app:score-certificate-proofs}

\textbf{Proof roadmap.}
This section proves the pointwise population surrogate certificate (\Cref{thm:T2P-app}, restating \Cref{thm:score-certificate}) and its simultaneous finite-library empirical counterpart (\Cref{cor:C2prime-app}, restating \Cref{cor:score-finite-library}).
We first establish an auxiliary lemma (\Cref{lem:score-envelopes}) proving measurability of the lexicographic plug-in quantities ($\hat b_{\rm S}$, $c$, and $\zeta_W$) together with the Euclidean ball containment $W s_{\rm S}(u), W c(u) \in B(0, R_{U1})$ and finite range envelopes ($B_\lambda$ and $B_\zeta$) for all $B_s, B_h, L_\ell \in [0, \infty)$.
\Cref{thm:T2P-app} invokes \Cref{lem:score-envelopes} to bound the target surrogate risk via one-hot plug-in interpolation, transfers $\zeta_W$ from the pushforward target marginal $\mu_g = g_\# P_{{\rm T}, X}$ to the source marginal $P_{{\rm S}, X}$, uses $\LossStar \ge 0$ to upper-bound surrogate excess risk, and applies the nondecreasing multiclass regret transform $\psi_\ell^{-1}$ to obtain the calibrated $0$--$1$ excess-risk bound.
Finally, \Cref{cor:C2prime-app} combines a deterministic four-term comparison between $\tauscore(g, W)$ and $\widehat{\tauscore}(g, W)$ (\eqref{eq:C2prime-det-slack}, which accounts for the factor $2$ on the source-sample deviation) with upfront zero-envelope checks, two-sided Hoeffding concentration, and a union bound across all $3M$ scalar sample means.

\subsection{Auxiliary measurability and envelope lemma}

Recall from \Cref{app:scoreoutput:certificate} that for measurable $(g, W) \in \Gfamily \times \Wclass$, the score-induced source classifier is $\hat b_{\rm S}(u) := \lexargmax_{k \in [K_{\rm S}]} s_{{\rm S}, k}(u)$, its one-hot embedding is $c(u) := e_{\hat b_{\rm S}(u)} \in \mathbb{R}^{K_{\rm S}}$, the pushforward target marginal is $\mu_g := g_\# P_{{\rm T}, X}$, and the population certificate is $\tauscore(g, W) := \lamscore(g, W) + \epsscore(W) + \Deltascore(g, W)$.
Before proving \Cref{thm:T2P-app,cor:C2prime-app} (and the uniform Rademacher certificate in \Cref{app:rademacher}), we record the measurability and range envelopes of the underlying plug-in and distortion maps across all $B_s, B_h, L_\ell \in [0, \infty)$.

\begin{lemma}[Measurability and range envelopes of score-output plug-in quantities]
\label{lem:score-envelopes}
Let $\XS$ be a measurable space, let $K_{\rm S}, K_{\rm T} \ge 1$ be integers, and let $B_s, B_h, L_\ell \in [0, \infty)$.
Suppose $s_{\rm S} : \XS \to [-B_s, B_s]^{K_{\rm S}}$ is measurable, $W \in \mathbb{R}^{K_{\rm T} \times K_{\rm S}}$ satisfies $\|W\|_F \le B_h$, and $\ell : [K_{\rm T}] \times \mathbb{R}^{K_{\rm T}} \to [0, \infty)$ is measurable and $L_\ell$-Lipschitz in its second argument with respect to $\|\cdot\|_2$ on the closed Euclidean ball $B(0, R_{U1}) \subseteq \mathbb{R}^{K_{\rm T}}$ of radius $R_{U1} := B_h\max\{B_s\sqrt{K_{\rm S}}, 1\}$.
Set $\hat b_{\rm S}(u) := \lexargmax_{k \in [K_{\rm S}]} s_{{\rm S}, k}(u)$, $c(u) := e_{\hat b_{\rm S}(u)} \in \mathbb{R}^{K_{\rm S}}$, and $\zeta_W(u) := \max_{y \in [K_{\rm T}]}\bigl|\ell(y, W s_{\rm S}(u)) - \ell(y, W c(u))\bigr|$ for $u \in \XS$, and define
\begin{equation*}
  B_\zeta \;:=\; L_\ell B_h\bigl(B_s\sqrt{K_{\rm S}} + 1\bigr),
  \qquad
  B_\lambda \;:=\; \sup_{y \in [K_{\rm T}],\, q \in B(0, R_{U1})} \ell(y, q).
\end{equation*}
Then:
\begin{enumerate}
  \item[(i)] $\hat b_{\rm S} : \XS \to [K_{\rm S}]$, $c : \XS \to \mathbb{R}^{K_{\rm S}}$, and $\zeta_W : \XS \to [0, \infty)$ are measurable, and $B_\lambda \in [0, \infty)$;
  \item[(ii)] $W s_{\rm S}(u) \in B(0, R_{U1})$ and $W c(u) \in B(0, R_{U1})$ for every $u \in \XS$; and
  \item[(iii)] $\ell(y, W c(u)) \in [0, B_\lambda]$ and $\zeta_W(u) \in [0, B_\zeta]$ for every $(y, u) \in [K_{\rm T}] \times \XS$.
\end{enumerate}
\end{lemma}

\begin{proof}
For claim~(i), deterministic lexicographic tie-breaking (\Cref{app:assumptions}) writes each level set of $\hat b_{\rm S}$ as
\begin{equation*}
  \bigl\{u \in \XS : \hat b_{\rm S}(u) = k\bigr\}
  \;=\;
  \bigcap_{r < k}\bigl\{u \in \XS : s_{{\rm S}, k}(u) > s_{{\rm S}, r}(u)\bigr\}
  \;\cap\;
  \bigcap_{r > k}\bigl\{u \in \XS : s_{{\rm S}, k}(u) \ge s_{{\rm S}, r}(u)\bigr\}
\end{equation*}
for every $k \in [K_{\rm S}]$.
Because the coordinate functions $s_{{\rm S}, 1}, \ldots, s_{{\rm S}, K_{\rm S}}$ are measurable, each level set is a finite intersection of measurable sets, so $\hat b_{\rm S}$ and its one-hot embedding $c(u) = \sum_{k=1}^{K_{\rm S}} \one\{\hat b_{\rm S}(u) = k\}\, e_k$ are measurable.
Composing the continuous linear map $v \mapsto W v$ and the measurable function $q \mapsto \ell(y, q)$ with $s_{\rm S}$ and $c$ shows that $u \mapsto \ell(y, W s_{\rm S}(u))$ and $u \mapsto \ell(y, W c(u))$ are measurable for every $y \in [K_{\rm T}]$; taking absolute differences and the pointwise maximum over the finite set $[K_{\rm T}]$ establishes measurability of $\zeta_W$.
Furthermore, for each $y \in [K_{\rm T}]$, $L_\ell$-Lipschitz continuity on $B(0, R_{U1})$ implies continuity of $q \mapsto \ell(y, q)$ on the nonempty compact ball $B(0, R_{U1})$ (including $B(0, 0) = \{0\}$ when $R_{U1} = 0$), so $\sup_{q \in B(0, R_{U1})} \ell(y, q) < \infty$; taking the maximum over the finite alphabet $[K_{\rm T}]$ and using $\ell \ge 0$ gives $B_\lambda \in [0, \infty)$.

For claim~(ii), $s_{\rm S}(u) \in [-B_s, B_s]^{K_{\rm S}}$ implies $\|s_{\rm S}(u)\|_2 = \bigl(\sum_{k=1}^{K_{\rm S}} s_{{\rm S}, k}(u)^2\bigr)^{1/2} \le B_s\sqrt{K_{\rm S}}$, while $\|c(u)\|_2 = \|e_{\hat b_{\rm S}(u)}\|_2 = 1$ and $\|W\|_{\mathrm{op}} \le \|W\|_F \le B_h$.
For every $u \in \XS$, submultiplicativity of the operator norm yields
\begin{align*}
  \|W s_{\rm S}(u)\|_2
  &\le \|W\|_{\mathrm{op}}\,\|s_{\rm S}(u)\|_2
    &&\text{(operator-norm bound)} \\
  &\le B_h B_s\sqrt{K_{\rm S}}
    &&\text{(norm bounds)} \\
  &\le B_h\max\bigl\{B_s\sqrt{K_{\rm S}},\, 1\bigr\} \;=\; R_{U1},
\end{align*}
and similarly
\begin{align*}
  \|W c(u)\|_2
  &\le \|W\|_{\mathrm{op}}\,\|c(u)\|_2
    &&\text{(operator-norm bound)} \\
  &\le B_h
    &&\text{(norm bounds)} \\
  &\le B_h\max\bigl\{B_s\sqrt{K_{\rm S}},\, 1\bigr\} \;=\; R_{U1},
\end{align*}
so both $W s_{\rm S}(u)$ and $W c(u)$ lie in $B(0, R_{U1})$.

For claim~(iii), nonnegativity of $\ell$ and $W c(u) \in B(0, R_{U1})$ from claim~(ii) immediately imply $0 \le \ell(y, W c(u)) \le B_\lambda$ for every $(y, u) \in [K_{\rm T}] \times \XS$.
Likewise, because both $W s_{\rm S}(u)$ and $W c(u)$ belong to $B(0, R_{U1})$, $L_\ell$-Lipschitz continuity of $\ell(y, \cdot)$ on $B(0, R_{U1})$ gives, for every $u \in \XS$,
\begin{align*}
  0 \;\le\; \zeta_W(u)
  &= \max_{y \in [K_{\rm T}]}\bigl|\ell\bigl(y, W s_{\rm S}(u)\bigr) - \ell\bigl(y, W c(u)\bigr)\bigr| \\
  &\le L_\ell\,\bigl\|W s_{\rm S}(u) - W c(u)\bigr\|_2
    \quad\text{($L_\ell$-Lipschitz on $B(0, R_{U1})$)} \\
  &\le L_\ell\,\|W\|_{\mathrm{op}}\,\|s_{\rm S}(u) - c(u)\|_2
    \quad\text{(operator-norm bound)} \\
  &\le L_\ell\,\|W\|_{\mathrm{op}}\bigl(\|s_{\rm S}(u)\|_2 + \|c(u)\|_2\bigr)
    \quad\text{(triangle inequality)} \\
  &\le L_\ell B_h\bigl(B_s\sqrt{K_{\rm S}} + 1\bigr) \;=\; B_\zeta,
\end{align*}
where the final inequality substitutes $\|W\|_{\mathrm{op}} \le B_h$, $\|s_{\rm S}(u)\|_2 \le B_s\sqrt{K_{\rm S}}$, and $\|c(u)\|_2 = 1$.
\end{proof}

\subsection{Proof of \Cref{thm:score-certificate}}

\begin{theorem}[Source-conditional surrogate certificate; restatement of \Cref{thm:score-certificate}]
\label{thm:T2P-app}
For every target distribution $\PT$ over $\XT \times [K_{\rm T}]$, every source-input marginal $P_{{\rm S}, X}$ on $\XS$, and constants $B_s, B_h, L_\ell \in [0, \infty)$, suppose $s_{\rm S}: \XS \to [-B_s, B_s]^{K_{\rm S}}$ is measurable, every $W \in \Wclass$ satisfies $\|W\|_F \le B_h$, and the measurable surrogate loss $\ell : [K_{\rm T}] \times \mathbb{R}^{K_{\rm T}} \to [0, \infty)$ is $L_\ell$-Lipschitz in its second argument with respect to $\|\cdot\|_2$ on the closed Euclidean ball $B(0, R_{U1}) \subseteq \mathbb{R}^{K_{\rm T}}$ of radius $R_{U1} := B_h \max\{B_s \sqrt{K_{\rm S}}, 1\}$ (\texttt{A8}--\texttt{A11} in \Cref{tab:assumptions}).
Then for every measurable pair $(g, W) \in \Gfamily \times \Wclass$,
\begin{equation*}
  \Loss(g, W) - \LossStar
  \;\le\;
  \tauscore(g, W)
  \;=\;
  \lamscore(g, W) + \epsscore(W) + \Deltascore(g, W).
\end{equation*}
Consequently, whenever $\ell$ is multiclass classification-calibrated with nondecreasing regret transform $\psi_\ell^{-1}: [0, \infty) \to [0, \infty)$,
\begin{equation*}
  \Risk(f_{g, W}) - \RiskStar
  \;\le\;
  \psi_\ell^{-1}\!\bigl(\tauscore(g, W)\bigr)
  \;=\;
  \psi_\ell^{-1}\!\bigl(\lamscore(g, W) + \epsscore(W) + \Deltascore(g, W)\bigr).
\end{equation*}
\end{theorem}

\begin{proof}
Fix a measurable pair $(g, W) \in \Gfamily \times \Wclass$ and write $\mu_g := g_\# P_{{\rm T}, X}$.
By \Cref{lem:score-envelopes}, $\hat b_{\rm S}$, $c$, and $\zeta_W$ are measurable, $W s_{\rm S}(u) \in B(0, R_{U1})$, $\ell(y, W c(u)) \in [0, B_\lambda]$, and $\zeta_W(u) \in [0, B_\zeta]$ for all $(y, u) \in [K_{\rm T}] \times \XS$.
In particular, $\ell(y, W s_{\rm S}(u)) \in [0, B_\lambda]$ as well, so $\Loss(g, W)$, $\lamscore(g, W)$, $\E_{\mu_g}[\zeta_W]$, and $\epsscore(W) = \E_{P_{{\rm S}, X}}[\zeta_W]$ are all finite.

By definition of $\zeta_W$, every $(y, u) \in [K_{\rm T}] \times \XS$ satisfies $\ell(y, W s_{\rm S}(u)) \le \ell(y, W c(u)) + \zeta_W(u)$.
Moreover, pointwise nonnegativity $\ell \ge 0$ implies $\LossStar = \inf_q \E_{(X, Y) \sim \PT}[\ell(Y, q(X))] \ge 0$.
Evaluating the plug-in bound at $(Y, g(X))$, taking expectation under $(X, Y) \sim \PT$, and transferring $\zeta_W$ from $\mu_g$ to $P_{{\rm S}, X}$ gives
\begin{align*}
  \Loss(g, W) - \LossStar
  &\le \Loss(g, W) \\
  &= \E_{(X, Y) \sim \PT}\bigl[\ell\bigl(Y, W s_{\rm S}(g(X))\bigr)\bigr] \\
  &\le \E_{(X, Y) \sim \PT}\bigl[\ell\bigl(Y, W c(g(X))\bigr)\bigr]
     + \E_{X \sim P_{{\rm T}, X}}\bigl[\zeta_W(g(X))\bigr] \\
  &= \lamscore(g, W) + \E_{\mu_g}[\zeta_W] \\
  &= \lamscore(g, W) + \E_{P_{{\rm S}, X}}[\zeta_W]
     + \bigl(\E_{\mu_g}[\zeta_W] - \E_{P_{{\rm S}, X}}[\zeta_W]\bigr) \\
  &\le \lamscore(g, W) + \E_{P_{{\rm S}, X}}[\zeta_W]
     + \bigl|\E_{\mu_g}[\zeta_W] - \E_{P_{{\rm S}, X}}[\zeta_W]\bigr| \\
  &= \lamscore(g, W) + \epsscore(W) + \Deltascore(g, W)
  \;=\; \tauscore(g, W),
\end{align*}
where the first inequality uses $\LossStar \ge 0$, the second uses $\ell(Y, W s_{\rm S}(g(X))) \le \ell(Y, W c(g(X))) + \zeta_W(g(X))$, the second equality applies the pushforward identity $\E_{X \sim P_{{\rm T}, X}}[\zeta_W(g(X))] = \E_{\mu_g}[\zeta_W]$, and the third inequality uses $t \le |t|$.

When $\ell$ is multiclass classification-calibrated~\citep{zhang2004statistical, tewari2007consistency}, the score-induced predictor $f_{g, W}(x) = \lexargmax_{y \in [K_{\rm T}]} (W s_{\rm S}(g(x)))_y$ satisfies $\Risk(f_{g, W}) - \RiskStar \le \psi_\ell^{-1}\bigl(\Loss(g, W) - \LossStar\bigr)$.
Since $\psi_\ell^{-1} : [0, \infty) \to [0, \infty)$ is nondecreasing and $0 \le \Loss(g, W) - \LossStar \le \tauscore(g, W)$, applying $\psi_\ell^{-1}$ yields $\Risk(f_{g, W}) - \RiskStar \le \psi_\ell^{-1}\bigl(\tauscore(g, W)\bigr)$.
\end{proof}

\begin{remark}[One-directional nature and family-level infimum]
Because the proof of \Cref{thm:T2P-app} lower-bounds $\LossStar$ by $0$ and bounds surrogate differences via the worst-label envelope $\zeta_W$, \Cref{thm:T2P-app} is a one-directional upper certificate rather than a two-sided characterization.
For nonempty $\Gfamily$ and $\Wclass$, setting $\tau_\ast := \inf_{(g, W) \in \Gfamily \times \Wclass} \tauscore(g, W)$ yields $a_{\mathrm{score}, \ell} \le \tau_\ast$ and $a_{\mathrm{score}, 0/1} \le \inf_{\varepsilon > 0}\psi_\ell^{-1}(\tau_\ast + \varepsilon)$; this right-hand limit equals $\psi_\ell^{-1}(\tau_\ast)$ when $\psi_\ell^{-1}$ is right-continuous at $\tau_\ast$, while $a_{\mathrm{score}, 0/1} \le \psi_\ell^{-1}(\tau_\ast)$ also follows directly whenever $\tau_\ast$ is attained at some $(g^\star, W^\star) \in \Gfamily \times \Wclass$ by evaluating \Cref{thm:T2P-app} at $(g^\star, W^\star)$.
\end{remark}

\subsection{Proof of \Cref{cor:score-finite-library}}

\begin{corollary}[Finite-library empirical certificate; restatement of \Cref{cor:score-finite-library}]
\label{cor:C2prime-app}
Under the assumptions of \Cref{thm:T2P-app}, set $B_\zeta := L_\ell B_h(B_s\sqrt{K_{\rm S}} + 1)$ and $B_\lambda := \sup_{y \in [K_{\rm T}],\, q \in B(0, R_{U1})} \ell(y, q) < \infty$.
For every integer $M \ge 1$, every finite candidate library $Q = \{(g_r, W_r)\}_{r=1}^M \subseteq \Gfamily \times \Wclass$ of measurable pairs, all sample sizes $\nT, n_{\rm S} \ge 1$, and every $\delta \in (0, 1)$, if $D_{\rm T} \sim \PT^{\nT}$ and $D_{\rm S} \sim P_{{\rm S}, X}^{n_{\rm S}}$ are mutually independent i.i.d.\ samples drawn independently of $Q$, then with probability at least $1 - \delta$ over $(D_{\rm T}, D_{\rm S})$, simultaneously for all $(g, W) \in Q$,
\begin{equation*}
  \Loss(g, W) - \LossStar
  \;\le\;
  \widehat{\tauscore}(g, W)
  + B_\lambda \sqrt{\frac{\log(6 M / \delta)}{2 \nT}}
  + B_\zeta \sqrt{\frac{\log(6 M / \delta)}{2 \nT}}
  + 2 B_\zeta \sqrt{\frac{\log(6 M / \delta)}{2 n_{\rm S}}}.
\end{equation*}
Consequently, whenever $\ell$ is multiclass classification-calibrated with nondecreasing regret transform $\psi_\ell^{-1}: [0, \infty) \to [0, \infty)$, on the same $(1 - \delta)$-probability event, simultaneously for all $(g, W) \in Q$,
\begin{equation*}
  \Risk(f_{g, W}) - \RiskStar
  \;\le\;
  \psi_\ell^{-1}\!\left(
    \widehat{\tauscore}(g, W)
    + (B_\lambda + B_\zeta)\sqrt{\frac{\log(6 M / \delta)}{2 \nT}}
    + 2 B_\zeta \sqrt{\frac{\log(6 M / \delta)}{2 n_{\rm S}}}
  \right).
\end{equation*}
\end{corollary}

\begin{proof}
For any candidate $(g, W) \in Q$, abbreviate its three population expectations and empirical sample means from \Cref{app:scoreoutput:empirical} by
\begin{align*}
  \lambda &:= \lamscore(g, W), &
  T &:= \E_{X \sim P_{{\rm T}, X}}\bigl[\zeta_W(g(X))\bigr], &
  S &:= \E_{U \sim P_{{\rm S}, X}}\bigl[\zeta_W(U)\bigr], \\
  \widehat\lambda &:= \widehat{\lamscore}(g, W), &
  \widehat T &:= \widehat T(g, W), &
  \widehat S &:= \widehat S(W),
\end{align*}
so that $\tauscore(g, W) = \lambda + S + |T - S|$ and $\widehat{\tauscore}(g, W) = \widehat\lambda + \widehat S + |\widehat T - \widehat S|$.
Bounding the difference deterministically via the scalar triangle inequality gives
\begin{align*}
  \tauscore(g, W) - \widehat{\tauscore}(g, W)
  &= (\lambda - \widehat\lambda) + (S - \widehat S) + \bigl(|T - S| - |\widehat T - \widehat S|\bigr) \\
  &\le |\lambda - \widehat\lambda| + |S - \widehat S| + \bigl|(T - S) - (\widehat T - \widehat S)\bigr| \\
  &= |\lambda - \widehat\lambda| + |S - \widehat S| + \bigl|(T - \widehat T) - (S - \widehat S)\bigr| \\
  &\le |\lambda - \widehat\lambda| + |T - \widehat T| + 2|S - \widehat S|,
\end{align*}
where the first inequality uses $u \le |u|$ and $||a| - |b|| \le |a - b|$, and the second applies the triangle inequality, rearranging to the deterministic certificate comparison
\begin{equation}
\label{eq:C2prime-det-slack}
  \tauscore(g, W) \;\le\; \widehat{\tauscore}(g, W) + |\lambda - \widehat\lambda| + |T - \widehat T| + 2|S - \widehat S|,
\end{equation}
where the factor $2$ on $|S - \widehat S|$ arises because $\widehat S$ estimates both $\epsscore(W) = S$ and the source-side reference inside $\Deltascore(g, W) = |T - S|$.

Now condition on $Q$ if $Q$ is constructed on an independent holdout split, so that every $(g_r, W_r) \in Q$ ($r \in [M]$) is fixed and independent of $(D_{\rm T}, D_{\rm S})$.
By \Cref{lem:score-envelopes}, for each fixed $(g_r, W_r) \in Q$, the $\nT$ summands $\ell(Y_i, W_r c(g_r(X_i)))$ in $\widehat{\lamscore}(g_r, W_r)$ are i.i.d.\ in $[0, B_\lambda]$, the $\nT$ summands $\zeta_{W_r}(g_r(X_i))$ in $\widehat T(g_r, W_r)$ are i.i.d.\ in $[0, B_\zeta]$, and the $n_{\rm S}$ summands $\zeta_{W_r}(U_j)$ in $\widehat S(W_r)$ are i.i.d.\ in $[0, B_\zeta]$.
For any integer $n \ge 1$ and envelope $B \ge 0$, set $t_n(B) := B\sqrt{\log(6M/\delta)/(2n)}$, and let $V_1, \ldots, V_n \in [0, B]$ be i.i.d.\ with mean $\mu = \E[V_1]$ and empirical mean $\widehat\mu = n^{-1}\sum_{i=1}^n V_i$.
In the degenerate zero-envelope case $B = 0$ (namely $B_\lambda = 0$ for $\widehat{\lamscore}$, or $B_\zeta = 0$ for $\widehat T$ and $\widehat S$), every $V_i$ vanishes identically, so $|\mu - \widehat\mu| = 0$ almost surely and $\Prob(|\mu - \widehat\mu| > t_n(0)) = \Prob(0 > 0) = 0 \le \delta/(3M)$.
When $B > 0$, two-sided Hoeffding's inequality on $[0, B]$ yields
\begin{align*}
  \Prob\bigl(|\mu - \widehat\mu| > t_n(B)\bigr)
  &\le 2\exp\!\left(-\frac{2n\, t_n(B)^2}{B^2}\right)
    &&\text{(Hoeffding's inequality)} \\
  &= 2\exp\!\bigl(-\log(6M/\delta)\bigr) \;=\; \frac{\delta}{3M}.
\end{align*}
Writing $T(g_r, W_r) := \E_{X \sim P_{{\rm T}, X}}[\zeta_{W_r}(g_r(X))]$ and $S(W_r) := \E_{U \sim P_{{\rm S}, X}}[\zeta_{W_r}(U)]$, define for each $r \in [M]$ the three deviation events
\begin{align*}
  E_{r, 1} &:= \bigl\{\bigl|\lamscore(g_r, W_r) - \widehat{\lamscore}(g_r, W_r)\bigr| > t_{\nT}(B_\lambda)\bigr\}, \\
  E_{r, 2} &:= \bigl\{\bigl|T(g_r, W_r) - \widehat T(g_r, W_r)\bigr| > t_{\nT}(B_\zeta)\bigr\}, \qquad
  E_{r, 3} := \bigl\{\bigl|S(W_r) - \widehat S(W_r)\bigr| > t_{n_{\rm S}}(B_\zeta)\bigr\}.
\end{align*}
In both zero- and positive-envelope regimes, $\Prob(E_{r, j}) \le \delta/(3M)$ for all $r \in [M]$ and $j \in \{1, 2, 3\}$, so a union bound across all $3M$ scalar means gives
\begin{align*}
  \Prob\!\left(\bigcup_{r=1}^{M}\bigl(E_{r, 1} \cup E_{r, 2} \cup E_{r, 3}\bigr)\right)
  &\le \sum_{r=1}^{M}\Bigl(\Prob(E_{r, 1}) + \Prob(E_{r, 2}) + \Prob(E_{r, 3})\Bigr)
    &&\text{(union bound)} \\
  &\le \sum_{r=1}^{M} 3 \cdot \frac{\delta}{3M} \;=\; \delta.
\end{align*}

On the complementary event of probability at least $1 - \delta$, combining \Cref{thm:T2P-app} with \eqref{eq:C2prime-det-slack} simultaneously for every $(g, W) \in Q$ gives
\begin{align*}
  \Loss(g, W) - \LossStar
  &\le \tauscore(g, W)
    \quad\text{(by \Cref{thm:T2P-app})} \\
  &\le \widehat{\tauscore}(g, W) + \bigl|\lamscore(g, W) - \widehat{\lamscore}(g, W)\bigr| \\
  &\quad + \bigl|T(g, W) - \widehat T(g, W)\bigr| + 2\bigl|S(W) - \widehat S(W)\bigr|
    \quad\text{(by \eqref{eq:C2prime-det-slack})} \\
  &\le \widehat{\tauscore}(g, W) + t_{\nT}(B_\lambda) + t_{\nT}(B_\zeta) + 2\, t_{n_{\rm S}}(B_\zeta),
\end{align*}
which establishes the simultaneous surrogate certificate.
When $\ell$ is multiclass classification-calibrated, applying the calibrated excess-risk bound of \Cref{thm:T2P-app} together with monotonicity of $\psi_\ell^{-1}: [0, \infty) \to [0, \infty)$ on the same $(1 - \delta)$-probability event yields the simultaneous $0$--$1$ excess-risk bound.
\end{proof}

\begin{remark}[Evaluation-independence requirement]
Simultaneity in \Cref{cor:C2prime-app} holds over the evaluation-independent library $Q$; whenever candidate wrappers are trained or selected using data, $D_{\rm T}$ and $D_{\rm S}$ must be held out from that selection step (for example via sample splitting or cross-fitting).
\end{remark}

%% file: appendix/rademacher_certificate.tex
\label{app:rademacher}

\textbf{Proof roadmap.}
This section proves the uniform empirical score-output certificate over a fixed (possibly infinite) candidate set (\Cref{thm:rad-score-output-certificate}, restating \Cref{thm:score-rademacher}), replacing the $\log(6M/\delta)$ finite-library union bound of \Cref{cor:C2prime-app} with empirical Rademacher complexities of three induced scalar classes.
\Cref{app:rad-assumptions} records the pointwise-measurability, range-envelope, and evaluation-independence conditions on $Q \subseteq \Gfamily \times \Wclass$ and $(D_{\rm T}, D_{\rm S})$.
\Cref{app:rad-certificate} first establishes a self-contained two-sided empirical Rademacher uniform-deviation lemma (\Cref{lem:two-sided-rad}), then applies it at failure parameter $\delta' = \delta/6$ across $(\mathcal{L}_Q, \mathcal{Z}_{{\rm T}, Q}, \mathcal{Z}_{{\rm S}, Q})$ together with \Cref{lem:score-envelopes}, the deterministic comparison \eqref{eq:C2prime-det-slack}, and \Cref{thm:T2P-app} to prove \Cref{thm:rad-score-output-certificate}, and concludes with optional vector-contraction refinements via \citet{maurer2016vector}.

Let $Q \subseteq \Gfamily \times \Wclass$ be a nonempty set of measurable candidate pairs fixed independently of the certification samples (or constructed on an independent holdout split so that it is independent of the certification folds), and denote its head projection by
\begin{equation*}
  \Wclass_Q \;:=\; \bigl\{W \in \Wclass : \exists\, g \in \Gfamily \text{ with } (g, W) \in Q\bigr\}.
\end{equation*}
Let $D_{\rm T} = \{(X_i, Y_i)\}_{i=1}^{\nT} \sim \PT^{\nT}$ (with target-input tuple $X_{1:\nT} := (X_1, \ldots, X_{\nT})$) and $D_{\rm S} = \{U_j\}_{j=1}^{n_{\rm S}} \sim P_{{\rm S}, X}^{n_{\rm S}}$ ($\nT, n_{\rm S} \ge 1$) be mutually independent i.i.d.\ samples drawn independently of $Q$.
As in \Cref{app:scoreoutput:empirical}, define the three induced scalar function classes
\begin{align*}
  \mathcal{L}_Q &:= \bigl\{(x, y) \mapsto \ell\bigl(y, W c(g(x))\bigr) : (g, W) \in Q\bigr\}, \\
  \mathcal{Z}_{{\rm T}, Q} &:= \bigl\{x \mapsto \zeta_W\bigl(g(x)\bigr) : (g, W) \in Q\bigr\}, \\
  \mathcal{Z}_{{\rm S}, Q} &:= \bigl\{u \mapsto \zeta_W(u) : (g, W) \in Q\bigr\} \;=\; \bigl\{u \mapsto \zeta_W(u) : W \in \Wclass_Q\bigr\},
\end{align*}
where $\hat b_{\rm S}(u) := \lexargmax_{k \in [K_{\rm S}]} s_{{\rm S}, k}(u)$ and $c(u) := e_{\hat b_{\rm S}(u)}$.
For a scalar function class $\mathcal{F}$ and sample $Z_{1:n} = (Z_1, \ldots, Z_n)$, write its empirical Rademacher complexity as
\begin{equation*}
  \widehat{\mathfrak{R}}_{Z_{1:n}}(\mathcal{F}) \;:=\; \frac{1}{n}\,\E_\sigma\!\left[\sup_{f \in \mathcal{F}} \sum_{i=1}^{n} \sigma_i f(Z_i) \;\middle|\; Z_{1:n}\right],
\end{equation*}
where $\sigma_1, \ldots, \sigma_n \in \{-1, +1\}$ are independent uniform Rademacher signs independent of $Z_{1:n}$.

\subsection{Assumptions for the uniform empirical certificate}
\label{app:rad-assumptions}

Assume $\mathcal{L}_Q$, $\mathcal{Z}_{{\rm T}, Q}$, and $\mathcal{Z}_{{\rm S}, Q}$ are pointwise measurable (i.e., admitting a countable subset dense under pointwise convergence, as in \Cref{app:assumptions}) so that their empirical suprema and Rademacher averages are measurable and symmetrization applies.
Assume the boundedness and surrogate-regularity hypotheses of \Cref{thm:T2P-app} (\Cref{thm:score-certificate}): for constants $B_s, B_h, L_\ell \in [0, \infty)$, $s_{\rm S} : \XS \to [-B_s, B_s]^{K_{\rm S}}$ is measurable, every $W \in \Wclass$ (and hence every $W \in \Wclass_Q$) satisfies $\|W\|_F \le B_h$, and $\ell : [K_{\rm T}] \times \mathbb{R}^{K_{\rm T}} \to [0, \infty)$ is measurable and $L_\ell$-Lipschitz in its second argument with respect to $\|\cdot\|_2$ on $B(0, R_{U1})$ with $R_{U1} := B_h\max\{B_s\sqrt{K_{\rm S}}, 1\}$, giving the finite envelopes $B_\lambda := \sup_{y \in [K_{\rm T}],\, q \in B(0, R_{U1})} \ell(y, q)$ and $B_\zeta := L_\ell B_h(B_s\sqrt{K_{\rm S}} + 1)$ by \Cref{lem:score-envelopes}.
Assume $(D_{\rm T}, D_{\rm S})$ and $Q$ satisfy the mutual- and evaluation-independence conditions stated at the start of \Cref{app:rademacher}.

\subsection{Uniform empirical certificate}
\label{app:rad-certificate}

We first establish the two-sided empirical Rademacher concentration inequality invoked for each of the three induced scalar classes.

\begin{lemma}[Two-sided empirical Rademacher uniform deviation]
\label{lem:two-sided-rad}
Let $\mathcal{Z}$ be a measurable space, let $P$ be a probability distribution on $\mathcal{Z}$, let $n \ge 1$ be an integer, and let $B \in [0, \infty)$.
Let $\mathcal{F}$ be a nonempty pointwise-measurable class of functions $f : \mathcal{Z} \to [0, B]$, and let $Z_{1:n} = (Z_1, \ldots, Z_n) \sim P^n$ be an i.i.d.\ sample with $\E[f] := \E_{Z \sim P}[f(Z)]$ and $\widehat{\E}_{Z_{1:n}}[f] := n^{-1}\sum_{i=1}^{n} f(Z_i)$.
Then for every $\delta' \in (0, 1/2)$,
\begin{equation*}
  \Prob_{Z_{1:n} \sim P^n}\!\left(
    \sup_{f \in \mathcal{F}}\bigl|\E[f] - \widehat{\E}_{Z_{1:n}}[f]\bigr|
    \;>\;
    2\,\widehat{\mathfrak{R}}_{Z_{1:n}}(\mathcal{F}) + 3B\sqrt{\frac{\log(2/\delta')}{2n}}
  \right) \;\le\; 2\delta'.
\end{equation*}
\end{lemma}

\begin{proof}
If $B = 0$, then every $f \in \mathcal{F}$ vanishes identically on $\mathcal{Z}$, so $\E[f] = \widehat{\E}_{Z_{1:n}}[f] = 0$ for all $f \in \mathcal{F}$ and $\widehat{\mathfrak{R}}_{Z_{1:n}}(\mathcal{F}) = 0$, whence the probability of exceeding $0$ is $0 \le 2\delta'$.
Assume henceforth that $B > 0$, and set $t := B\sqrt{\log(2/\delta')/(2n)}$ and $\mathfrak{R}_n(\mathcal{F}) := \E_{Z_{1:n} \sim P^n}\bigl[\widehat{\mathfrak{R}}_{Z_{1:n}}(\mathcal{F})\bigr]$.

Define the upper one-sided supremum $\Phi_+(Z_{1:n}) := \sup_{f \in \mathcal{F}}\bigl(\E[f] - \widehat{\E}_{Z_{1:n}}[f]\bigr)$, which is measurable along with $\widehat{\mathfrak{R}}_{Z_{1:n}}(\mathcal{F})$ by pointwise measurability of $\mathcal{F}$~\citep{bartlett2002rademacher, mohri2018foundations}.
For any two tuples $z_{1:n}, z'_{1:n} \in \mathcal{Z}^n$ differing only at coordinate $i \in [n]$, range containment $f(z_i), f(z'_i) \in [0, B]$ implies $|f(z_i) - f(z'_i)| \le B$, so replacing $z_i$ by $z'_i$ changes $\Phi_+(z_{1:n})$ by at most $\sup_{f \in \mathcal{F}} n^{-1}|f(z'_i) - f(z_i)| \le B/n$ and changes $\widehat{\mathfrak{R}}_{z_{1:n}}(\mathcal{F})$ by at most $\E_\sigma\bigl[\sup_{f \in \mathcal{F}} n^{-1}|\sigma_i(f(z_i) - f(z'_i))|\bigr] \le B/n$.
Because $\sum_{i=1}^{n} (B/n)^2 = B^2/n$ and $\exp(-2n t^2 / B^2) = \delta'/2$, McDiarmid's bounded-differences inequality yields $\Prob\bigl(\Phi_+(Z_{1:n}) - \E[\Phi_+(Z_{1:n})] > t\bigr) \le \delta'/2$ and $\Prob\bigl(\mathfrak{R}_n(\mathcal{F}) - \widehat{\mathfrak{R}}_{Z_{1:n}}(\mathcal{F}) > t\bigr) \le \delta'/2$.

Introducing an independent ghost sample $Z'_{1:n} = (Z'_1, \ldots, Z'_n) \sim P^n$ and independent uniform signs $\sigma_1, \ldots, \sigma_n \in \{-1, +1\}$, standard Rademacher symmetrization bounds $\E[\Phi_+(Z_{1:n})]$ via
\begin{align*}
  \E_{Z_{1:n}}\bigl[\Phi_+(Z_{1:n})\bigr]
  &= \E_{Z_{1:n}}\!\left[\sup_{f \in \mathcal{F}}\Bigl(\E_{Z'_{1:n}}\bigl[\widehat{\E}_{Z'_{1:n}}[f]\bigr] - \widehat{\E}_{Z_{1:n}}[f]\Bigr)\right] \\
  &\le \E_{Z_{1:n}, Z'_{1:n}}\!\left[\sup_{f \in \mathcal{F}}\frac{1}{n}\sum_{i=1}^{n}\bigl(f(Z'_i) - f(Z_i)\bigr)\right]
    \quad\text{(Jensen's inequality)} \\
  &= \E_{Z_{1:n}, Z'_{1:n}, \sigma}\!\left[\sup_{f \in \mathcal{F}}\frac{1}{n}\sum_{i=1}^{n}\sigma_i\bigl(f(Z'_i) - f(Z_i)\bigr)\right]
    \quad\text{(exchangeability)} \\
  &\le 2\,\E_{Z_{1:n}, \sigma}\!\left[\sup_{f \in \mathcal{F}}\frac{1}{n}\sum_{i=1}^{n}\sigma_i f(Z_i)\right] \;=\; 2\,\mathfrak{R}_n(\mathcal{F}),
    \quad\text{(subadditivity of $\sup$)}
\end{align*}
where the final inequality splits the supremum and uses $Z'_{1:n} \stackrel{d}{=} Z_{1:n}$ and $-\sigma \stackrel{d}{=} \sigma$.
On the intersection of the two McDiarmid events (which has failure probability at most $\delta'/2 + \delta'/2 = \delta'$ by the union bound), combining these three bounds gives
\begin{align*}
  \Phi_+(Z_{1:n})
  &\le \E_{Z_{1:n}}\bigl[\Phi_+(Z_{1:n})\bigr] + B\sqrt{\frac{\log(2/\delta')}{2n}}
    &&\text{(McDiarmid on $\Phi_+$)} \\
  &\le 2\,\mathfrak{R}_n(\mathcal{F}) + B\sqrt{\frac{\log(2/\delta')}{2n}}
    &&\text{(symmetrization)} \\
  &\le 2\,\widehat{\mathfrak{R}}_{Z_{1:n}}(\mathcal{F}) + 3B\sqrt{\frac{\log(2/\delta')}{2n}},
    &&\text{(McDiarmid on $\widehat{\mathfrak{R}}_{Z_{1:n}}$)}
\end{align*}
where the third step substitutes $2\,\mathfrak{R}_n(\mathcal{F}) \le 2\,\widehat{\mathfrak{R}}_{Z_{1:n}}(\mathcal{F}) + 2t$.

For the lower tail $\Phi_-(Z_{1:n}) := \sup_{f \in \mathcal{F}}\bigl(\widehat{\E}_{Z_{1:n}}[f] - \E[f]\bigr) = \sup_{h \in -\mathcal{F}}\bigl(\E[h] - \widehat{\E}_{Z_{1:n}}[h]\bigr)$, every $h = -f \in -\mathcal{F}$ takes values in $[-B, 0]$ of diameter $B$ (equivalently, $B - f \in [0, B]$ has $\E[B - f] - \widehat{\E}_{Z_{1:n}}[B - f] = \widehat{\E}_{Z_{1:n}}[f] - \E[f]$), so the $B/n$ bounded-differences property and symmetrization apply identically, while sign symmetry $-\sigma \stackrel{d}{=} \sigma$ gives $\widehat{\mathfrak{R}}_{Z_{1:n}}(-\mathcal{F}) = \widehat{\mathfrak{R}}_{Z_{1:n}}(\mathcal{F})$.
Hence $\Prob\bigl(\Phi_-(Z_{1:n}) > 2\,\widehat{\mathfrak{R}}_{Z_{1:n}}(\mathcal{F}) + 3t\bigr) \le \delta'$ as well.
Since $\sup_{f \in \mathcal{F}}|\E[f] - \widehat{\E}_{Z_{1:n}}[f]| = \max\{\Phi_+(Z_{1:n}), \Phi_-(Z_{1:n})\}$, a union bound over the two tails yields total failure probability at most $\delta' + \delta' = 2\delta'$.
\end{proof}

The next theorem proves \Cref{thm:score-rademacher}, upper-bounding the surrogate excess risk $\Loss(g, W) - \LossStar$ (and, when $\ell$ is calibrated, the $0$--$1$ excess risk $\Risk(f_{g, W}) - \RiskStar$) simultaneously over $(g, W) \in Q$ via the empirical plug-in certificate $\widehat{\tauscore}(g, W)$ and the concentration slack $\Gamma_Q$.

\begin{theorem}[Uniform empirical certificate over a fixed candidate set; restatement of \Cref{thm:score-rademacher}]
\label{thm:rad-score-output-certificate}
Under the assumptions of \Cref{thm:T2P-app} (\Cref{thm:score-certificate}), let $B_\lambda, B_\zeta$ be defined as in \Cref{cor:C2prime-app} (\Cref{cor:score-finite-library}), and let $Q \subseteq \Gfamily \times \Wclass$ be a nonempty set of measurable candidate pairs whose induced scalar classes $\mathcal{L}_Q$, $\mathcal{Z}_{{\rm T}, Q}$, and $\mathcal{Z}_{{\rm S}, Q}$ satisfy the measurability conditions of \Cref{app:rad-assumptions}.
For all sample sizes $\nT, n_{\rm S} \ge 1$ and every $\delta \in (0, 1)$, if $D_{\rm T} = \{(X_i, Y_i)\}_{i=1}^{\nT} \sim \PT^{\nT}$ (with input tuple $X_{1:\nT}$) and $D_{\rm S} = \{U_j\}_{j=1}^{n_{\rm S}} \sim P_{{\rm S}, X}^{n_{\rm S}}$ are mutually independent i.i.d.\ samples drawn independently of $Q$, then with probability at least $1 - \delta$ over $(D_{\rm T}, D_{\rm S})$, simultaneously for all $(g, W) \in Q$,
\begin{equation}
\label{eq:rad-surrogate}
  \Loss(g, W) - \LossStar
  \;\le\;
  \widehat{\tauscore}(g, W) + \Gamma_Q,
\end{equation}
where the uniform concentration slack $\Gamma_Q$ is
\begin{align*}
  \Gamma_Q
  \;:=\;& 2 \widehat{\mathfrak{R}}_{D_{\rm T}}(\mathcal{L}_Q)
    + 3 B_\lambda \sqrt{\frac{\log(12 / \delta)}{2 \nT}} \\
  &+ 2 \widehat{\mathfrak{R}}_{X_{1:\nT}}(\mathcal{Z}_{{\rm T}, Q})
    + 3 B_\zeta \sqrt{\frac{\log(12 / \delta)}{2 \nT}} \\
  &+ 4 \widehat{\mathfrak{R}}_{D_{\rm S}}(\mathcal{Z}_{{\rm S}, Q})
    + 6 B_\zeta \sqrt{\frac{\log(12 / \delta)}{2 n_{\rm S}}}.
\end{align*}
Consequently, whenever $\ell$ is multiclass classification-calibrated with nondecreasing regret transform $\psi_\ell^{-1}: [0, \infty) \to [0, \infty)$, on the same $(1 - \delta)$-probability event, simultaneously for all $(g, W) \in Q$,
\begin{equation}
\label{eq:rad-calibrated}
  \Risk(f_{g, W}) - \RiskStar
  \;\le\;
  \psi_\ell^{-1}\!\bigl(\widehat{\tauscore}(g, W) + \Gamma_Q\bigr).
\end{equation}
\end{theorem}

\begin{proof}
Condition on $Q$ whenever $Q$ is constructed on an independent holdout split, so that $Q$ is fixed and independent of $(D_{\rm T}, D_{\rm S})$.
Set the three component slacks
\begin{align*}
  \Delta_\lambda &:= 2\widehat{\mathfrak{R}}_{D_{\rm T}}(\mathcal{L}_Q) + 3B_\lambda\sqrt{\frac{\log(12/\delta)}{2\nT}}, \\
  \Delta_{\rm T} &:= 2\widehat{\mathfrak{R}}_{X_{1:\nT}}(\mathcal{Z}_{{\rm T}, Q}) + 3B_\zeta\sqrt{\frac{\log(12/\delta)}{2\nT}}, \\
  \Delta_{\rm S} &:= 2\widehat{\mathfrak{R}}_{D_{\rm S}}(\mathcal{Z}_{{\rm S}, Q}) + 3B_\zeta\sqrt{\frac{\log(12/\delta)}{2n_{\rm S}}},
\end{align*}
so that $\Gamma_Q = \Delta_\lambda + \Delta_{\rm T} + 2\Delta_{\rm S}$.

By \Cref{lem:score-envelopes}, every function in $\mathcal{L}_Q$ is measurable and takes values in $[0, B_\lambda]$, while every function in $\mathcal{Z}_{{\rm T}, Q}$ and $\mathcal{Z}_{{\rm S}, Q}$ is measurable and takes values in $[0, B_\zeta]$.
Since $\delta \in (0, 1)$, setting $\delta' := \delta/6 \in (0, 1/6) \subset (0, 1/2)$ gives $2/\delta' = 12/\delta$ and $2\delta' = \delta/3$.
Applying \Cref{lem:two-sided-rad} at $\delta' = \delta/6$ separately to $\mathcal{L}_Q$ on $D_{\rm T} \sim \PT^{\nT}$ (envelope $B_\lambda$), to $\mathcal{Z}_{{\rm T}, Q}$ on $X_{1:\nT} \sim P_{{\rm T}, X}^{\nT}$ (envelope $B_\zeta$), and to $\mathcal{Z}_{{\rm S}, Q}$ on $D_{\rm S} \sim P_{{\rm S}, X}^{n_{\rm S}}$ (envelope $B_\zeta$) yields three bad events of probability at most $\delta/3$ each.
By the union bound, their complements hold simultaneously with probability at least $1 - 3(\delta/3) = 1 - \delta$, on which
\begin{align*}
  \sup_{(g, W) \in Q} \bigl|\lamscore(g, W) - \widehat{\lamscore}(g, W)\bigr| &\le \Delta_\lambda, \\
  \sup_{(g, W) \in Q} \bigl|T(g, W) - \widehat T(g, W)\bigr| &\le \Delta_{\rm T}, \qquad T(g, W) := \E_{X \sim P_{{\rm T}, X}}\bigl[\zeta_W(g(X))\bigr], \\
  \sup_{W \in \Wclass_Q} \bigl|S(W) - \widehat S(W)\bigr| &\le \Delta_{\rm S}, \qquad S(W) := \E_{U \sim P_{{\rm S}, X}}\bigl[\zeta_W(U)\bigr].
\end{align*}

On this $(1 - \delta)$-probability event, combining \Cref{thm:T2P-app} with the deterministic comparison \eqref{eq:C2prime-det-slack} simultaneously for every $(g, W) \in Q$ (noting $W \in \Wclass_Q$) yields
\begin{align*}
  \Loss(g, W) - \LossStar
  &\le \tauscore(g, W)
    \quad\text{(by \Cref{thm:T2P-app})} \\
  &\le \widehat{\tauscore}(g, W) + \bigl|\lamscore(g, W) - \widehat{\lamscore}(g, W)\bigr| \\
  &\quad + \bigl|T(g, W) - \widehat T(g, W)\bigr| + 2\bigl|S(W) - \widehat S(W)\bigr|
    \quad\text{(by \eqref{eq:C2prime-det-slack})} \\
  &\le \widehat{\tauscore}(g, W) + \Delta_\lambda + \Delta_{\rm T} + 2\Delta_{\rm S}
  \;=\; \widehat{\tauscore}(g, W) + \Gamma_Q,
\end{align*}
establishing \eqref{eq:rad-surrogate}.
When $\ell$ is multiclass classification-calibrated, applying the calibrated excess-risk bound of \Cref{thm:T2P-app} together with monotonicity of $\psi_\ell^{-1}: [0, \infty) \to [0, \infty)$ to \eqref{eq:rad-surrogate} on the same event yields \eqref{eq:rad-calibrated} simultaneously for all $(g, W) \in Q$.
\end{proof}

\paragraph{Vector-contraction reductions.}
When the associated vector-valued classes are measurable on the sample, the scalar empirical Rademacher complexities in \Cref{thm:rad-score-output-certificate} admit further upper bounds via the vector-contraction inequality of \citet{maurer2016vector}.
For a vector class $\mathcal{V} \subseteq (\mathbb{R}^K)^{\mathcal{Z}}$ on $Z_{1:n}$, write
\begin{equation*}
  \widehat{\mathfrak{V}}_{Z_{1:n}}(\mathcal{V}) \;:=\; \frac{1}{n}\,\E_\sigma\!\left[\sup_{v \in \mathcal{V}}\sum_{i=1}^{n}\sum_{k=1}^{K} \sigma_{i, k} v_k(Z_i) \;\middle|\; Z_{1:n}\right]
\end{equation*}
with independent uniform Rademacher signs $\sigma_{i, k} \in \{-1, +1\}$, under which sample-wise composition $\varphi \circ \mathcal{V} := \{z_i \mapsto \varphi_i(v(z_i)) : v \in \mathcal{V}\}$ for $L$-Lipschitz scalar maps $\varphi_i : \mathbb{R}^K \to \mathbb{R}$ (with respect to $\|\cdot\|_2$) satisfies $\widehat{\mathfrak{R}}_{Z_{1:n}}(\varphi \circ \mathcal{V}) \le \sqrt{2}\,L\,\widehat{\mathfrak{V}}_{Z_{1:n}}(\mathcal{V})$~\citep[Corollary~4]{maurer2016vector}.
First, by \Cref{lem:score-envelopes}, every map in $\mathcal{C}_Q := \{x \mapsto W c(g(x)) : (g, W) \in Q\}$ takes values in $B(0, R_{U1})$, where $q \mapsto \ell(Y_i, q)$ is $L_\ell$-Lipschitz with respect to $\|\cdot\|_2$; contracting $\mathcal{L}_Q$ conditioned on $Y_{1:\nT}$ therefore gives
\begin{equation*}
  \widehat{\mathfrak{R}}_{D_{\rm T}}(\mathcal{L}_Q) \;\le\; \sqrt{2}\,L_\ell\,\widehat{\mathfrak{V}}_{X_{1:\nT}}(\mathcal{C}_Q).
\end{equation*}
Second, consider $\Psi(a, b) := \max_{y \in [K_{\rm T}]} |\ell(y, a) - \ell(y, b)|$ on $B(0, R_{U1}) \times B(0, R_{U1}) \subseteq \mathbb{R}^{2K_{\rm T}}$ equipped with the Euclidean product norm $\|(a, b)\|_2 = (\|a\|_2^2 + \|b\|_2^2)^{1/2}$.
For any $(a, b), (a', b') \in B(0, R_{U1})^2$, expanding $|\Psi(a, b) - \Psi(a', b')|$ one step per line yields
\begin{align*}
  \bigl|\Psi(a, b) - \Psi(a', b')\bigr|
  &= \Bigl|\max_{y \in [K_{\rm T}]}\bigl|\ell(y, a) - \ell(y, b)\bigr|
         - \max_{y \in [K_{\rm T}]}\bigl|\ell(y, a') - \ell(y, b')\bigr|\Bigr| \\
  &\le \max_{y \in [K_{\rm T}]}\Bigl|\bigl|\ell(y, a) - \ell(y, b)\bigr|
         - \bigl|\ell(y, a') - \ell(y, b')\bigr|\Bigr| \\
  &\le \max_{y \in [K_{\rm T}]}\bigl|\bigl(\ell(y, a) - \ell(y, a')\bigr)
         - \bigl(\ell(y, b) - \ell(y, b')\bigr)\bigr| \\
  &\le \max_{y \in [K_{\rm T}]}\Bigl(\bigl|\ell(y, a) - \ell(y, a')\bigr|
         + \bigl|\ell(y, b) - \ell(y, b')\bigr|\Bigr) \\
  &\le L_\ell\bigl(\|a - a'\|_2 + \|b - b'\|_2\bigr)
    \quad\text{($L_\ell$-Lipschitz on $B(0, R_{U1})$)} \\
  &\le \sqrt{2}\,L_\ell\sqrt{\|a - a'\|_2^2 + \|b - b'\|_2^2}
    \quad\text{(Cauchy--Schwarz)} \\
  &= \sqrt{2}\,L_\ell\,\bigl\|(a, b) - (a', b')\bigr\|_2,
\end{align*}
where the first three inequalities apply $|\max_y u_y - \max_y v_y| \le \max_y |u_y - v_y|$, the reverse triangle inequality $||r| - |s|| \le |r - s|$, and the triangle inequality, so $\Psi$ is $\sqrt{2}\,L_\ell$-Lipschitz on $B(0, R_{U1})^2$.
Because $\mathcal{D}_{{\rm T}, Q} := \{x \mapsto (W s_{\rm S}(g(x)),\, W c(g(x))) : (g, W) \in Q\}$ and $\mathcal{D}_{{\rm S}, Q} := \{u \mapsto (W s_{\rm S}(u),\, W c(u)) : W \in \Wclass_Q\}$ take values in $B(0, R_{U1})^2$ by \Cref{lem:score-envelopes}, applying \citet[Corollary~4]{maurer2016vector} with factor $\sqrt{2} \cdot (\sqrt{2}\,L_\ell) = 2L_\ell$ gives
\begin{equation*}
  \widehat{\mathfrak{R}}_{X_{1:\nT}}(\mathcal{Z}_{{\rm T}, Q}) \;\le\; 2L_\ell\,\widehat{\mathfrak{V}}_{X_{1:\nT}}(\mathcal{D}_{{\rm T}, Q}),
  \qquad
  \widehat{\mathfrak{R}}_{D_{\rm S}}(\mathcal{Z}_{{\rm S}, Q}) \;\le\; 2L_\ell\,\widehat{\mathfrak{V}}_{D_{\rm S}}(\mathcal{D}_{{\rm S}, Q}),
\end{equation*}
providing optional vector-complexity refinements of \Cref{thm:rad-score-output-certificate}.

%% file: appendix/worked_examples.tex
\label{app:examples}

\textbf{Section overview.}
This section provides the complete combinatorial derivation behind \Cref{ex:gap} in \Cref{sec:T1:conditional-growth}.
Computing the conditional-relabeling growth $\PihatH(m)$ and the structure-oblivious proxy $\PihatHall(m)$ in closed form for the unrestricted output-mapping class $\Hfamily = [K_{\rm T}]^{[K_{\rm S}]}$ shows that exact alphabet-exponent identities hold for all $K_{\rm T} \ge 1$, while for $K_{\rm T} \ge 2$ conditioning on reachable source-label patterns eliminates an artificial $O(\nT \log K_{\rm T})$ log-growth penalty in \Cref{thm:T1} (\Cref{thm:T1-app}) whenever $K_{\rm S}$ is large relative to the reached source alphabet.

\begin{example}[Strict improvement on small reachable alphabets; derivation of \Cref{ex:gap}]
\label{ex:reachable-alphabet-derivation}
Under the wrapper setup of \Cref{sec:prelim:wrapper} (fixed measurable $\fs: \XS \to [K_{\rm S}]$, nonempty $\XT$, and nonempty $\Gfamily$), let $K_{\rm T} \ge 1$ and $K_{\rm S} \ge 1$, and consider the unrestricted output-mapping class
\begin{equation}
\label{eq:full-head-class}
  \Hfamily \;=\; [K_{\rm T}]^{[K_{\rm S}]}.
\end{equation}
For every integer $m \ge 1$ and every source-label vector $z = (z_1, \ldots, z_m) \in [K_{\rm S}]^m$, recall from \Cref{ex:gap} and \Cref{app:assumptions} that $\reachset(z) = \{z_1, \ldots, z_m\} \subseteq [K_{\rm S}]$ denotes the set of distinct source labels appearing in $z$, with cardinality $|z|_{\#} = |\reachset(z)| \in \{1, \ldots, \min\{m, K_{\rm S}\}\}$.
Two maps $h, h' \in \Hfamily$ induce the same target-label vector $(h(z_1), \ldots, h(z_m)) = (h'(z_1), \ldots, h'(z_m))$ if and only if their restrictions $h|_{\reachset(z)}$ and $h'|_{\reachset(z)}$ coincide, and under \eqref{eq:full-head-class} every map in $[K_{\rm T}]^{\reachset(z)}$ extends to $\Hfamily$ (for instance by assigning label $1 \in [K_{\rm T}]$ outside $\reachset(z)$).
The restriction map $h \mapsto h|_{\reachset(z)}$ therefore induces a bijection between $\RelClass(z)$ and $[K_{\rm T}]^{\reachset(z)}$, giving the exact pattern count for every $K_{\rm T} \ge 1$:
\begin{equation}
\label{eq:count-full-head}
  |\RelClass(z)| \;=\; \bigl|[K_{\rm T}]^{\reachset(z)}\bigr| \;=\; K_{\rm T}^{|\reachset(z)|} \;=\; K_{\rm T}^{|z|_{\#}}.
\end{equation}

To evaluate both growth functions from \eqref{eq:count-full-head}, note that $t \mapsto K_{\rm T}^t$ is nondecreasing on $\mathbb{R}$ for every $K_{\rm T} \ge 1$.
For the structure-oblivious proxy, every $z \in [K_{\rm S}]^m$ satisfies $|z|_{\#} \le \min\{m, K_{\rm S}\}$, with equality attained at $z^\star := (1, 2, \ldots, q, 1, \ldots, 1) \in [K_{\rm S}]^m$ where $q := \min\{m, K_{\rm S}\}$, whence
\begin{align*}
  \PihatHall(m)
  &= \sup_{z \in [K_{\rm S}]^m} |\RelClass(z)|
    &&\text{(by definition)} \\
  &= \sup_{z \in [K_{\rm S}]^m} K_{\rm T}^{|z|_{\#}}
    &&\text{(by \eqref{eq:count-full-head})} \\
  &= K_{\rm T}^{\max_{z \in [K_{\rm S}]^m} |z|_{\#}}
    &&\text{($t \mapsto K_{\rm T}^t$ nondecreasing)} \\
  &= K_{\rm T}^{\min\{m, K_{\rm S}\}}.
    &&\text{(attained at $z^\star$)}
\end{align*}
For the conditional-relabeling growth, because $\XT$ and $\Gfamily$ are nonempty, the integer $|z|_{\#} \in \{1, \ldots, \min\{m, K_{\rm S}\}\}$ attains its maximum $\tR(m) = \sup_{x_{1:m} \in \XT^m}\sup_{z \in \ReaClass(x_{1:m})} |z|_{\#}$ over reachable patterns, giving
\begin{align*}
  \PihatH(m)
  &= \sup_{x_{1:m} \in \XT^m} \sup_{z \in \ReaClass(x_{1:m})} |\RelClass(z)|
    &&\text{(by \Cref{def:growth})} \\
  &= \sup_{x_{1:m} \in \XT^m} \sup_{z \in \ReaClass(x_{1:m})} K_{\rm T}^{|z|_{\#}}
    &&\text{(by \eqref{eq:count-full-head})} \\
  &= K_{\rm T}^{\tR(m)}.
    &&\text{($t \mapsto K_{\rm T}^t$ nondecreasing)}
\end{align*}
Consequently, for every $K_{\rm T} \ge 1$ and every integer $m \ge 1$, matching \Cref{ex:gap},
\begin{equation}
\label{eq:illustration-growths}
  \PihatHall(m) \;=\; K_{\rm T}^{\min\{m, K_{\rm S}\}},
  \qquad
  \PihatH(m) \;=\; K_{\rm T}^{\tR(m)}.
\end{equation}

When $K_{\rm T} = 1$, $\log K_{\rm T} = 0$ and both $\log\PihatHall(2\nT)$ and $\log\PihatH(2\nT)$ vanish identically.
When $K_{\rm T} \ge 2$, we have $\log K_{\rm T} > 0$, and evaluating \eqref{eq:illustration-growths} at $m = 2\nT$ in the large-source-alphabet regime $K_{\rm S} \ge 2\nT$ gives
\begin{align*}
  \log\PihatHall(2\nT)
  &= \min\{2\nT, K_{\rm S}\}\log K_{\rm T}
  \;=\; 2\nT\log K_{\rm T}, \\
  \log\PihatH(2\nT)
  &= \tR(2\nT)\log K_{\rm T},
\end{align*}
which yields the strict separation $\log\PihatH(2\nT) < \log\PihatHall(2\nT)$ whenever $\tR(2\nT) < 2\nT$.
In particular, when $K_{\rm T} \ge 2$, $K_{\rm S} \ge 2\nT$, and $\tR(2\nT) = O(1)$, substituting these logarithmic counts into the statistical term of \Cref{thm:T1} (\Cref{thm:T1-app}) shows that the structure-oblivious proxy contributes $\sqrt{\log\PihatHall(2\nT)/\nT} = \sqrt{2\log K_{\rm T}}$, yielding a vacuous $\Omega(1)$ penalty, whereas the conditional-relabeling contribution $\sqrt{\log\PihatH(2\nT)/\nT} = \sqrt{\tR(2\nT)(\log K_{\rm T})/\nT} = O\bigl(\sqrt{(\log K_{\rm T})/\nT}\bigr)$ vanishes at rate $\nT^{-1/2}$.
\end{example}

%% file: appendix/empirical_verification.tex
\label{app:empirical}

This section provides the complete experimental protocols, formal definitions of the hypothesis-class interventions, sample-split and evaluation governance, full numerical tables across all sample sizes and checkpoints, control comparisons, and compute disclosures for the two controlled empirical studies summarized in \Cref{sec:empirical}. Code, frozen pretrained checkpoints, and reproduction scripts are publicly available at \url{https://github.com/zyecs/repgm-slt}.

\subsection{Study 1: Synthetic Exact-ERM Study with Analytical Population Risks}
\label{app:empirical:synthetic}

\paragraph{Objective and design rationale.}
The synthetic study isolates the two core structural mechanisms of our theory under conditions free of optimization slack ($\eta_n = 0$) and Monte Carlo evaluation noise:
\begin{enumerate}
  \item \textbf{Estimation invariance and capacity penalty (\Cref{thm:T1}):} Since \Cref{thm:T1} bounds finite-sample estimation error through the sum $\log \PihatR(2\nT) + \log \PihatH(2\nT)$, shifting the exact same hypothesis-class capacity between input-side reachability and output-side relabeling while preserving the induced target predictor class $\FMR$ should leave exact ERM behavior invariant. Simultaneously, enlarging active capacity should increase small-sample estimation error and vanish as $\nT$ grows.
  \item \textbf{Approximation mechanism attribution (\Cref{prop:T2P}):} When two wrapper designs share the exact same suboptimal in-class oracle risk $\RiskClsOptimum$, total risk alone cannot identify why reprogramming fails, whereas \Cref{prop:T2P} separates source-label information loss from output-mapping restriction.
\end{enumerate}

\paragraph{Data-generating distribution and frozen source classifier.}
The target input space $\XT = \{e_1, \ldots, e_{16}\} \subset \mathbb{R}^{16}$ consists of the $16$ standard basis vectors, grouped into $8$ ordered pairs $P_j = (e_{2j-1}, e_{2j})$ for $j \in [8]$. The target marginal $P_{{\rm T},X}$ is uniform over $\XT$ ($P_{{\rm T},X}(X = e_i) = 1/16$). For each pair $P_j$, the noiseless Bayes label is $0$ on the first element $e_{2j-1}$ and $1$ on the second element $e_{2j}$. The observed target label $Y \in \{0, 1\}$ is corrupted by independent symmetric flip noise of probability $0.1$, yielding a target Bayes risk of $\RiskStar = 0.1$ exactly. The baseline frozen source classifier $\fs : \mathbb{R}^{16} \to [16]$ assigns each basis vector $e_i$ its unique coordinate index $\fs(e_i) = i$ with multiclass score margin $1$, preserving full input distinguishability prior to any controlled bottleneck.

\paragraph{Part A protocol: Input-side reachability vs.\ output-side relabeling.}
For each capacity parameter $r \in \{0, 2, 4, 6, 8\}$ and sample size $\nT \in \{8, 16, 32, 64, 128, 256, 512\}$, we construct two matched arms of cardinality $2^r$:
\begin{itemize}
  \item \textbf{Input-side reachability arm ($\mathrm{A\text{-}R}$):} The output mapping is fixed to the canonical parity head $h_{\mathrm{can}}(k) = \one\{k \text{ is even}\}$. The input transformation class $\Gfamily_r$ consists of $2^r$ coordinate permutations that independently either preserve or swap the two basis vectors within each of the first $r$ pairs $P_1, \ldots, P_r$ while leaving pairs $P_{r+1}, \ldots, P_8$ fixed. Here $|\ReaClass(x_{1:m})| = 2^{q(x_{1:m})}$ where $q(x_{1:m}) \le r$ is the number of active pairs observed in $x_{1:m}$, while $|\RelClass(z)| = 1$.
  \item \textbf{Output-side relabeling arm ($\mathrm{A\text{-}H}$):} The input transformation is fixed to the identity $g_{\mathrm{id}}(x) = x$. The output mapping class $\Hfamily_r$ consists of $2^r$ relabelings that independently either preserve $h_{\mathrm{can}}$ or flip the binary target assignments on both source labels $\{2j-1, 2j\}$ of each of the first $r$ pairs. Here $|\ReaClass(x_{1:m})| = 1$, while $|\RelClass(z)| = 2^{q(x_{1:m})}$.
\end{itemize}
Both arms induce the exact same composite predictor class $\FMR$ of size $2^r$ and contain the target Bayes rule ($\RiskClsOptimum = \RiskStar = 0.1$, so $a_{\mathrm{MR}} = 0$). For each $(r, \nT)$ cell, we run $200$ paired repetitions from root seed $\texttt{20260725}$. In each repetition, exhaustive finite-class ERM selects $\widehat f_n \in \FMR$ using a shared label-independent canonical tie-breaking order. We evaluate the exact analytical population estimation excess $\EstRisk(\widehat f_n) = \Risk(\widehat f_n) - 0.1$, alongside descriptive checks on a $20{,}000$-example held-out sample and $10{,}000$ paired whole-repetition bootstrap draws.

\begin{table}[h]
\centering
\caption{Part A complete results across all capacity levels $r$ and target sample sizes $\nT$ ($200$ paired repetitions). Shown are mean exact population estimation excess $\EstRisk(\widehat f_n)$ with descriptive $95\%$ paired-bootstrap intervals, and mean realized $\log_2$ active growth $\E[q(X_{1:\nT})]$. In every $(r, \nT)$ cell, the paired difference between $\mathrm{A\text{-}R}$ and $\mathrm{A\text{-}H}$ is identically $0.0000$ with bootstrap interval $[0.0000, 0.0000]$.}
\label{tab:app-synth-slice-a}
\resizebox{\textwidth}{!}{%
\begin{tabular}{@{}lcccccc@{}}
\toprule
Capacity $r$ & $\nT = 8$ & $\nT = 16$ & $\nT = 32$ & $\nT = 64$ & $\nT = 128$ & $\nT \in \{256, 512\}$ \\
\midrule
$r = 0$ ($\EstRisk$; $\E[q]=0$) & $0.0000$ $[0.000, 0.000]$ & $0.0000$ $[0.000, 0.000]$ & $0.0000$ $[0.000, 0.000]$ & $0.0000$ $[0.000, 0.000]$ & $0.0000$ $[0.000, 0.000]$ & $0.0000$ $[0.000, 0.000]$ \\
$r = 2$ ($\EstRisk$) & $0.0475$ $[0.040, 0.056]$ & $0.0220$ $[0.017, 0.028]$ & $0.0080$ $[0.004, 0.012]$ & $0.0020$ $[0.001, 0.004]$ & $0.0000$ $[0.000, 0.000]$ & $0.0000$ $[0.000, 0.000]$ \\
$r = 4$ ($\EstRisk$) & $0.0995$ $[0.088, 0.111]$ & $0.0455$ $[0.037, 0.053]$ & $0.0155$ $[0.011, 0.021]$ & $0.0035$ $[0.001, 0.006]$ & $0.0000$ $[0.000, 0.000]$ & $0.0000$ $[0.000, 0.000]$ \\
$r = 6$ ($\EstRisk$) & $0.1340$ $[0.120, 0.147]$ & $0.0705$ $[0.060, 0.081]$ & $0.0255$ $[0.019, 0.033]$ & $0.0045$ $[0.002, 0.007]$ & $0.0000$ $[0.000, 0.000]$ & $0.0000$ $[0.000, 0.000]$ \\
$r = 8$ ($\EstRisk$) & $0.1765$ $[0.162, 0.192]$ & $0.0990$ $[0.087, 0.111]$ & $0.0295$ $[0.022, 0.037]$ & $0.0050$ $[0.002, 0.008]$ & $0.0005$ $[0.000, 0.002]$ & $0.0000$ $[0.000, 0.000]$ \\
\midrule
$\E[q]$ ($r=2 / 4 / 6 / 8$) & $1.32 / 2.57 / 3.89 / 5.17$ & $1.78 / 3.58 / 5.34 / 7.10$ & $1.97 / 3.95 / 5.92 / 7.90$ & $2.00 / 4.00 / 6.00 / 8.00$ & $2.00 / 4.00 / 6.00 / 8.00$ & $2.00 / 4.00 / 6.00 / 8.00$ \\
\bottomrule
\end{tabular}%
}
\end{table}

\paragraph{Part B protocol: Source-label information loss vs.\ output-mapping restriction.}
For each severity level $m \in \{0, 2, 4, 6, 8\}$ (number of affected pairs) and sample size $\nT \in \{32, 128, 512, 2048, 8192\}$, we compare two interventions whose in-class oracle risk is identically $\RiskClsOptimum = 0.1 + 0.05m$ (so $a_{\mathrm{MR}} = 0.05m$):
\begin{itemize}
  \item \textbf{Information-loss intervention ($\mathrm{B\text{-}I}$):} The source channel merges both inputs of each affected pair $P_j$ ($1 \le j \le m$) onto a single shared source label, while the output mapping class $\Hfamily = \{0, 1\}^{[16]}$ is unrestricted. Because each merged cell pools equal probability mass from Bayes-label $0$ and Bayes-label $1$ points, no relabeling can do better than random guessing on the first $m$ pairs. Analytically, the source-label information loss is $I := R_{\mathrm{all}}^{\ast}(g) - \RiskStar = 0.05m$, and the output-mapping restriction is $H := R_{\Hfamily}^{\ast}(g) - R_{\mathrm{all}}^{\ast}(g) = 0$.
  \item \textbf{Head-restriction intervention ($\mathrm{B\text{-}H}$):} The input transformation is the identity and all $16$ source labels remain distinct ($R_{\mathrm{all}}^{\ast}(g) = \RiskStar = 0.1$, so $I = 0$), but every $h \in \Hfamily$ is constrained to output the incorrect target label $1$ on the first member $e_{2j-1}$ (whose Bayes label is $0$) of each affected pair $P_j$ ($1 \le j \le m$) while leaving the remaining $16 - m$ source labels free in $\{0, 1\}$, forcing one Bayes disagreement per affected pair ($|\Hfamily| = 2^{16-m}$, matching $|\FMR| = 2^{16-m}$ of $\mathrm{B\text{-}I}$) and yielding $H = 0.05m$.
\end{itemize}
In addition to exact population evaluation, we compute empirical plug-in estimates $(\widehat I, \widehat H)$ on an independent $20{,}000$-example diagnostic sample per repetition by comparing empirical risk minima over the nested reference classes $\mathcal{C}_{\mathrm{actual}} \subseteq \mathcal{C}_{\mathrm{unrestricted\text{-}head}} \subseteq \mathcal{C}_{\mathrm{unrestricted\text{-}channel}}$.

\begin{table}[h]
\centering
\caption{Part B results across severity levels $m \in \{0, 2, 4, 6, 8\}$ ($200$ repetitions; bracketed ranges denote descriptive $95\%$ bootstrap intervals for $\Risk(\widehat f_n)$, which have width $0.000$ for all $\nT \in \{512, 2048, 8192\}$). Both arms share identical in-class oracle risk $\RiskClsOptimum = 0.1 + 0.05m$, while \Cref{prop:T2P} and $(\widehat I, \widehat H)$ isolate the underlying mechanism.}
\label{tab:app-synth-slice-b}
\resizebox{\textwidth}{!}{%
\begin{tabular}{@{}llcccccc@{}}
\toprule
$m$ & Arm & Oracle $\RiskClsOptimum$ & Risk ($\nT=32$) [95\% CI] & Risk ($\nT=128$) [95\% CI] & Risk ($\nT \in \{512, 2048, 8192\}$) [95\% CI] & Exact $(I, H)$ & Plug-in $(\widehat I, \widehat H)$ \\
\midrule
$0$ & $\mathrm{B\text{-}I}$ / $\mathrm{B\text{-}H}$ & $0.1000$ & $0.1988$ $[0.190, 0.208]$ & $0.1037$ $[0.102, 0.106]$ & $0.1000$ $[0.100, 0.100]$ & $(0.0000, 0.0000)$ & $(0.0000, 0.0000)$ \\
\midrule
$2$ & $\mathrm{B\text{-}I}$ (info.\ loss) & $0.2000$ & $0.2725$ $[0.266, 0.280]$ & $0.2025$ $[0.201, 0.204]$ & $0.2000$ $[0.200, 0.200]$ & $(0.1000, 0.0000)$ & $(0.0979, 0.0000)$ \\
    & $\mathrm{B\text{-}H}$ (head restr.) & $0.2000$ & $0.2823$ $[0.275, 0.290]$ & $0.2032$ $[0.202, 0.205]$ & $0.2000$ $[0.200, 0.200]$ & $(0.0000, 0.1000)$ & $(0.0000, 0.1000)$ \\
\midrule
$4$ & $\mathrm{B\text{-}I}$ (info.\ loss) & $0.3000$ & $0.3520$ $[0.346, 0.358]$ & $0.3018$ $[0.301, 0.303]$ & $0.3000$ $[0.300, 0.300]$ & $(0.2000, 0.0000)$ & $(0.1958, 0.0000)$ \\
    & $\mathrm{B\text{-}H}$ (head restr.) & $0.3000$ & $0.3723$ $[0.365, 0.380]$ & $0.3028$ $[0.301, 0.304]$ & $0.3000$ $[0.300, 0.300]$ & $(0.0000, 0.2000)$ & $(0.0000, 0.1999)$ \\
\midrule
$6$ & $\mathrm{B\text{-}I}$ (info.\ loss) & $0.4000$ & $0.4258$ $[0.421, 0.430]$ & $0.4005$ $[0.400, 0.401]$ & $0.4000$ $[0.400, 0.400]$ & $(0.3000, 0.0000)$ & $(0.2941, 0.0000)$ \\
    & $\mathrm{B\text{-}H}$ (head restr.) & $0.4000$ & $0.4578$ $[0.451, 0.465]$ & $0.4022$ $[0.401, 0.404]$ & $0.4000$ $[0.400, 0.400]$ & $(0.0000, 0.3000)$ & $(0.0000, 0.2998)$ \\
\midrule
$8$ & $\mathrm{B\text{-}I}$ (info.\ loss) & $0.5000$ & $0.5000$ $[0.500, 0.500]$ & $0.5000$ $[0.500, 0.500]$ & $0.5000$ $[0.500, 0.500]$ & $(0.4000, 0.0000)$ & $(0.3923, 0.0000)$ \\
    & $\mathrm{B\text{-}H}$ (head restr.) & $0.5000$ & $0.5435$ $[0.538, 0.550]$ & $0.5020$ $[0.501, 0.504]$ & $0.5000$ $[0.500, 0.500]$ & $(0.0000, 0.4000)$ & $(0.0000, 0.4000)$ \\
\bottomrule
\end{tabular}%
}
\end{table}

\subsection{Study 2: Pretrained MLP and LeNet Sources on Rotated-MNIST Parity Prediction}
\label{app:empirical:mnist}

\paragraph{Source-model pretraining and hard-label interface.}
We pretrain three independent checkpoints of a two-hidden-layer MLP ($784 \to 256 \to \mathrm{ReLU} \to 128 \to \mathrm{ReLU} \to 10$) and three independent checkpoints of a minimal convolutional network (LeNet: $\mathrm{Conv}(1,6,5) \to \mathrm{ReLU} \to \mathrm{AvgPool}(2) \to \mathrm{Conv}(6,16,5) \to \mathrm{ReLU} \to \mathrm{AvgPool}(2) \to 256 \to 120 \to \mathrm{ReLU} \to 84 \to \mathrm{ReLU} \to 10$) on upright $10$-class MNIST~\citep{lecun1998gradient} in PyTorch~\citep{paszke2019pytorch}. The official $60{,}000$-image training set is split into $55{,}000$ training images and a digit-balanced $5{,}000$-image validation split (normalized with mean $0.1307$ and standard deviation $0.3081$). Each network is trained for up to $20$ epochs using Adam (learning rate $10^{-3}$, batch size $128$, cross-entropy loss) without data augmentation, selecting the checkpoint with highest validation accuracy (all six checkpoints exceed $0.976$ validation accuracy: MLP checkpoints attain $0.9790, 0.9782, 0.9768$; LeNet checkpoints attain $0.9864, 0.9886, 0.9890$). Every checkpoint is then frozen, and its output interface is restricted to the single hard source class ID $\fs(u) \in \{0, \ldots, 9\}$; logits, probabilities, internal activations, and gradients are never exposed to the downstream learner.

\paragraph{Target task, sample partitions, and wrapper/head hypothesis classes.}
The downstream pool consists of the unseen $10{,}000$-image official MNIST test set. Every target image is rotated counterclockwise by $90^\circ$ ($\mathrm{rot}_{90}(x)$), and the target label is binary digit parity ($Y = \text{digit} \bmod 2 \in \{0, 1\}$). Using deterministic stratified allocation, the $10{,}000$ target images are partitioned into a $4{,}000$-image target training pool, a $2{,}000$-image diagnostic split, and a $4{,}000$-image held-out final evaluation split. Across $20$ paired target seeds per source checkpoint, we draw nested, digit-balanced training subsets of sizes $\nT \in \{50, 200, 800, 3200\}$ from the training pool. We evaluate a $2 \times 2$ factorial design crossing two input-transformation classes and two output-mapping classes:
\begin{itemize}
  \item \textbf{Input transformations:} $\Gfamily_{\mathrm{id}} = \{\mathrm{rot}_0\}$ ($|\Gfamily_{\mathrm{id}}| = 1$, leaving the $90^\circ$ rotation uncorrected) versus $\Gfamily_{\mathrm{rot}} = \{\mathrm{rot}_0, \mathrm{rot}_{90}, \mathrm{rot}_{180}, \mathrm{rot}_{270}\}$ ($|\Gfamily_{\mathrm{rot}}| = 4$, which includes the exact inverse rotation $\mathrm{rot}_{270}$).
  \item \textbf{Output mappings:} $\Hfamily_{\mathrm{thr}}$ ($|\Hfamily_{\mathrm{thr}}| = 20$, comprising the $2$ constant maps and both orientations $\one\{k \ge t\}$, $\one\{k < t\}$ of single-threshold maps for $t \in \{1, \ldots, 9\}$, which cannot represent alternating even/odd parity) versus $\Hfamily_{\mathrm{full}} = \{0, 1\}^{\{0,\ldots,9\}}$ ($|\Hfamily_{\mathrm{full}}| = 2^{10} = 1{,}024$, containing all binary relabelings of the $10$ source digit IDs).
\end{itemize}
For every checkpoint, target seed, sample size $\nT$, and factorial cell, exact finite-class ERM enumerates all candidate wrapper-head pairs on the target training subset (breaking ties lexicographically by canonical wrapper and head indices). All $960$ predictor selections per architecture were locked on the $2{,}000$-image diagnostic split before evaluating them once, without reselection, on the $4{,}000$-image final evaluation split.

\paragraph{Estimands, controls, and complete learning curves.}
At $\nT = 3200$, we measure the input-side reachability effect $\Delta_{\mathrm{in}} := \Risk(\Gfamily_{\mathrm{id}}, \Hfamily_{\mathrm{full}}) - \Risk(\Gfamily_{\mathrm{rot}}, \Hfamily_{\mathrm{full}})$, the output-side relabeling effect $\Delta_{\mathrm{out}} := \Risk(\Gfamily_{\mathrm{rot}}, \Hfamily_{\mathrm{thr}}) - \Risk(\Gfamily_{\mathrm{rot}}, \Hfamily_{\mathrm{full}})$, their interaction $\Delta_{\mathrm{int}} := [\Risk(\Gfamily_{\mathrm{id}}, \Hfamily_{\mathrm{thr}}) - \Risk(\Gfamily_{\mathrm{id}}, \Hfamily_{\mathrm{full}})] - \Delta_{\mathrm{out}}$, and the contrast $\Delta_{\mathrm{train}} := \Risk(\text{untrained}, \Gfamily_{\mathrm{rot}}, \Hfamily_{\mathrm{full}}) - \Risk(\text{trained}, \Gfamily_{\mathrm{rot}}, \Hfamily_{\mathrm{full}})$ against one prespecified randomly initialized untrained checkpoint of the same architecture. Target-seed averages are computed within each checkpoint and equally weighted across the three checkpoints.

\begin{table}[h]
\centering
\caption{Complete final-split learning curves and $\nT = 3200$ mechanism contrasts on rotated-MNIST parity prediction (equally weighted across $3$ pretrained checkpoints $\times$ $20$ paired target seeds per architecture; bracketed ranges denote descriptive $2.5\%$--$97.5\%$ target-seed quantiles).}
\label{tab:app-mnist-full}
\footnotesize
\setlength{\tabcolsep}{4pt}
\begin{tabular}{@{}llcccc@{}}
\toprule
Model & Sample size & $(\Gfamily_{\mathrm{id}}, \Hfamily_{\mathrm{thr}})$ & $(\Gfamily_{\mathrm{id}}, \Hfamily_{\mathrm{full}})$ & $(\Gfamily_{\mathrm{rot}}, \Hfamily_{\mathrm{thr}})$ & $(\Gfamily_{\mathrm{rot}}, \Hfamily_{\mathrm{full}})$ \\
\midrule
MLP & $\nT = 50$   & $0.4701$ $[0.443, 0.520]$ & $0.4416$ $[0.407, 0.494]$ & $0.3967$ $[0.361, 0.458]$ & $0.0098$ $[0.010, 0.010]$ \\
    & $\nT = 200$  & $0.4515$ $[0.435, 0.476]$ & $0.4069$ $[0.396, 0.430]$ & $0.3718$ $[0.362, 0.391]$ & $0.0098$ $[0.010, 0.010]$ \\
    & $\nT = 800$  & $0.4439$ $[0.436, 0.472]$ & $0.4006$ $[0.395, 0.408]$ & $0.3636$ $[0.359, 0.381]$ & $0.0098$ $[0.010, 0.010]$ \\
    & $\nT = 3200$ & $0.4385$ $[0.435, 0.440]$ & $0.3980$ $[0.394, 0.400]$ & $0.3613$ $[0.359, 0.368]$ & $0.0098$ $[0.010, 0.010]$ \\
\midrule
LeNet & $\nT = 50$   & $0.4206$ $[0.401, 0.453]$ & $0.4006$ $[0.362, 0.449]$ & $0.3621$ $[0.335, 0.438]$ & $0.0058$ $[0.006, 0.006]$ \\
      & $\nT = 200$  & $0.4029$ $[0.396, 0.416]$ & $0.3634$ $[0.354, 0.382]$ & $0.3460$ $[0.335, 0.371]$ & $0.0058$ $[0.006, 0.006]$ \\
      & $\nT = 800$  & $0.3979$ $[0.396, 0.403]$ & $0.3552$ $[0.354, 0.363]$ & $0.3374$ $[0.335, 0.345]$ & $0.0058$ $[0.006, 0.006]$ \\
      & $\nT = 3200$ & $0.3960$ $[0.396, 0.396]$ & $0.3538$ $[0.354, 0.354]$ & $0.3352$ $[0.335, 0.335]$ & $0.0058$ $[0.006, 0.006]$ \\
\bottomrule
\end{tabular}

\vspace{0.4em}
\scriptsize
\setlength{\tabcolsep}{2.5pt}
\begin{tabular}{@{}lcccc@{}}
Model ($\nT = 3200$) & Input effect $\Delta_{\mathrm{in}}$ & Output effect $\Delta_{\mathrm{out}}$ & Interaction $\Delta_{\mathrm{int}}$ & Untrained contrast $\Delta_{\mathrm{train}}$ \\
\midrule
MLP (3-ckpt mean) & $0.3881$ $[0.384, 0.390]$ & $0.3514$ $[0.349, 0.358]$ & $-0.3109$ $[-0.317, -0.308]$ & $0.4257$ $[0.426, 0.426]$ \\
LeNet (3-ckpt mean) & $0.3480$ $[0.348, 0.348]$ & $0.3293$ $[0.329, 0.329]$ & $-0.2872$ $[-0.287, -0.287]$ & $0.5017$ $[0.502, 0.502]$ \\
\bottomrule
\end{tabular}
\end{table}

\paragraph{Control checks, empirical growth accounting, and compute resources.}
Across all $60$ $(\Gfamily_{\mathrm{rot}}, \Hfamily_{\mathrm{full}})$ runs per architecture and all $\nT \in \{50, 200, 800, 3200\}$, exact ERM selected the true inverse rotation $\mathrm{rot}_{270}$ in $100\%$ of runs, matching the risk of the mathematical oracle $(\mathrm{rot}_{270}, k \mapsto k \bmod 2)$ ($0.0098$ on MLP and $0.0058$ on LeNet). By contrast, majority-class prediction on the final split has risk $0.4925$, and the single untrained source control under $(\Gfamily_{\mathrm{rot}}, \Hfamily_{\mathrm{full}})$ at $\nT = 3200$ attains mean risk $0.4355$ on MLP and $0.5075$ on LeNet, confirming that low target risk requires both learned source structure and a wrapper-head pair where neither reachability nor relabeling is bottlenecked. Direct enumeration of prediction patterns on each target sample $x_{1:\nT}$ verified the sample-wise factorization bound $|\{(f(x_1),\ldots,f(x_{\nT})) : f \in \FMR\}| \le |\ReaClass(x_{1:\nT})|\max_{z \in \ReaClass(x_{1:\nT})}|\RelClass(z)|$ in all $2{,}400$ audited configurations per architecture (at $\nT = 3200$, $|\ReaClass(x_{1:\nT})|$ is $1$ under $\Gfamily_{\mathrm{id}}$ and $4$ under $\Gfamily_{\mathrm{rot}}$, yielding $20$, $1{,}024$, $74$, and $4{,}090$ composed prediction vectors across the four cells). The synthetic study required $18.46$ CPU seconds ($139.2$ MB peak RSS), and the MNIST study required $377.1$ seconds for MLP and $397.4$ seconds for LeNet on a single NVIDIA A100-80GB MIG ($1\text{g}.10\text{gb}$) slice.

%% file: checklist.tex
\section*{NeurIPS Paper Checklist}

\begin{enumerate}

\item {\bf Claims}
    \item[] Question: Do the main claims made in the abstract and introduction accurately reflect the paper's contributions and scope?
    \item[] Answer: \answerYes{}.
    \item[] Justification: The abstract and introduction state the paper's scope as a wrapper-side finite-sample learnability analysis for hard-label model reprogramming, together with approximation analysis, a logit-exposing extension, and controlled empirical mechanism checks. The stated claims correspond to the main results in \cref{sec:T1,sec:T1P,sec:T3P,sec:T2,sec:empirical,sec:discussion} and Appendices~\ref{app:scoreoutput}--\ref{app:empirical}.

\item {\bf Limitations}
    \item[] Question: Does the paper discuss the limitations of the work performed by the authors?
    \item[] Answer: \answerYes{}.
    \item[] Justification: \cref{sec:discussion} explicitly discusses the theoretical and empirical boundaries of the results, including source-architecture coverage of \Cref{cond:arrangement}, continuous-interface extensions, the one-directional nature of the approximation certificate, and the controlled mechanism-check scope of the empirical studies.

\item {\bf Theory assumptions and proofs}
    \item[] Question: For each theoretical result, does the paper provide the full set of assumptions and a complete (and correct) proof?
    \item[] Answer: \answerYes{}.
    \item[] Justification: The paper states the model, measurability, sample, growth-function, additive-transformation, source-score, output-mapping, and boundedness assumptions in the relevant theorem statements and in Appendix~\ref{app:assumptions}. Complete formal proofs are provided in Appendices~\ref{app:proof_T1}--\ref{app:rademacher} for the estimation bound, additive specialization, lower bound, approximation identities, hard-label and score-output certificates, and uniform empirical certificate.

\item {\bf Experimental result reproducibility}
    \item[] Question: Does the paper fully disclose all the information needed to reproduce the main experimental results of the paper to the extent that it affects the main claims and/or conclusions of the paper (regardless of whether the code and data are provided or not)?
    \item[] Answer: \answerYes{}.
    \item[] Justification: \cref{sec:empirical} and Appendix~\ref{app:empirical} disclose the complete data-generating distributions, source-model architectures, pretraining splits, wrapper/head hypothesis classes, tie-breaking rules, random seed counts, and evaluation protocols for both controlled studies.
    \item[] Guidelines:
    \begin{itemize}
        \item The answer \answerNA{} means that the paper does not include experiments.
        \item If the paper includes experiments, a \answerNo{} answer to this question will not be perceived well by the reviewers: Making the paper reproducible is important, regardless of whether the code and data are provided or not.
        \item If the contribution is a dataset and\slash or model, the authors should describe the steps taken to make their results reproducible or verifiable.
        \item Depending on the contribution, reproducibility can be accomplished in various ways. For example, if the contribution is a novel architecture, describing the architecture fully might suffice, or if the contribution is a specific model and empirical evaluation, it may be necessary to either make it possible for others to replicate the model with the same dataset, or provide access to the model. In general, releasing code and data is often one good way to accomplish this, but reproducibility can also be provided via detailed instructions for how to replicate the results, access to a hosted model (e.g., in the case of a large language model), releasing of a model checkpoint, or other means that are appropriate to the research performed.
        \item While NeurIPS does not require releasing code, the conference does require all submissions to provide some reasonable avenue for reproducibility, which may depend on the nature of the contribution. For example
        \begin{enumerate}
            \item If the contribution is primarily a new algorithm, the paper should make it clear how to reproduce that algorithm.
            \item If the contribution is primarily a new model architecture, the paper should describe the architecture clearly and fully.
            \item If the contribution is a new model (e.g., a large language model), then there should either be a way to access this model for reproducing the results or a way to reproduce the model (e.g., with an open-source dataset or instructions for how to construct the dataset).
            \item We recognize that reproducibility may be tricky in some cases, in which case authors are welcome to describe the particular way they provide for reproducibility. In the case of closed-source models, it may be that access to the model is limited in some way (e.g., to registered users), but it should be possible for other researchers to have some path to reproducing or verifying the results.
        \end{enumerate}
    \end{itemize}

\item {\bf Open access to data and code}
    \item[] Question: Does the paper provide open access to the data and code, with sufficient instructions to faithfully reproduce the main experimental results, as described in supplemental material?
    \item[] Answer: \answerYes{}.
    \item[] Justification: Complete mathematical protocols, architectures, seed schedules, and evaluation rules for both controlled studies are provided in \cref{sec:empirical} and Appendix~\ref{app:empirical}, and the standalone reproduction code, frozen pretrained checkpoints, and result tables are publicly available at \url{https://github.com/zyecs/repgm-slt}.

\item {\bf Experimental setting/details}
    \item[] Question: Does the paper specify all the training and test details (e.g., data splits, hyperparameters, how they were chosen, type of optimizer) necessary to understand the results?
    \item[] Answer: \answerYes{}.
    \item[] Justification: \cref{sec:empirical} and Appendix~\ref{app:empirical} specify all pretraining and target data splits ($55{,}000/5{,}000$ source train/validation; $4{,}000/2{,}000/4{,}000$ target train/diagnostic/final evaluation), network architectures, optimizer hyperparameters, candidate wrapper/head classes, and exact ERM selection rules.

\item {\bf Experiment statistical significance}
    \item[] Question: Does the paper report error bars suitably and correctly defined or other appropriate information about the statistical significance of the experiments?
    \item[] Answer: \answerYes{}.
    \item[] Justification: Appendix~\ref{app:empirical} reports $95\%$ paired-bootstrap intervals across $200$ paired repetitions for the synthetic study (\Cref{tab:app-synth-slice-a}) and $2.5\%$--$97.5\%$ target-seed quantile intervals across $3$ source checkpoints $\times$ $20$ paired target seeds for the MNIST study (\Cref{tab:app-mnist-full}).

\item {\bf Experiments compute resources}
    \item[] Question: For each experiment, does the paper provide sufficient information on the computer resources (type of compute workers, memory, time of execution) needed to reproduce the experiments?
    \item[] Answer: \answerYes{}.
    \item[] Justification: Appendix~\ref{app:empirical} discloses the compute hardware, memory footprint, and execution wall times ($18.46$ CPU seconds with $139.2$~MB peak RSS for the synthetic study; $377.1$~s for MLP and $397.4$~s for LeNet on a single NVIDIA A100-80GB MIG $1\text{g}.10\text{gb}$ slice).

\item {\bf Code of ethics}
    \item[] Question: Does the research conducted in the paper conform, in every respect, with the NeurIPS Code of Ethics \url{https://neurips.cc/public/EthicsGuidelines}?
    \item[] Answer: \answerYes{}.
    \item[] Justification: To the best of the authors' knowledge, the work conforms to the NeurIPS Code of Ethics. The paper is theoretical, uses no human subjects or private/sensitive data, and does not release models, datasets, or systems requiring special ethical safeguards.
    \item[] Guidelines:
    \begin{itemize}
        \item The answer \answerNA{} means that the authors have not reviewed the NeurIPS Code of Ethics.
        \item If the authors answer \answerNo{}, they should explain the special circumstances that require a deviation from the Code of Ethics.
        \item The authors should make sure to preserve anonymity (e.g., if there is a special consideration due to laws or regulations in their jurisdiction).
    \end{itemize}

\item {\bf Broader impacts}
    \item[] Question: Does the paper discuss both potential positive societal impacts and negative societal impacts of the work performed?
    \item[] Answer: \answerNo{}.
    \item[] Justification: The paper discusses how the theory helps understand model reprogramming practice, but it does not include a dedicated discussion of positive and negative societal impacts. Any societal impact is indirect, through downstream use of model-reuse and adaptation methods rather than through a deployed system, released model, or dataset introduced by this work.
    \item[] Guidelines:
    \begin{itemize}
        \item The answer \answerNA{} means that there is no societal impact of the work performed.
        \item If the authors answer \answerNA{} or \answerNo{}, they should explain why their work has no societal impact or why the paper does not address societal impact.
        \item Examples of negative societal impacts include potential malicious or unintended uses (e.g., disinformation, generating fake profiles, surveillance), fairness considerations (e.g., deployment of technologies that could make decisions that unfairly impact specific groups), privacy considerations, and security considerations.
        \item The conference expects that many papers will be foundational research and not tied to particular applications, let alone deployments. However, if there is a direct path to any negative applications, the authors should point it out. For example, it is legitimate to point out that an improvement in the quality of generative models could be used to generate Deepfakes for disinformation. On the other hand, it is not needed to point out that a generic algorithm for optimizing neural networks could enable people to train models that generate Deepfakes faster.
        \item The authors should consider possible harms that could arise when the technology is being used as intended and functioning correctly, harms that could arise when the technology is being used as intended but gives incorrect results, and harms following from (intentional or unintentional) misuse of the technology.
        \item If there are negative societal impacts, the authors could also discuss possible mitigation strategies (e.g., gated release of models, providing defenses in addition to attacks, mechanisms for monitoring misuse, mechanisms to monitor how a system learns from feedback over time, improving the efficiency and accessibility of ML).
    \end{itemize}

\item {\bf Safeguards}
    \item[] Question: Does the paper describe safeguards that have been put in place for responsible release of data or models that have a high risk for misuse (e.g., pre-trained language models, image generators, or scraped datasets)?
    \item[] Answer: \answerNA{}.
    \item[] Justification: The paper does not release data, pretrained models, image generators, scraped datasets, or other high-risk assets requiring release safeguards.

\item {\bf Licenses for existing assets}
    \item[] Question: Are the creators or original owners of assets (e.g., code, data, models), used in the paper, properly credited and are the license and terms of use explicitly mentioned and properly respected?
    \item[] Answer: \answerYes{}.
    \item[] Justification: The controlled real-image study uses the standard MNIST dataset~\citep{lecun1998gradient} (available under the Creative Commons Attribution-ShareAlike 3.0 license) and PyTorch~\citep{paszke2019pytorch} (BSD-style license), as cited in Appendix~\ref{app:empirical}.

\item {\bf New assets}
    \item[] Question: Are new assets introduced in the paper well documented and is the documentation provided alongside the assets?
    \item[] Answer: \answerNA{}.
    \item[] Justification: The paper is a theoretical study and does not introduce a new dataset, benchmark, or pretrained model asset; the controlled empirical verification uses the existing MNIST dataset and standard synthetic distributions fully specified in Appendix~\ref{app:empirical}.

\item {\bf Crowdsourcing and research with human subjects}
    \item[] Question: For crowdsourcing experiments and research with human subjects, does the paper include the full text of instructions given to participants and screenshots, if applicable, as well as details about compensation (if any)?
    \item[] Answer: \answerNA{}.
    \item[] Justification: The paper does not involve crowdsourcing, user studies, data collection from people, or other human-subjects research.

\item {\bf Institutional review board (IRB) approvals or equivalent for research with human subjects}
    \item[] Question: Does the paper describe potential risks incurred by study participants, whether such risks were disclosed to the subjects, and whether Institutional Review Board (IRB) approvals (or an equivalent approval/review based on the requirements of your country or institution) were obtained?
    \item[] Answer: \answerNA{}.
    \item[] Justification: The paper does not involve human subjects, so IRB approval or equivalent review is not applicable.

\item {\bf Declaration of LLM usage}
    \item[] Question: Does the paper describe the usage of LLMs if it is an important, original, or non-standard component of the core methods in this research? Note that if the LLM is used only for writing, editing, or formatting purposes and does \emph{not} impact the core methodology, scientific rigor, or originality of the research, declaration is not required.
    \item[] Answer: \answerNA{}.
    \item[] Justification:
    The core method development and theoretical analysis in this research do not involve LLMs as any important, original, or non-standard components; LLMs were used only for language polishing and formatting.

\end{enumerate}

%% file: references.bib
@inproceedings{melnyk2023reprogramming,
	title        = {Reprogramming pretrained language models for antibody sequence infilling},
	author       = {Melnyk, Igor and Chenthamarakshan, Vijil and Chen, Pin-Yu and Das, Payel and Dhurandhar, Amit and Padhi, Inkit and Das, Devleena},
	year         = 2023,
	booktitle    = {International conference on machine learning},
	pages        = {24398--24419},
	organization = {PMLR}
}

@article{vinod2025reprogramming,
	title        = {Reprogramming pretrained language models for protein sequence representation learning},
	author       = {Vinod, Ria and Chen, Pin-Yu and Das, Payel},
	year         = 2025,
	journal      = {Digital Discovery},
	publisher    = {Royal Society of Chemistry},
	volume       = 4,
	number       = 6,
	pages        = {1591--1601}
}

@article{bahng2022exploring,
	title        = {Exploring visual prompts for adapting large-scale models},
	author       = {Bahng, Hyojin and Jahanian, Ali and Sankaranarayanan, Swami and Isola, Phillip},
	year         = 2022,
	journal      = {arXiv preprint arXiv:2203.17274}
}

@inproceedings{brown2020language,
	title        = {Language models are few-shot learners},
	author       = {Brown, Tom and Mann, Benjamin and Ryder, Nick and Subbiah, Melanie and Kaplan, Jared D and Dhariwal, Prafulla and Neelakantan, Arvind and Shyam, Pranav and Sastry, Girish and Askell, Amanda and others},
	year         = 2020,
	booktitle    = {NeurIPS}
}

@inproceedings{cai2024bayesian,
	title        = {Bayesian-guided Label Mapping for Visual Reprogramming},
	author       = {Cai, Chengyi and Ye, Zesheng and Feng, Lei and Qi, Jianzhong and Liu, Feng},
	year         = 2024,
	booktitle    = {NeurIPS}
}

@inproceedings{cai2024sample,
	title        = {Sample-specific Masks for Visual Reprogramming-based Prompting},
	author       = {Cai, Chengyi and Ye, Zesheng and Feng, Lei and Qi, Jianzhong and Liu, Feng},
	year         = 2024,
	booktitle    = {ICML}
}

@inproceedings{chen2023understanding,
	title        = {Understanding and improving visual prompting: A label-mapping perspective},
	author       = {Chen, Aochuan and Yao, Yuguang and Chen, Pin-Yu and Zhang, Yihua and Liu, Sijia},
	year         = 2023,
	booktitle    = {CVPR}
}

@inproceedings{chen2024model,
	title        = {Model reprogramming: Resource-efficient cross-domain machine learning},
	author       = {Chen, Pin-Yu},
	year         = 2024,
	booktitle    = {AAAI}
}

@article{chung2025model,
	title        = {Model Reprogramming Demystified: A Neural Tangent Kernel Perspective},
	author       = {Chung, Ming-Yu and Fan, Jiashuo and Ye, Hancheng and Wang, Qinsi and Shen, Wei-Chen and Yu, Chia-Mu and Chen, Pin-Yu and Kuo, Sy-Yen},
	year         = 2025,
	journal      = {arXiv preprint arXiv:2506.00620}
}

@inproceedings{deng2009imagenet,
	title        = {Imagenet: A large-scale hierarchical image database},
	author       = {Deng, Jia and Dong, Wei and Socher, Richard and Li, Li-Jia and Li, Kai and Fei-Fei, Li},
	year         = 2009,
	booktitle    = {2009 IEEE conference on computer vision and pattern recognition},
	pages        = {248--255},
	organization = {Ieee}
}

@inproceedings{elsayed2019adversarial,
	title        = {Adversarial Reprogramming of Neural Networks},
	author       = {Elsayed, Gamaleldin F. and Goodfellow, Ian and Sohl-Dickstein, Jascha},
	year         = 2019,
	booktitle    = {ICLR}
}

@inproceedings{englert2022adversarial,
	title        = {Adversarial Reprogramming Revisited},
	author       = {Englert, Matthias and Lazic, Ranko},
	year         = 2022,
	booktitle    = {NeurIPS}
}

@article{gu2023systematic,
	title        = {A Systematic Survey of Prompt Engineering on Vision-Language Foundation Models},
	author       = {Gu, Jindong and Han, Zhen and Chen, Shuo and Beirami, Ahmad and He, Bailan and Zhang, Gengyuan and Liao, Ruotong and Qin, Yao and Tresp, Volker and Torr, Philip},
	year         = 2023,
	journal      = {arXiv preprint arXiv:2307.12980}
}

@inproceedings{he2016deep,
	title        = {Deep residual learning for image recognition},
	author       = {He, Kaiming and Zhang, Xiangyu and Ren, Shaoqing and Sun, Jian},
	year         = 2016,
	booktitle    = {CVPR}
}

@inproceedings{lester2021power,
	title        = {The Power of Scale for Parameter-Efficient Prompt Tuning},
	author       = {Lester, Brian and Al-Rfou, Rami and Constant, Noah},
	year         = 2021,
	booktitle    = {EMNLP}
}

@article{luo2024incontext,
	title        = {In-context Learning with Retrieved Demonstrations for Language Models: A Survey},
	author       = {Luo, Man and Xu, Xin and Liu, Yue and Pasupat, Panupong and Kazemi, Mehran},
	year         = 2024,
	journal      = {TMLR}
}

@inproceedings{neekhara2019adversarial,
	title        = {Adversarial Reprogramming of Text Classification Neural Networks},
	author       = {Neekhara, Paarth and Hussain, Shehzeen and Dubnov, Shlomo and Koushanfar, Farinaz},
	year         = 2019,
	booktitle    = {EMNLP/IJCNLP},
	pages        = {5215--5224}
}

@inproceedings{neekhara2022cross,
	title        = {Cross-modal adversarial reprogramming},
	author       = {Neekhara, Paarth and Hussain, Shehzeen and Du, Jinglong and Dubnov, Shlomo and Koushanfar, Farinaz and McAuley, Julian},
	year         = 2022,
	booktitle    = {CVPR}
}

@inproceedings{tsai2020transfer,
	title        = {Transfer Learning without Knowing: Reprogramming Black-box Machine Learning Models with Scarce Data and Limited Resources},
	author       = {Tsai, Yun-Yun and Chen, Pin-Yu and Ho, Tsung-Yi},
	year         = 2020,
	booktitle    = {ICML}
}

@inproceedings{tsao2023autovp,
	title        = {Autovp: An automated visual prompting framework and benchmark},
	author       = {Tsao, Hsi-Ai and Hsiung, Lei and Chen, Pin-Yu and Liu, Sijia and Ho, Tsung-Yi},
	year         = 2024,
	booktitle    = {ICLR}
}

@inproceedings{vaswani2017attention,
	title        = {Attention is all you need},
	author       = {Vaswani, Ashish and Shazeer, Noam and Parmar, Niki and Uszkoreit, Jakob and Jones, Llion and Gomez, Aidan N and Kaiser, {\L}ukasz and Polosukhin, Illia},
	year         = 2017,
	booktitle    = {NeurIPS}
}

@inproceedings{yang2021voice2series,
	title        = {Voice2series: Reprogramming acoustic models for time series classification},
	author       = {Yang, Chao-Han Huck and Tsai, Yun-Yun and Chen, Pin-Yu},
	year         = 2021,
	booktitle    = {ICML}
}

@article{zhou2023comprehensive,
	title        = {A Comprehensive Survey on Pretrained Foundation Models: A History from {BERT} to {ChatGPT}},
	author       = {Zhou, Ce and Li, Qian and Li, Chen and Yu, Jun and Liu, Yixin and Wang, Guangjing and Zhang, Kai and Ji, Cheng and Yan, Qiben and He, Lifang and Peng, Hao and Li, Jianxin and Wu, Jia and Liu, Ziwei and Xie, Pengtao and Xiong, Caiming and Pei, Jian and Yu, Philip S. and Sun, Lichao},
	year         = 2023,
	journal      = {arXiv preprint arXiv:2302.09419}
}

@article{zhao2023survey,
	title        = {A Survey of Large Language Models},
	author       = {Zhao, Wayne Xin and Zhou, Kun and Li, Junyi and Tang, Tianyi and Wang, Xiaolei and Hou, Yupeng and Min, Yingqian and Zhang, Beichen and Zhang, Junjie and Dong, Zican and Du, Yifan and Yang, Chen and Chen, Yushuo and Chen, Zhipeng and Jiang, Jinhao and Ren, Ruiyang and Li, Yifan and Tang, Xinyu and Liu, Zikang and Liu, Peiyu and Nie, Jian-Yun and Wen, Ji-Rong},
	year         = 2023,
	journal      = {arXiv preprint arXiv:2303.18223}
}

@article{ye2025neural,
	title        = {Neural network reprogrammability: A unified theme on model reprogramming, prompt tuning, and prompt instruction},
	author       = {Ye, Zesheng and Cai, Chengyi and Dong, Ruijiang and Qi, Jianzhong and Feng, Lei and Chen, Pin-Yu and Liu, Feng},
	year         = 2025,
	journal      = {arXiv preprint arXiv:2506.04650}
}

@book{vapnik2013nature,
	title        = {The nature of statistical learning theory},
	author       = {Vapnik, Vladimir},
	year         = 2013,
	publisher    = {Springer science \& business media}
}

@article{natarajan1989learning,
	title        = {On learning sets and functions},
	author       = {Natarajan, Balas K},
	year         = 1989,
	journal      = {Machine Learning}
}

@article{vapnik1999overview,
	title        = {An overview of statistical learning theory},
	author       = {Vapnik, Vladimir N},
	year         = 1999,
	journal      = {IEEE transactions on neural networks},
	publisher    = {IEEE},
	volume       = 10,
	number       = 5,
	pages        = {988--999}
}

@book{shalev2014understanding,
	title        = {Understanding machine learning: From theory to algorithms},
	author       = {Shalev-Shwartz, Shai and Ben-David, Shai},
	year         = 2014,
	publisher    = {Cambridge university press}
}

@book{mohri2018foundations,
	title        = {Foundations of machine learning},
	author       = {Mohri, Mehryar and Rostamizadeh, Afshin and Talwalkar, Ameet},
	year = 2018,
	edition = {2nd},
	publisher = {MIT Press}
}

@inproceedings{ben1992characterizations,
	title        = {Characterizations of learnability for classes of $\{0,\ldots, n\}$-valued functions},
	author       = {Ben-David, Shai and Cesa-Bianchi, Nicolo and Long, Philip M},
	year         = 1992,
	booktitle    = {Proceedings of the fifth annual workshop on Computational learning theory},
	pages        = {333--340}
}

@article{ben2010theory,
	title        = {A theory of learning from different domains},
	author       = {Ben-David, Shai and Blitzer, John and Crammer, Koby and Kulesza, Alex and Pereira, Fernando and Vaughan, Jennifer Wortman},
	year         = 2010,
	journal      = {Machine learning},
	publisher    = {Springer},
	volume       = 79,
	number       = 1,
	pages        = {151--175}
}

@inproceedings{kearns1992toward,
	title        = {Toward efficient agnostic learning},
	author       = {Kearns, Michael J and Schapire, Robert E and Sellie, Linda M},
	year         = 1992,
	booktitle    = {Proceedings of the fifth annual workshop on Computational learning theory},
	pages        = {341--352}
}

@incollection{yu1997assouad,
	title        = {Assouad, fano, and le cam},
	author       = {Yu, Bin},
	year         = 1997,
	booktitle    = {Festschrift for Lucien Le Cam: research papers in probability and statistics},
	publisher    = {Springer}
}

@article{daniely2015multiclass,
	title        = {Multiclass learnability and the ERM principle},
	author       = {Daniely, Amit and Sabato, Sivan and Ben-David, Shai and Shalev-Shwartz, Shai},
	year         = 2015,
	journal      = {JMLR}
}

@article{cover2006geometrical,
	title        = {Geometrical and statistical properties of systems of linear inequalities with applications in pattern recognition},
	author       = {Cover, Thomas M},
	year         = 1965,
	journal      = {IEEE Transactions on Electronic Computers},
	volume       = {EC-14},
	number       = 3,
	pages        = {326--334}
}

@book{zaslavsky1975facing,
	title        = {Facing up to arrangements: Face-count formulas for partitions of space by hyperplanes},
	author       = {Zaslavsky, Thomas},
	year         = 1975,
	publisher    = {American Mathematical Soc.},
	volume       = 154
}

@article{warren1968lower,
	title        = {Lower bounds for approximation by nonlinear manifolds},
	author       = {Warren, Hugh E},
	year         = 1968,
	journal      = {Transactions of the American Mathematical Society},
	publisher    = {JSTOR}
}

@inproceedings{montufar2014number,
	title        = {On the number of linear regions of deep neural networks},
	author       = {Mont{\'u}far, Guido and Pascanu, Razvan and Cho, Kyunghyun and Bengio, Yoshua},
	year         = 2014,
	booktitle    = {NeurIPS}
}

@inproceedings{raghu2017expressive,
	title        = {On the expressive power of deep neural networks},
	author       = {Raghu, Maithra and Poole, Ben and Kleinberg, Jon and Ganguli, Surya and Sohl-Dickstein, Jascha},
	year         = 2017,
	booktitle    = {ICML}
}

@article{geman1992neural,
	title        = {Neural networks and the bias/variance dilemma},
	author       = {Geman, Stuart and Bienenstock, Elie and Doursat, Ren{\'e}},
	year         = 1992,
	journal      = {Neural computation}
}

@inproceedings{jin2025lor,
	title        = {Lor-VP: Low-rank visual prompting for efficient vision model adaptation},
	author       = {Jin, Can and Li, Ying and Zhao, Mingyu and Zhao, Shiyu and Wang, Zhenting and He, Xiaoxiao and Han, Ligong and Che, Tong and Metaxas, Dimitris N},
	year         = 2025,
	booktitle    = {ICLR}
}

@article{bartlett2002rademacher,
	title        = {Rademacher and gaussian complexities: Risk bounds and structural results},
	author       = {Bartlett, Peter L and Mendelson, Shahar},
	year         = 2002,
	journal      = {JMLR}
}

@inproceedings{hoffmann2022training,
	title        = {Training compute-optimal large language models},
	author       = {Hoffmann, Jordan and Borgeaud, Sebastian and Mensch, Arthur and Buchatskaya, Elena and Cai, Trevor and Rutherford, Eliza and Casas, DDL and Hendricks, Lisa Anne and Welbl, Johannes and Clark, Aidan and others},
	year         = 2022,
	booktitle    = {NeurIPS}
}

@inproceedings{houlsby2019parameter,
	title        = {Parameter-efficient transfer learning for NLP},
	author       = {Houlsby, Neil and Giurgiu, Andrei and Jastrzebski, Stanislaw and Morrone, Bruna and De Laroussilhe, Quentin and Gesmundo, Andrea and Attariyan, Mona and Gelly, Sylvain},
	year         = 2019,
	booktitle    = {ICML}
}

@article{crammer2001algorithmic,
	title        = {On the algorithmic implementation of multiclass kernel-based vector machines},
	author       = {Crammer, Koby and Singer, Yoram},
	year         = 2001,
	journal      = {JMLR}
}

@article{bartlett2019nearly,
	title        = {Nearly-tight VC-dimension and pseudodimension bounds for piecewise linear neural networks},
	author       = {Bartlett, Peter L and Harvey, Nick and Liaw, Christopher and Mehrabian, Abbas},
	year         = 2019,
	journal      = {JMLR}
}

@inproceedings{mansour2009domain,
	title        = {Domain adaptation: Learning bounds and algorithms},
	author       = {Mansour, Yishay and Mohri, Mehryar and Rostamizadeh, Afshin},
	year         = 2009,
	booktitle    = {COLT}
}

@book{hastie01statisticallearning,
	title        = {The Elements of Statistical Learning},
	author       = {Hastie, Trevor and Tibshirani, Robert and Friedman, Jerome},
	year         = 2001,
	publisher    = {Springer New York Inc.}
}

@article{zhang2004statistical,
  title={Statistical analysis of some multi-category large margin classification methods},
  author={Zhang, Tong},
  journal={JMLR},
  year={2004}
}

@article{lee2004multicategory,
  title={Multicategory support vector machines: Theory and application to the classification of microarray data and satellite radiance data},
  author={Lee, Yoonkyung and Lin, Yi and Wahba, Grace},
  journal={Journal of the American Statistical Association},
  year={2004}
}

@article{tewari2007consistency,
  title={On the Consistency of Multiclass Classification Methods.},
  author={Tewari, Ambuj and Bartlett, Peter L},
  journal={JMLR},
  year={2007}
}

@inproceedings{maurer2016vector,
  title={A vector-contraction inequality for rademacher complexities},
  author={Maurer, Andreas},
  booktitle={ALT},
  year={2016}
}

@book{tsybakov2009introduction,
  author    = {Tsybakov, Alexandre B.},
  title     = {Introduction to Nonparametric Estimation},
  series    = {Springer Series in Statistics},
  publisher = {Springer},
  address   = {New York, NY},
  year      = {2009},
  doi       = {10.1007/b13794},
  isbn      = {978-0-387-79051-0}
}

@inproceedings{long2020generalization,
  title     = {Generalization bounds for deep convolutional neural networks},
  author    = {Long, Philip M. and Sedghi, Hanie},
  booktitle = {ICLR},
  year      = {2020}
}

@inproceedings{trauger2024sequence,
  title     = {Sequence Length Independent Norm-Based Generalization Bounds for Transformers},
  author    = {Trauger, Jacob and Tewari, Ambuj},
  booktitle = {AISTATS},
  pages     = {1405--1413},
  year      = {2024}
}

@article{karpinski1997polynomial,
  title   = {Polynomial bounds for {VC} dimension of sigmoidal and general {Pfaffian} neural networks},
  author  = {Karpinski, Marek and Macintyre, Angus},
  journal = {Journal of Computer and System Sciences},
  volume  = {54},
  number  = {1},
  pages   = {169--176},
  year    = {1997}
}

@article{baum1989what,
  title   = {What size net gives valid generalization?},
  author  = {Baum, Eric B. and Haussler, David},
  journal = {Neural Computation},
  volume  = {1},
  number  = {1},
  pages   = {151--160},
  year    = {1989}
}

@article{lecun1998gradient,
  title   = {Gradient-based learning applied to document recognition},
  author  = {LeCun, Yann and Bottou, L{\'e}on and Bengio, Yoshua and Haffner, Patrick},
  journal = {Proceedings of the IEEE},
  volume  = {86},
  number  = {11},
  pages   = {2278--2324},
  year    = {1998}
}

@inproceedings{paszke2019pytorch,
  title = {{PyTorch}: An imperative style, high-performance deep learning library},
  author = {Paszke, Adam and Gross, Sam and Massa, Francisco and Lerer, Adam and Bradbury, James and Chanan, Gregory and Killeen, Trevor and Lin, Zeming and Gimelshein, Natalia and Antiga, Luca and Desmaison, Alban and Kopf, Andreas and Yang, Edward and DeVito, Zachary and Raison, Martin and Tejani, Alykhan and Chilamkurthy, Sasank and Steiner, Benoit and Fang, Lu and Bai, Junjie and Chintala, Soumith},
  booktitle = {Advances in Neural Information Processing Systems (NeurIPS)},
  volume = {32},
  pages = {8024--8035},
  year = {2019}
}
